\documentclass[mnsc,nonblindrev]{informs3_hide}

\OneAndAHalfSpacedXI 

\usepackage{mathabx, makecell}
\usepackage{natbib, multirow, multicol, xspace, enumitem, amsmath, pifont, algorithm, algorithmic, subcaption}
\usepackage{accents}
\usepackage{graphicx}
\usepackage{multicol}
\bibpunct[, ]{(}{)}{,}{a}{}{,}%
\def\bibfont{\small}%
\usepackage{bm}
\usepackage[normalem]{ulem}
\usepackage[dvipsnames]{xcolor}
\usepackage[all]{xy}
\usepackage{hyperref}
\hypersetup{
    colorlinks=true,
    linkcolor=blue,
    filecolor=magenta,      
    urlcolor=cyan,
    citecolor=blue
    }
\usepackage{ulem}
\usepackage{framed}

\newcommand{\defeq}{\overset{\text{\tiny def}}{=}}
\newcommand{\unld}{\underline{d}}
\newcommand{\ovld}{\overline{d}}
\theoremstyle{TH}%
\newtheorem{observation}{Observation}

\TheoremsNumberedThrough     
\ECRepeatTheorems

\EquationsNumberedThrough    

\MANUSCRIPTNO{}

\begin{document}

\RUNAUTHOR{Chen et al.}

\RUNTITLE{A Minimax-MDP Framework for Learning-augmented Problems}

\TITLE{A Minimax-MDP Framework with Future-imposed Conditions for Learning-augmented Problems}

\ARTICLEAUTHORS{
\AUTHOR{Xin Chen\footnotemark[1]}
\AFF{H. Milton Stewart School of Industrial and Systems Engineering, Georgia Institute of Technology, Atlanta, Georgia 30332, USA, \EMAIL{xin.chen@isye.gatech.edu}}
 \AUTHOR{Yuze Chen\footnotemark[1]}
 \AFF{Qiuzhen College, Tsinghua University, Beijing 100084, China, \EMAIL{yz-chen21@mails.tsinghua.edu.cn}}
 \AUTHOR{Yuan Zhou\footnotemark[1]}
 \AFF{Yau Mathematical Sciences Center \& Department of Mathematical  Sciences, Tsinghua University, Beijing 100084, China, \\ Beijing Institute of Mathematical Sciences and Applications, Beijing 101408, China, \EMAIL{yuan-zhou@tsinghua.edu.cn}}
}
\renewcommand{\thefootnote}{\fnsymbol{footnote}}
\footnotetext[1]{Author names listed in alphabetical order.}
\renewcommand{\thefootnote}{\arabic{footnote}}

\ABSTRACT{We study a class of sequential decision-making problems with augmented predictions, potentially provided by a machine learning algorithm.  In this setting, the decision-maker receives prediction intervals for unknown parameters that become progressively refined over time, and seeks decisions that are competitive with the hindsight optimal under all possible realizations of both parameters and predictions. We propose a minimax Markov Decision Process (minimax-MDP) framework, where the system state consists of an adversarially evolving environment state and an internal state controlled by the decision-maker. We introduce a set of \emph{future-imposed conditions} that characterize the feasibility of minimax-MDPs and enable the design of efficient, often closed-form, robustly competitive policies. We illustrate the framework through three applications: multi-period inventory ordering with refining demand predictions, resource allocation with uncertain utility functions, and a multi-phase extension of the minimax-MDP applied to the inventory problem with time-varying ordering costs. Our results provide a tractable and versatile approach to robust online decision-making under predictive uncertainty.
}

\maketitle

\section{Introduction}

In recent years, \emph{learning-augmented algorithms}, also known as \emph{robust decision-making with predictions}, have emerged as a popular framework that integrates algorithmic decision-making with machine-learned predictions. These algorithms aim to exploit predictive information to improve performance, while preserving rigorous guarantees when facing uncertainty. A central focus in this area is to design algorithms that are \emph{prediction-aware} yet \emph{robust}, performing well even when the structure or quality of the predictive information varies. This line of work has found various operational applications in inventory control~\citep{feng2024robust}, scheduling~\citep{lattanzi2020online,li2021online}, energy systems and sustainability~\citep{lechowicz2024online}, online matching~\citep{jin2022online}, and dynamic resource allocation~\citep{mahdian2007allocating}, where predictive signals, although often uncertain, can significantly inform better decisions.

In this paper, we study a class of sequential decision-making problems where the decision-maker, before making each decision, receives a prediction interval for key problem parameters (e.g., future demand or production elasticity). These intervals are assumed to be valid in the sense that they always contain the true parameter values, but may initially be coarse and become increasingly refined over time --- reflecting a common setting in practice where predictive accuracy improves with more data and a shorter forecasting horizon. The decision-maker must make irrevocable decisions over time, without knowing the exact realization of the parameter within each interval. The goal is to design robust online algorithms whose performance is \emph{competitive with the hindsight optimal solution}, under the worst-case realization of the parameters and the prediction intervals. This setting naturally fits within the learning-augmented framework, where predictive information is  trusted but incomplete, and highlights the algorithmic challenges in balancing \emph{adaptivity to improving information} with \emph{robustness to uncertainty}.

\subsection{Our Contributions}

\noindent \textbf{The minimax-MDP framework.} To address the algorithmic challenges in the above setting, we introduce a general \emph{minimax Markov Decision Process (minimax-MDP)} framework (Section~\ref{sec:minimax-MDP-formulation}). In this formulation, the system state consists of two components: an \emph{environment state}, which captures exogenous factors such as the evolving prediction intervals, and an \emph{internal state}, which reflects the decision-maker’s own dynamics and is influenced by their actions. The environment state evolves independently of the decision-maker's actions and is modeled adversarially: at each time step, the next environment state is chosen from a prescribed set of feasible refinements. This adversarial modeling captures the uncertainty in the evolution of predictive information while preserving the structural assumption that all predictions remain valid. The minimax-MDP framework provides a powerful abstraction for analyzing online decisions that robustly utilize progressively refining but uncertain predictive information. 

We focus on a special class of minimax-MDPs in which the internal state is a one-dimensional real variable (which will also be referred to as the \emph{inventory level}) and the decision-maker takes actions by making additive adjustments to this state. Both the internal state and the action are constrained above and below, with the bounds determined by the environment state. This setting naturally models many operations problems, such as inventory management and resource allocation, where the internal state may represent the inventory level or cumulative investment in a project. As we will see later in the paper, for many such problems, a policy with a prescribed robustness guarantee exists \emph{if and only if} the corresponding minimax-MDP admits a feasible policy that satisfies all embedded constraints.

\medskip
\noindent \textbf{The future-imposed conditions for solving minimax-MDPs.}  In Section~\ref{sec:mainthm}, we establish our main result under the minimax-MDP framework --- our main theorem reduces the feasibility problem for minimax-MDPs to checking a family of \emph{future-imposed conditions}, where an inequality condition should hold for each time period $\alpha$ and each potential environment state $s_\alpha$ at time $\alpha$. The intuition is as follows: a future environment state at some time $\beta \geq \alpha$ may impose an upper bound on the internal state, namely $x_\alpha$, at time $\alpha$, which is a necessary condition so that a feasible policy can evolve from $(s_\alpha, x_\alpha)$. Specifically, regardless of how the environment evolves from time $\alpha$ to $\beta$, if the decision-maker consistently selects the minimum allowable actions, the internal state must stay below the upper bound at time $\beta$. This constraint induces a \emph{future-imposed upper bound} on $x_\alpha$. Similarly, by considering maximum allowable actions and the lower bound constraints for the internal state at a future time step, we obtain a \emph{future-imposed lower bound} constraint on $x_\alpha$. Thus, a necessary condition for the existence of a feasible policy is that the lower bound does not exceed the upper bound, which forms the prototype of the future-imposed condition. Surprisingly, we prove that a collection of slightly refined versions of these future-imposed conditions is not only necessary but also sufficient for the minimax-MDP to admit a feasible policy. This result reduces the task of solving a minimax-MDP to verifying a set of future-imposed inequality conditions.

The future-imposed conditions have simple forms. As a result, our minimax-MDP framework leads to highly efficient solutions for a wide range of robust decision-making problems with predictions. These operational applications span a variety of prediction structures, objective functions, and competitiveness metrics. In many cases, our framework yields \emph{closed-form solutions}, significantly advancing the tractability and practical relevance of robust online decision-making under structured predictive information.

\medskip
\noindent \textbf{Application I: robust multi-period ordering with predictions (RMOwP).}  To demonstrate the versatility of our minimax-MDP framework, in Section~\ref{sec:app-I-robust-multi-period-ordering}, we apply the framework to a robust variation of the multi-period newsvendor problem, RMOwP, where a decision-maker places orders over several preparation days in anticipation of uncertain demand on a final operating day. Each day, the decision-maker receives a prediction interval for the final demand, which becomes progressively narrower as more information becomes available. The goal is to make ordering decisions that remain competitive with the optimal hindsight solution under all valid (and potentially adversarial) sequences of prediction intervals. This problem captures practical settings such as workforce planning, where recruitment decisions are made gradually and incrementally based on improving, but still uncertain, demand forecasts (see \citet{feng2024robust}, as well as the discussion in Section~\ref{sec:app-I-robust-multi-period-ordering}). In this problem, we consider two metrics to measure the competitiveness of the robust online decision policy: \emph{regret} and \emph{competitive ratio}. Our future-imposed conditions yield optimal policies under both metrics.

A recent work by \citet{feng2024robust} studied the same problem by observing that the adversarial predictions can be restricted to a ``single-switching'' structure, and proposed an LP-based emulator technique to derive robust policies. In contrast, our minimax-MDP framework offers three key advantages. First, our approach yields explicit closed-form expressions for the optimal competitiveness, which can be computed in time linear in the problem size. In comparison, \citet{feng2024robust} obtain the optimal competitiveness by solving a linear program. Second, their algorithm critically depends on structural properties specific to the regret metric to justify the single-switching assumption, making it unclear how to extend their method to the competitive ratio setting (please refer to Section~\ref{sec:app-I-robust-competitive-ratio} for a detailed discussion). In contrast, our approach applies uniformly to both metrics. Third, our minimax-MDP framework naturally extends to a more general setting with \emph{scale-dependent prediction sequences}, where the length of prediction intervals may vary with the predicted demand scale, which could be a common feature in practice, as higher-demand scenarios typically carry greater uncertainty under the same amount of historical information. A summary of the comparison between our minimax-MDP-based algorithms and \citet{feng2024robust} can be found in Table~\ref{tab:comparison-between-our-results-for-app-I-and-related-work}.

\begin{table}[t]
  \centering
  \caption{Comparison between our results for RMOwP and \citet{feng2024robust}.}
  \label{tab:comparison-between-our-results-for-app-I-and-related-work}
  \renewcommand{\arraystretch}{2.1}
\scalebox{0.9}{
\begin{tabular}{l|m{4.5cm}|m{4.5cm}|m{4.5cm}}
\hline
&  Optimal robust regret & \makecell[l]{Optimal robust\\  competitive ratio} & \makecell[l]{Scale-dependent \\ prediction sequence} \\ 
\hline
\cite{feng2024robust}       & \makecell[l]{Solving a polynomial-size \\linear program$^{*}$}           & \makecell[c]{\textbf{---}}                  & \makecell[c]{\textbf{---}}  \\ \hline
Our results & \makecell[l]{\emph{Closed-form} solution\\ in \emph{linear} time$^{*}$}      & \makecell[l]{\emph{Closed-form} solution\\ in \emph{linear} time$^{*}$}         & \makecell[l]{\emph{Closed-form} solution\\ in polynomial time$^{*,**}$}     \\ \hline
\multicolumn{4}{l}{\makecell[l]{\small $^{*~}$ With respect to problem size.\\
$^{**}$ For illustration purposes, we present a solution for the optimal robust competitive ratio; however, \\ our method can also derive the optimal robust regret.}} \\
\end{tabular}
}
\end{table}

\medskip
\noindent \textbf{Application II: robust resource allocation with predictions (RRAwP).} In Section~\ref{sec:robust-resource-allocation-with-predictions}, we present the second application of our minimax-MDP framework: the RRAwP problem. In this setting, a decision-maker allocates budgeted resources between two sectors over multiple periods, aiming to maximize a cumulative utility function that depends on an unknown parameter vector. The utility function can be drawn from a broad class satisfying mild assumptions such as monotonicity and concavity, encompassing commonly used forms including perfect substitutes (linear), perfect complements (Leontief), and Cobb-Douglas utilities. At each time step, the decision-maker receives prediction intervals for the unknown parameters, which become progressively more accurate as additional data becomes available. This modeling framework captures a variety of practical scenarios, such as investment planning in manufacturing, where capital and labor are allocated over time based on evolving productivity estimates, or public budgeting, where resources are incrementally distributed between sectors like education and healthcare.

Unlike Application I, the utility function in RRAwP takes more general forms and may involve multiple parameters (e.g., one elasticity parameter per sector). As a result, the decision-maker receives multiple prediction intervals at each time step. The flexible environment-state structure of our minimax-MDP framework naturally accommodates this richer setting. Using the Leontief production function as an example, we demonstrate that our future-imposed conditions yield a \emph{closed-form} expression for the optimal robust competitive ratio, which can be computed in \emph{linear time} with respect to the problem size.

\medskip
\noindent \textbf{Multi-phase minimax-MDP and Application III: RMOwP under changing costs.} We further consider an extension of the RMOwP problem (introduced in Application I) in which the ordering costs may vary across preparation days. This setting can be modeled as a \emph{multi-phase minimax-MDP}, where each phase corresponds to a time interval during which the ordering cost remains constant and is represented by an ordinary minimax-MDP. In Section~\ref{sec:multi-phase-minimax-mdp}, we formally define the multi-phase minimax-MDP and introduce a phase reduction theorem, which enables us to reduce the feasibility problem of a $K$-phase minimax-MDP to that of a $(K-1)$-phase minimax-MDP by applying the future-imposed conditions to the final phase. By recursively applying this reduction $(K-1)$ times, the problem is ultimately reduced to a standard minimax-MDP, which can then be solved by directly verifying the future-imposed conditions.

In Section~\ref{sec:Multi-period-Ordering-Decisions-with-Multi-phase-costs-and-Predictions}, we apply the phase reduction theorem to the RMOwP problem under changing costs. While powerful, each phase reduction process incurs a polynomial blow-up in the size of the resulting multi-phase minimax-MDP. Consequently, the algorithm for computing the optimal robust policy has a worst-case runtime of $\exp((\log T) \cdot 2^K)$, which can be further improved to $T^{O(K)}$ by exploiting the specific structure of the RMOwP problem. Here, $T$ denotes the number of preparation days and $K$ is the number of cost changes (equivalently, the number of phases). This runtime remains polynomial only when $K$ is constant. To handle the general case, we further develop an algorithm that reduces the number of cost changes while approximately preserving the robust competitive ratio. By combining this reduction with the phase reduction technique, we obtain a polynomial-time approximation scheme (PTAS) that, for any $\epsilon > 0$, computes a policy achieving the optimal robust competitive ratio up to an $\epsilon$ error in time $T^{O(1/\epsilon)}$. As in the previous applications, our algorithms here can also be adapted to the regret-based competitiveness metric.

We also note that \citet{feng2024robust} considered a version of the RMOwP problem with changing ordering costs. However, as discussed in Application I, their algorithm is limited to the regret metric and does not extend to competitive ratio. Moreover, their definition of regret differs from ours: their regret directly (and only) includes the full ordering cost incurred by the online algorithm (referred to as hiring cost in their context), whereas our regret is defined as the difference between the hindsight-optimal profit and the profit of the online algorithm. In particular, the cost component of our regret reflects the difference in ordering costs between the two policies. Due to this difference, our results in this application are not directly comparable with those of \citet{feng2024robust}.

\section{Related Works}

\medskip
\noindent\textbf{Learning-augmented problems.} The learning-augmented framework, first introduced by \citet{vee2010optimal} and \citet{mahdian2012online}, aims to enhance the worst-case performance of online algorithms by incorporating predictions (or advice), typically generated by machine learning models. While these predictions may be unreliable or even adversarially chosen, learning-augmented algorithms are designed to benefit from correct predictions while maintaining robustness against incorrect ones. This influential research direction has been applied to a broad range of online decision problems, including ski rental~\citep{purohit2018improving}, caching~\citep{lykouris2021competitive}, online matching~\citep{dinitz2021faster,jin2022online,chen2022faster}, energy-efficient scheduling~\citep{balkanski2023energy}, single-leg revenue management~\citep{balseiro2023single}, online resource allocation~\citep{an2024best,golrezaei2023online}, online lead time quotation~\citep{huo2024online}, and Nash social welfare maximization~\citep{banerjee2022online}. In comparison, \citet{feng2024robust} and our work represent a complementary line of research that refines the adversarial modeling paradigm by incorporating predictions whose quality improves over time in an adversarially controlled manner.

\medskip
\noindent\textbf{Robust Markov Decision Processes.} Robust Markov Decision Processes (RMDPs), first introduced by \citet{iyengar2005robust} and \citet{nilim2005robust}, address uncertainty in transition probabilities by optimizing policies for the worst-case scenario within a prescribed uncertainty set. This framework is motivated by the uncertainties arising from parameter estimation errors~\citep{mannor2007bias}, especially those caused by noisy or corrupted data~\citep{zhang2021robust}, discrepancies between simulators and real-world environments (i.e., the sim-to-real gap)~\citep{pinto2017robust,tessler2019action,li2022policy}, and the presence of unavoidable missing information~\citep{he2023robust}. In comparison, our minimax-MDP formulation focuses on a specialized class of RMDPs where the transition dynamics are binary, akin to the setting studied in \citet{bertsekas1971control}. Furthermore, the state in our model is composed of two components: an environment state that evolves under uncertainty and a one-dimensional internal state with known transition rules. This structural decomposition is crucial for deriving our future-imposed conditions, while still retaining the flexibility to model a range of learning-augmented decision-making problems.

\medskip
\noindent\textbf{Inventory control.}
Inventory control, a fundamental problem in operations research and supply chain management, focuses on determining optimal inventory policies that meet customer demand while minimizing associated costs. A central challenge lies in balancing the trade-off between overage and underage costs. The classical newsvendor model~\citep{edgeworth1888mathematical} captures this trade-off under demand uncertainty. Our RMOwP problems (Applications I and III) can be viewed as multi-period extensions of the newsvendor model, where a single target demand is considered and the decision-maker incrementally adjusts the final ordering quantity over multiple periods, which is similar in spirit to the model studied by \citet{song2012newsvendor}. While early works on the newsvendor model and its extensions assume full knowledge of the demand distribution~\citep{arrow1951optimal,clark1960optimal}, more recent research addresses the lack of distributional information through robust optimization~\citep{bertsimas2006robust}, distributionally robust optimization~\citep{xin2022distributionally}, and data-driven approaches~\citep{levi2007nearoptimal,besbes2013implications,levi2015data,lin2022data}. In contrast, our RMOwP problems introduce a different form of robustness that integrates the (potentially data-driven) predictions: the demand information is revealed gradually through a black-box machine learning algorithm, with prediction quality improving over time in an adversarial manner.


\medskip
\noindent\textbf{Resource allocation.} Effective resource allocation is a fundamental challenge across economics, operations, and computer systems. 
A central framework for addressing this challenge is utility maximization under budget constraints --- introduced and developed by \citet{jevons1871theory}, \citet{edgeworth1881mathematical}, and \citet{walras1900elements} --- which captures the trade-off between resource consumption and decision-maker preferences. While some works address this problem assuming full knowledge of the utility function~\citep{srikantan1963problem,bitran1981disaggregation,bretthauer1995nonlinear} (see the survey in~\citet{patriksson2008survey} for further references), another stream of works considers settings with unknown utility or preferences, leading to a variety of estimation techniques. These include Afriat’s Efficiency Index~\citep{afriat1967construction}, maximum likelihood estimation~\citep{aigner1977formulation}, the Kmenta approximation~\citep{kmenta1967estimation}, the UTA method~\citep{jacquet1982assessing}, inverse optimization~\citep{keshavarz2011imputing}, two-step instrumental variables (TSIV) approach~\citep{de2021estimating}, and inverse reinforcement learning~\citep{grzeskiewicz2025uncovering}.
In our RRAwP problem (Application II), we demonstrate how to robustly optimize the allocation policy when the utility function is only partially known and iteratively refined, potentially using any of the aforementioned estimation methods, within a prediction-driven, adversarially evolving environment.

\section{The Minimax-MDP Framework}
\label{sec:minimax-MDP-formulation}

We consider a \emph{minimax Markov Decision Process (minimax-MDP)}, introduced through a natural inventory management setting where the internal state is interpreted as the inventory level, though it may represent any real-valued quantity depending on the application. The minimax-MDP framework consists of $T$ discrete time periods, where the system state at time $t$ is represented as a pair $(s_t, x_t)$, with $s_t$ denoting the environment state and $x_t \in \mathbb{R}$ representing the system's inventory level. 
The initial inventory level is given by:
\begin{align}\label{eq:minimax-MDP-initial-inventory-level}
x_1 = 0 .
\end{align}
At any time $t \in [T-1]$, the decision-maker needs to select an ordering action $a_t \in \mathbb{R}$ and the inventory level at the next time period becomes
\begin{align}
x_{t+1} = x_t + a_t .
\end{align}
Unlike standard Markov Decision Processes, where policies are optimized for expected performance, the minimax-MDP requires a feasible strategy under all possible transition trajectories, motivating the term ``minimax''.  The environment state influences action feasibility and inventory constraints (which will be soon described in detail) but evolves independently of the decision-maker's actions. 

We formally define a minimax-MDP as follows.
\begin{definition}
\label{def:minimax-MDP}
A \emph{minimax Memoryless Decision Process (minimax-MDP)} is defined as a tuple:
\[
\mathcal{I} = \left\{T, \{S_t\}_{t \in [T]}, s_1, x_1, \{F_t\}_{t \in [T-1]},\{U_t, V_t\}_{t \in [T-1]}, \{L_t, R_t\}_{t \in [T]} \right\} ,
\]
where 
\begin{enumerate}
\item \emph{Time horizon} $T$ is the total number of discrete time periods.
\item \emph{Environment state set} $S_t$ for each $t \in [T]$ consists of all possible environment states at time $t$.
\item \emph{Initial environment state} $s_1 \in S_1$ is the state of the environment at time $1$.
\item \emph{Initial inventory level} $x_1$ is as introduced at the beginning of this section.
\item \emph{Environment state transition function} $F_t : S_t  \to \mathcal{P}(S_{t+1}) \setminus \{ \emptyset \}$, where $\mathcal{P}(S)$ denotes the power set of $S$, specifies the set of possible next states for each $s_t$. That is, for any $s_t \in S_t$, the environment may transit to any state in $F_t(s_t)$ at time $t+1$.
\item \emph{Environment-dependent action constraints} $U_t, V_t : S \to \mathbb{R}$ restricts the action $a_t$ at time $t \in [T-1]$ to the interval:
\begin{align} \label{eq:environment-dependent-action-constraint}
U_t(s_t) \leq a_t \leq V_t(s_t);
\end{align}
\item \emph{Environment-dependent inventory constraints} $L_t, R_t : S_t \to \mathbb{R}$ require the inventory level $x_t$ at each time $t \in [T]$ to satisfy
\begin{align} \label{eq:environment-dependent-inventory-constraint}
L_t(s_t) \leq x_t \leq R_t(s_t),
\end{align}
where for convenience, we set
\begin{align}\label{eq:minimax-MDP-initial-LR}
L_1(s_1) = R_1(s_1) = x_1 = 0 .
\end{align}
\end{enumerate}
\end{definition}

A policy $\pi = (\pi_1, \pi_2, \dots, \pi_{T-1})$ maps the system state at any time $t \in \{1, 2, \dots, T - 1\}$ to an action: 
\begin{align}\label{eq:definition-of-policy}
\pi_t : S_t \times \mathbb{R} \to \mathbb{R}, \qquad \forall t \in \{1, 2, \dots, T - 1\}.
\end{align}

Given a policy $\pi$, we say a trajectory $\phi$ is \emph{$\pi$-compatible} if it can occur under the system $\mathcal{I}$ when following the policy $\pi$. Formally, we define:
\begin{definition}\label{def:pi-compatible-trajectory}
For two time periods with $\alpha, \beta \in [T]$ and $\alpha \leq \beta$, a sequence $\phi = (s_\alpha, x_\alpha, s_{\alpha + 1}, x_{\alpha + 1}, \dots, s_\beta, x_\beta)$ is called a \emph{full trajectory} (or simply \emph{trajectory}) if $\alpha = 1$ and $\beta = T$ and a \emph{partial trajectory} otherwise. A (partial) trajectory $\phi$ is said to be \emph{$\pi$-compatible} under the system $\mathcal{I}$ if it satisfies
\begin{align}
s_{t+1} \in F_t(s_t), \qquad \text{and} \qquad a_t = x_{t+1}-x_t = \pi_t(s_t, x_t)
\end{align}
for all $t \in \{\alpha, \alpha + 1, \dots, \beta - 1\}$. 

Furthermore, a (partial) trajectory $\phi$ is deemed  \emph{feasible} if it satisfies all environment-dependent action and inventory constraints, i.e., it satisfies Eq.~\eqref{eq:environment-dependent-action-constraint} and Eq.~\eqref{eq:environment-dependent-inventory-constraint} for all $t \in \{\alpha, \alpha + 1, \dots, \beta - 1\}$.
\end{definition}

\medskip
\noindent \uline{{\bfseries The Feasible Policy Problem.}} Given a minimax-MDP $\mathcal{I}$, a policy $\pi$ is said to be \emph{feasible} if every $\pi$-compatible trajectory under $\mathcal{I}$ is also feasible. In other words, a feasible policy $\pi$ ensures that all possible trajectories generated under $\pi$ and $\mathcal{I}$ satisfy all environment-dependent action and inventory constraints. The main question studied in this work is:
\begin{center}
\begin{framed}
{\it Given a minimax-MDP, is there an efficient algorithm to determine whether a feasible policy exists, and if so, to compute such a policy?}
\end{framed}
\end{center}

In the following sections, we will present an affirmative answer to the above question, and demonstrate its applicability in several practically meaningful scenarios.

\section{Solving the Minimax-MDPs}
\label{sec:mainthm}

This section aims to provide our main efficient algorithms to solve a minimax-MDP: to determine whether there is a feasible policy and to compute such a policy if there is one.

\subsection{Additional Notations}\label{sec:additional-notations}

To facilitate our analysis, we first introduce additional notations that extend those in Section~\ref{sec:minimax-MDP-formulation}.

\medskip
\noindent \underline{Compatible environment state trajectories.} For any two time periods $\alpha, \beta \in [T]$ with $\alpha \leq \beta$, we define a \emph{(partial) environment state trajectory} $(s_\alpha, s_{\alpha+1}, \dots, s_\beta) \in S_\alpha \times S_{\alpha+1}  \times \dots \times S_{\beta}$ as \emph{compatible} if $s_{t+1} \in F_t(s_t)$ holds for all $t \in \{\alpha, \alpha+1, \dots, \beta-1\}$. That is, such an environment state trajectory may occur if we start from state $s_\alpha$ at time $\alpha$. 
\begin{observation} \label{obs:compatible-trajectory-environment-state-trajectory}
It is straightforward to verify that the following statements:
\begin{enumerate}[label=(\alph*)]
\item \label{obs:compatible-trajectory-environment-state-trajectory-a} For any policy $\pi$, a $\pi$-compatible (partial) trajectory $\phi = (s_\alpha, x_\alpha, s_{\alpha + 1}, x_{\alpha + 1}, \dots, s_\beta, x_\beta)$ naturally induces a compatible (partial) environment state trajectory $(s_\alpha, s_{\alpha + 1}, \dots, s_\beta)$.
\item \label{obs:compatible-trajectory-environment-state-trajectory-b} For any policy $\pi = (\pi_1, \pi_2, \dots, \pi_{T-1})$ and initial inventory level $x_\alpha$, a compatible (partial) environment state trajectory $(s_\alpha, s_{\alpha + 1}, \dots, s_\beta)$ naturally introduces a $\pi$-compatible (partial) trajectory $(s_\alpha, x_\alpha, s_{\alpha + 1}, x_{\alpha + 1}, \dots, s_\beta, x_{\beta})$ where $x_{t + 1} = x_t + \pi_t(s_t, x_t)$ for $t \in \{\alpha, \alpha + 1, \dots, \beta - 1\}$.
\end{enumerate}    
\end{observation} 

\medskip
\noindent \underline{Multi-time-period environment state transition.} For any two time periods $\alpha, \beta \in [T]$ with $\alpha \leq \beta$, we define the transition function $F_{\alpha\to \beta} : S_\alpha \to \mathcal{P}(S_\beta)$ as follows:
\begin{align}\label{eq:def-F-a-to-b}
F_{\alpha\to \beta}(s_\alpha) \defeq \{s_\beta \in S_\beta : \exists (s_{\alpha+1}, \dots, s_{\beta-1}) \in S_{\alpha+1} \times  \dots \times S_{\beta-1} \text{~s.t.~} (s_\alpha, s_{\alpha+1}, \dots, s_\beta) \text{~is compatible} \} .
\end{align}
In words, the set $F_{\alpha\to \beta}(s_\alpha)$ contains all states at time $\beta$ that the environment may reach when starting from state $s_\alpha$ at time $\alpha$. In particular, we have $F_{t\to t}(s_t) = \{s_t\}$ and $F_{t \to t+1}(s_t) = F_t(s_t)$.

\medskip
\noindent \underline{Multi-time-period action constraints.} For any two time periods $\alpha, \beta \in [T]$ with $\alpha \leq \beta$, and for two states $s_\alpha \in S_\alpha, s_\beta \in S_\beta$ such that $s_\beta \in F_{\alpha\to \beta}(s_\alpha)$, we define:
\begin{align} \label{eq:multi-time-action-constraint-U}
U_{\alpha\to \beta}(s_\alpha, s_\beta) \defeq \max_{(s_\alpha, s_{\alpha+1}, \dots, s_\beta)  \atop \text{~compatible}} \left\{\sum_{t = \alpha}^{\beta-1} U_t(s_t) \right\} .
\end{align}
In words, $U_{\alpha\to \beta}(s_\alpha, s_\beta)$ is the minimum increase to the inventory level from time $\alpha$ to time $\beta$ that a policy may achieve, irrespective of how the environment state evolves, provided that the initial environment state $s_\alpha$ and the final environment state $s_\beta$. The $\max$ operator in Eq.~\eqref{eq:multi-time-action-constraint-U} reflects the adversarial (minimax) nature of the environment evolution. In particular, we have $U_{t\to t}(\cdot, \cdot) = 0$ and $U_{t\to t+1}(s_t, s_{t+1}) = U_t(s_t)$ if $s_{t+1} \in F_t(s_t)$. Similarly, for $s_\beta \in F_{\alpha\to \beta}(s_\alpha)$, we also define 
\begin{align} \label{eq:multi-time-action-constraint-V}
V_{\alpha\to \beta}(s_\alpha, s_\beta) \defeq \min_{(s_\alpha, s_{\alpha+1}, \dots, s_\beta)  \atop \text{~compatible}} \left\{\sum_{t = \alpha}^{\beta-1} V_t(s_t) \right\} 
\end{align}
to be the maximum increase to the inventory level from time $\alpha$ to time $\beta$ that a policy may achieve, regardless of how the environment evolves, provided that the initial environment state $s_\alpha$ and the final environment state $s_\beta$.

\subsection{Future-imposed Inventory Level Bounds} \label{sec:future-imposed-inventory-level-constraints}

In this subsection, we present the key lemmas that will be used in our algorithm to solve the minimax-MDPs. These lemmas establish bounds on the inventory level at any given state (at any time) of the minimax-MDP to ensure the existence of a feasible policy starting from that state. These bounds are derived from the direct inventory level constraints imposed on states in future time periods. They naturally contribute to a class of necessary conditions for the existence of feasible policies in a minimax-MDP, which we will introduce in the next subsection. 

For any two time periods $\alpha, \beta \in [T]$ with $\alpha \leq \beta$ and any state $s_\alpha \in S_\alpha$, we define:
\begin{align}\label{eq:def-R-beta-alpha}
R_{\beta \leadsto \alpha}(s_\alpha) & \defeq \min_{s_\beta \in F_{\alpha\to \beta}(s_\alpha)} \left\{ R_\beta(s_\beta) - U_{\alpha\to \beta}(s_\alpha, s_\beta)\right\},\\
L_{\beta \leadsto \alpha}(s_\alpha) & \defeq \max_{s_\beta \in F_{\alpha\to \beta}(s_\alpha)} \left\{ L_\beta(s_\beta) - V_{\alpha\to \beta}(s_\alpha, s_\beta)\right\} . \label{eq:def-L-beta-alpha}
\end{align}
The intuition behind these definitions is as follows: if we start from environment state $s_\alpha$ with inventory level $x_\alpha$ at time $\alpha$, then for any $s_\beta \in F_{\alpha \to \beta}(s_\alpha)$, if we reach environment state $s_\beta$ at time $\beta$, the worst-case (due to the adversarial environment evolution) inventory level increase is at least $U_{\alpha\to \beta}(s_\alpha, s_\beta)$. Therefore, $x_\beta \geq x_\alpha + U_{\alpha\to \beta}(s_\alpha, s_\beta)$. In order to satisfy the environment-dependent inventory constraint at time $\beta$, we need to have
$x_\alpha + U_{\alpha\to \beta}(s_\alpha, s_\beta) \leq x_\beta \leq R_\beta(s_\beta)$, i.e., 
\[
x_\alpha \leq R_\beta(s_\beta) - U_{\alpha\to \beta}(s_\alpha, s_\beta).
\]
Therefore, $R_{\beta \leadsto \alpha}(s_\alpha)$ provides an upper bound on the inventory level at state $s_\alpha$ and time $\alpha$ implied by the constraints at time $\beta$. Similarly,  $L_{\beta \leadsto \alpha}(s_\alpha)$ provides a lower bound for the  inventory level.

We further define
\begin{align}\label{eq:def-R-star}
R_{\star\leadsto \alpha}(s_\alpha) \defeq \min_{\beta \geq \alpha} \left\{ R_{\beta \leadsto \alpha}(s_\alpha) \right\}, \text{~~~~and~~~~} L_{\star\leadsto \alpha}(s_\alpha) \defeq \max_{\beta \geq \alpha} \left\{ L_{\beta \leadsto \alpha}(s_\alpha) \right\},
\end{align}
which incorporates upper and lower bounds implied by the inventory level constraints from all future time periods of $\alpha$.  We formalize the intuition discussed above in the following lemma, the proof of which is deferred to Section~\ref{sec:proof-of-lemma-future-imposed-upper-bound}.

\begin{lemma}[Future-imposed inventory bounds]
\label{lem:future-imposed-upper-bound}
Given a minimax-MDP instance $\mathcal{I}$, if there exists a feasible policy $\pi$, then for any $t \in \{1,2,\dots,T\}$ and $\pi$-compatible (partial) trajectory $(s_1, x_1, s_2, x_2, \dots, s_t, x_t)$, we have that
$L_{\star \leadsto t}(s_t) \leq x_t \leq R_{\star \leadsto t}(s_t) $.
\end{lemma}

To facilitate the understanding of the future-imposed inventory bounds, we present the following observations, where Observation~\ref{obs:R-star-simple} is straightforward by definition and we omit its proof, and Observation~\ref{obs:R-star-neighbors} provides a relationship between the future-imposed inventory upper bounds for two temporally consecutive states:
\begin{observation} \label{obs:R-star-simple}
For any time period $\alpha \in [T]$ and any state $s_\alpha \in S_\alpha$, we have $
R_{\alpha \leadsto \alpha}(s_\alpha) = R_{\alpha}(s_\alpha)$ and $L_{\alpha \leadsto \alpha}(s_\alpha) = L_{\alpha}(s_\alpha)$, and therefore, 
$R_{\star \leadsto \alpha}(s_\alpha) \leq R_{\alpha}(s_\alpha)$ and $L_{\star \leadsto \alpha}(s_\alpha) \geq L_{\alpha}(s_\alpha)$.
\end{observation}

\begin{observation} \label{obs:R-star-neighbors}
For any time period $\alpha \in [T - 1]$ and any states $s_\alpha \in S_\alpha, s_{\alpha + 1} \in F_{\alpha}(s_\alpha)$, we have 
$
R_{\star \leadsto \alpha + 1}(s_{\alpha+1})  - R_{\star \leadsto \alpha}(s_\alpha)\geq U_{\alpha}(s_{\alpha})$.
\end{observation}
\proof{Proof.}
Let $\beta \geq \alpha + 1$ and $s_\beta \in F_{\alpha + 1 \to \beta}(s_{\alpha + 1})$ be such that $R_{\star \leadsto \alpha + 1}(s_{\alpha+1}) = R_\beta(s_\beta) - U_{\alpha + 1 \to \beta}(s_{\alpha+1}, s_{\beta})$ (because of Eq.~\eqref{eq:def-R-beta-alpha} and Eq.~\eqref{eq:def-R-star}). We have
\[
R_{\star \leadsto \alpha}(s_\alpha) \leq R_{\beta \leadsto \alpha}(s_{\alpha}) \leq R_\beta(s_\beta) - U_{\alpha \to \beta}(s_\alpha, s_\beta),
\]
where the first inequality is due to Eq.~\eqref{eq:def-R-star} and the second inequality is because of Eq.~\eqref{eq:def-R-beta-alpha} and $s_{\alpha + 1} \in F_\alpha(s_\alpha)$ (so that $s_\beta \in F_{\alpha\to \beta}(s_\alpha)$).
In all, we have
\[
R_{\star \leadsto \alpha + 1}(s_{\alpha+1})  - R_{\star \leadsto \alpha}(s_\alpha)\geq U_{\alpha \to \beta}(s_\alpha, s_\beta) - U_{\alpha + 1 \to \beta}(s_{\alpha+1}, s_\beta) \geq U_{\alpha}(s_{\alpha}) ,
\]
where the second inequality is due to the definition of $U_{\cdot \to \cdot}(\cdot, \cdot)$ in Eq.~\eqref{eq:multi-time-action-constraint-U}.
\hfill\Halmos
\endproof

For each $\alpha \in \{1, 2, \dots, T-1\}$ and $s_\alpha \in S_\alpha$, we also consider the collection of future-imposed inventory bounds for all states in $F_\alpha(s_\alpha)$. Formally, we define
\begin{align} \label{eq:def-R-star-bracket}
R_{\star\leadsto \alpha \to [\alpha+1]}(s_\alpha) \defeq \min_{s_{\alpha+1} \in F_{\alpha}(s_\alpha)} \left\{R_{\star \leadsto \alpha+1}(s_{\alpha+1})\right\}, ~~ L_{\star\leadsto \alpha \to [\alpha+1]}(s_\alpha) \defeq \max_{s_{\alpha+1} \in F_{\alpha}(s_\alpha)} \left\{L_{\star \leadsto \alpha+1}(s_{\alpha+1})\right\}.
\end{align}
By definition, we have that $R_{\star\leadsto \alpha \to [\alpha+1]}(s_\alpha) \leq R_{\star\leadsto \alpha+1}(s_{\alpha+1})$ and $L_{\star\leadsto \alpha \to [\alpha+1]}(s_\alpha) \geq L_{\star\leadsto \alpha+1}(s_{\alpha+1})$ hold for any $s_{\alpha+1} \in F_\alpha(s_\alpha)$.
In Section~\ref{sec:proof-of-lemma-future-imposed-upper-bound-strengthened}, we prove the following lemma, which, whenever applicable, strengthens the upper bound in Lemma~\ref{lem:future-imposed-upper-bound}.
\begin{lemma}[Strengthened future-imposed inventory bounds]\label{lem:future-imposed-upper-bound-strengthened}
Given a minimax-MDP instance $\mathcal{I}$, if there exists a feasible policy $\pi$, then for any $t \in \{2,3,\dots,T\}$ and $\pi$-compatible (partial) trajectory $(s_1, x_1, s_2, x_2, \dots, s_t, x_t)$, we have that
$
L_{\star\leadsto t-1 \to [t]}(s_{t-1}) \leq x_t \leq R_{\star\leadsto t-1 \to [t]}(s_{t-1}) $.
\end{lemma}

\subsection{Future-imposed Conditions and the Main Equivalence Theorem}

In this subsection, we introduce a class of conditions that a minimax-MDP must satisfy for a feasible policy to exist. These conditions check time period and the corresponding environment states, leveraging the key relationships in Lemma~\ref{lem:future-imposed-upper-bound} and Lemma~\ref{lem:future-imposed-upper-bound-strengthened}. According to these lemmas, it is straightforward to see that these conditions are necessary for the existence of a feasible policy. On the other hand, surprisingly, we present our main theorem stating that the future-imposed conditions are also sufficient, implying that verifying these conditions alone is enough to determine the existence of a feasible policy.

\begin{definition}[Future-imposed conditions]\label{def:two-point-condition}
For each time period $\alpha \in \{2, 3, \dots, T\}$, the \emph{future-imposed condition} requires that for each $s_{\alpha-1} \in F_{1 \to \alpha-1}(s_1)$, it holds that
\begin{align}  \label{eq:two-point-condition}
R_{\star \leadsto \alpha -1 \to [\alpha]}(s_{\alpha-1})   \geq L_{\star \leadsto \alpha -1 \to [\alpha]}(s_{\alpha-1}). 
\end{align}
For $\alpha = 1$, the future-imposed condition requires that
\begin{align}\label{eq:two-point-condition-special-case}
R_{\star \leadsto 1}(s_1) \geq L_{\star \leadsto 1}(s_1).
\end{align}
\end{definition}

\begin{lemma} \label{lemma:two-point-condition-necessary}
The future-imposed conditions are necessary for a minimax-MDP to admit a feasible policy. 
\end{lemma}
\proof{Proof.} This is a direct corollary of Lemma~\ref{lem:future-imposed-upper-bound} and Lemma~\ref{lem:future-imposed-upper-bound-strengthened}. Suppose that $\pi$ is a feasible policy for the minimax-MDP. When $\alpha > 1$, because of $s_{\alpha-1} \in F_{1\to \alpha-1}(s_1)$, $s_\alpha \in F_{\alpha-1}(s_{\alpha-1})$, and Observation~\ref{obs:compatible-trajectory-environment-state-trajectory}.\ref{obs:compatible-trajectory-environment-state-trajectory-b}, there exists a $\pi$-compatible partial trajectory $\phi = (s_1, x_1, s_2, x_2, \dots, s_\alpha, x_\alpha)$. Applying the two lemmas to $\phi$, we have that $R_{\star \leadsto \alpha -1 \to [\alpha]}(s_{\alpha-1})   \geq x_\alpha \geq  L_{\star \leadsto \alpha -1 \to [\alpha]}(s_{\alpha-1})$, implying Eq.~\eqref{eq:two-point-condition}. For $\alpha = 1$, Lemma~\ref{lem:future-imposed-upper-bound} directly implies Eq.~\eqref{eq:two-point-condition-special-case}.
\hfill\Halmos
\endproof

Next, we present the main theorem of our paper.
\begin{theorem}[Equivalent conditions.] \label{thm:main-equivalent-conditions}
A minimax-MDP admits a feasible policy $\pi$ if and only if it satisfies the future-imposed conditions.
\end{theorem}
\proof{Proof.} 
Thanks to Lemma~\ref{lemma:two-point-condition-necessary}, we only need to establish the sufficiency of the future-imposed conditions: when a minimax-MDP $\mathcal{I}$ satisfies the future-imposed conditions, a feasible policy $\pi$ exists. We will explicitly construct such a feasible policy (Eq.~\eqref{eq:pi-alpha}) and prove its feasibility by induction on the horizon length.

For convenience, we first introduce a few notations. For each $\alpha \in [T]$, $s_\alpha \in S_\alpha$, and $x_\alpha \in \mathbb{R}$, let 
\[
\mathcal{I}_{\alpha, s_\alpha, x_\alpha} = (T - \alpha + 1, \{S_t\}_{t=\alpha}^{T}, s_\alpha, x_\alpha, \{F_t\}_{t=\alpha}^{T-1}, \{U_t, V_t\}_{t=\alpha}^{T-1}, \{L_t, R_t\}_{t=\alpha}^{T}\})
\]
be the $(T-\alpha+1)$-horizon minimax-MDP that starts from the system state $(s_\alpha, x_\alpha)$. Here we slightly extend the definition of minimax-MDPs so that the initial time step is $t = \alpha$.

For each $\beta \in [T]$ and $s_\beta \in S_\beta \cap F_{1 \to \beta}(s_1)$, we define
\begin{align} \label{eq:def-tilde-L-beta}
\tilde{L}_\beta(s_\beta) \defeq \min \left\{
\begin{array}{cc}
\displaystyle{\min_{\substack{1<\alpha \leq \beta: s_{\alpha-1} \in F_{1 \to (\alpha-1)}(s_1)\\ \wedge s_\alpha\in F_{\alpha-1}(s_{\alpha-1}) \wedge s_\beta \in F_{\alpha\to\beta}(s_\alpha)}} \left\{R_{\star \leadsto \alpha-1 \to [\alpha]}(s_{\alpha-1}) + V_{\alpha \to \beta}(s_\alpha, s_\beta)\right\} } \\
\displaystyle{R_{\star \leadsto 1}(s_1) + V_{1 \to \beta}(s_1, s_\beta) }
\end{array} \right\} .
\end{align}
The intuition behind the above definition is as follows. If we expand the right-hand sides of the future-imposed conditions (both Eq.~\eqref{eq:two-point-condition} and Eq.~\eqref{eq:two-point-condition-special-case}) by the definitions of $L_{\star \leadsto \alpha -1 \to [\alpha]}(\cdot)$ and $L_{\star \leadsto \alpha}(\cdot)$, we obtain a collection of inequalities that upper bound $L_\beta(s_\beta)$ for each $\beta \geq \alpha$ and corresponding environment state $s_\beta$. We aggregate these upper bounds and define $\tilde{L}_\beta(s_\beta)$ as their minimum. In other words, $\tilde{L}_\beta(\cdot)$ serves as an upper bound on $L_\beta(\cdot)$, derived from the future-imposed conditions through algebraic manipulation. Formally, we have the following claim, whose proof is deferred to Section~\ref{sec:proof-of-tilde-L-larger-L}.
\begin{claim}\label{claim:tilde-L-larger-L}
    Assuming the future-imposed conditions, for any $\beta\in[T]$ and $s_\beta\in F_{1\to \beta}(s_1)$, we have 
    $
    \tilde{L}_\beta(s_\beta)\geq L_{\beta}(s_\beta)$.
\end{claim}

We also defer the proof of the following property of $\tilde{L}$ to Section~\ref{sec:proof-of-claim-tilde-L-beta-property}.
\begin{claim}\label{claim:tilde-L-beta-property}
For any $\beta \in [T-1]$, $s_\beta \in S_\beta \cap F_{1 \to \beta}(s_1)$, and $s_{\beta+1} \in F_{\beta}(s_\beta)$, we have
$\tilde{L}_{\beta + 1}(s_{\beta+1}) \leq \tilde{L}_{\beta}(s_\beta) + V_\beta(s_\beta)$.
\end{claim}

To establish the sufficiency of the future-imposed conditions, we prove the following proposition.
\begin{proposition}\label{prop:main-thm-sufficiency}
Assuming the future-imposed conditions, for each $\alpha \in [T]$, $s_\alpha \in S_\alpha$, and $x_\alpha \in \mathbb{R}$ such that $s_\alpha \in F_{1\to\alpha}(s_1)$ and $x_\alpha \in [\tilde{L}_\alpha(s_\alpha), R_{\star \to \alpha}(s_\alpha)]$, we have that $\mathcal{I}_{\alpha, s_\alpha, x_\alpha}$ admits a feasible policy.
\end{proposition}
By the future-imposed conditions and Observation~\ref{obs:R-star-simple}, we have $0 = R_1(s_1) \geq R_{\star \leadsto 1}(s_1)  \geq L_{\star \leadsto 1}(s_1) \geq L_1(s_1) = 0$. Therefore, $R_{\star \leadsto 1}(s_1) = L_{\star \leadsto 1}(s_1) = 0$. Furthermore, $\tilde{L}_1(s_1) = R_{\star \leadsto 1}(s_1) + V_{1 \to 1}(s_1, s_1) \leq 0 + 0 = 0$.  In all, we have that $x_1 = 0 \in [\tilde{L}_1(s_1), R_{\star \leadsto 1}(s_1)]$. Thus, once we establish Proposition~\ref{prop:main-thm-sufficiency}, the future-imposed conditions imply that a feasible policy exists for $\mathcal{I} = \mathcal{I}_{1, s_1, x_1}$.

It remains to prove Proposition~\ref{prop:main-thm-sufficiency}, and we prove the statement by induction on $\alpha$.

\medskip
\noindent\underline{Induction basis.} When $\alpha = T$, there is only one trivial policy $\pi = \emptyset$ (because no decision needs to be made). This policy is feasible because $x_T \in [\tilde{L}_T(s_T), R_{\star \to T}(s_T)] \subseteq [L_T(s_T), R_T(s_T)]$, where the left interval endpoint part of the ``$\subseteq$'' relation is because of Claim~\ref{claim:tilde-L-larger-L} and the right endpoint part is due to Observation~\ref{obs:R-star-simple}.

\medskip
\noindent\underline{Induction step.} Now we assume that Proposition~\ref{prop:main-thm-sufficiency} holds for all $\alpha > \alpha_0$ (where $\alpha_0 \in \{1, 2, \dots, T - 1\})$. For $\alpha = \alpha_0$, we first verify that the environment-dependent inventory constraints are satisfied at time $\alpha$, i.e., $x_\alpha \in [\tilde{L}_\alpha(s_\alpha), R_{\star \to \alpha}(s_\alpha)] \subseteq [L_\alpha(s_\alpha), R_\alpha(s_\alpha)]$, which can be done similarly as in the induction basis part.

We then consider the following policy at time $\alpha$:
\begin{align} \label{eq:pi-alpha}
\pi_\alpha(s_\alpha, x_\alpha) = \min\left\{V_\alpha(s_\alpha), R_{\star \leadsto \alpha \to [\alpha + 1]}(s_{\alpha}) - x_\alpha\right\}.
\end{align}
The intuition about the above policy construction is that the policy greedily orders as much as possible subject to two known constraints: $\pi_\alpha(s_\alpha, x_\alpha) \leq V_\alpha(s_\alpha)$ and $x_{\alpha + 1} = \pi_\alpha(s_\alpha, x_\alpha) + x_\alpha \leq R_{\star \leadsto \alpha \to [\alpha + 1]}(s_{\alpha})$ (where the second constraint is due to Lemma~\ref{lem:future-imposed-upper-bound-strengthened}). We also note that a symmetric policy that orders as little as possible subject to the mirrored constraints should also work.

To establish the feasibility of the constructed policy, we first show that $\pi_\alpha$ satisfies the environment-dependent action constraints (Eq.~\eqref{eq:environment-dependent-action-constraint}). The upper bound is straightforward, and we only need to prove that $\pi_\alpha(x_\alpha, s_\alpha) \geq U_\alpha(s_\alpha)$. Let $s_{\alpha+1}^* \in \arg\min_{s_{\alpha+1} \in F_\alpha(s_\alpha)} \{R_{\star \leadsto \alpha+1}(s_{\alpha+1})\}$, we have 
\[
R_{\star \leadsto \alpha \to [\alpha + 1]}(s_{\alpha}) - x_\alpha = R_{\star \leadsto \alpha+1}(s_{\alpha+1}^*) - x_\alpha \geq R_{\star \leadsto \alpha+1}(s_{\alpha+1}^*) - R_{\star \leadsto \alpha}(s_\alpha) \geq U_\alpha(s_\alpha) ,
\]
where the equality is by the definition in Eq.~\eqref{eq:def-R-star-bracket}, the first inequality is due to the proposition assumption and the last inequality is because of Observation~\ref{obs:R-star-neighbors}. Combining with the condition $V_{\alpha}(s_\alpha)\geq U_\alpha(s_\alpha)$,
we deduce that $\pi_\alpha$ satisfies the environment-dependent action constraints.

After applying $\pi_\alpha$ at time period $\alpha$, for any $s_{\alpha+1} \in F_\alpha(s_\alpha)$, the system state may transit to $(s_{\alpha+1}, x_{\alpha+1})$, where $x_{\alpha+1}= x_\alpha + \pi_\alpha(s_\alpha, x_\alpha)$. Then, the minimax-MDP reduces to  $\mathcal{I}_{\alpha+1, s_{\alpha+1}, x_{\alpha+1}}$. We have that $s_{\alpha+1} \in F_{1\to\alpha+1}(s_1)$ (because $s_\alpha \in F_{1\to\alpha}(s_1)$ and $s_{\alpha+1} \in F_\alpha(s_\alpha)$). Therefore, we only need to verify that $x_{\alpha+1} \in [\tilde{L}_{\alpha+1}(s_{\alpha+1}), R_{\star \leadsto \alpha+1}(s_{\alpha+1})]$ so that we can apply the inductive hypothesis which implies that  $\mathcal{I}_{\alpha+1, s_{\alpha+1}, x_{\alpha+1}}$ admits a feasible policy. Then, together with $\pi_\alpha$, we can construct a feasible policy for $\mathcal{I}_{\alpha, s_\alpha, x_\alpha}$.

Finally, we verify that $x_{\alpha+1} \in [\tilde{L}_{\alpha+1}(s_{\alpha+1}), R_{\star \leadsto \alpha+1}(s_{\alpha+1})]$. For the upper bound, we have
\begin{align} 
x_{\alpha+1} = x_\alpha + \pi_\alpha(s_\alpha, x_\alpha)& \leq x_\alpha + R_{\star \leadsto \alpha \to [\alpha + 1]}(s_{\alpha})   - x_\alpha = \min_{s' \in F_\alpha(s_\alpha)} \{R_{\star \leadsto \alpha+1}(s')\} \leq R_{\star \leadsto \alpha+1}(s_{\alpha+1}) . \nonumber
\end{align}

For the lower bound, we discuss the following two cases:

\noindent \underline{Case 1:} $V_\alpha(s_\alpha)$ is the greater one in Eq.~\eqref{eq:pi-alpha}. In this case, we have
\begin{align*}
x_{\alpha+1} & = x_\alpha + R_{\star \leadsto \alpha \to [\alpha + 1]}(s_{\alpha})   - x_\alpha 
 = R_{\star \leadsto \alpha \to [\alpha + 1]}(s_{\alpha}) + V_{\alpha+1 \to \alpha+1}(s_{\alpha+1}, s_{\alpha+1}) \geq \tilde{L}_{\alpha+1}(s_{\alpha+1}),  
\end{align*}
where the last inequality is by the definition of $\tilde{L}_{\alpha+1}(s_{\alpha+1})$ (Eq.~\eqref{eq:def-tilde-L-beta}).


\noindent \underline{Case 2:} $V_\alpha(s_\alpha)$ is the smaller one (or equally small) in Eq.~\eqref{eq:pi-alpha}. In this case, we have 
\[
x_{\alpha+1} = x_\alpha + V_\alpha(s_\alpha) \geq \tilde{L}_\alpha(s_\alpha) + V_\alpha(s_\alpha) \geq \tilde{L}_{\alpha+1}(s_{\alpha+1}),
\]
where the last inequality is because of Claim~\ref{claim:tilde-L-beta-property}.
\hfill\Halmos
\endproof

A direct corollary of the proof of Theorem~\ref{thm:main-equivalent-conditions} is as follows.
\begin{corollary}[Feasible policy] \label{cor:feasible-policy}
Suppose the future-imposed conditions hold for a minimax-MDP. Let $\pi = (\pi_1, \pi_2, \dots, \pi_{T-1})$ be 
$\pi_t(s_t, x_t) = \min\left\{V_t(s_t), R_{\star \leadsto t \to [t+1]}(s_t) - x_t\right\}$ (for each $t \in \{1, 2, \dots, T-1\}$).
Then $\pi$ is a feasible policy.
\end{corollary}

Given a minimax-MDP with a finite state space, one may use dynamic programming to calculate $R_{\star \leadsto \alpha}(\cdot)$, $L_{\star \leadsto \alpha}(\cdot)$, $R_{\star \leadsto [\alpha - 1] \to [\alpha]}(\cdot)$, and $L_{\star \leadsto [\alpha - 1] \to [\alpha]}(\cdot)$ in $\mathrm{poly}(S, T)$ time (where $S$ denotes the number of possible states). Thus, according to Theorem~\ref{thm:main-equivalent-conditions} and Corollary~\ref{cor:feasible-policy}, one may also verify the feasibility of the minimax-MDP and compute a feasible policy (if there is one) in polynomial time. In the following sections, we demonstrate that in many applications, the future-imposed conditions can lead to more efficient (often closed-form) solutions, even when the state space is infinite.

\section{Application I: Robust Multi-period Ordering with Predictions}
\label{sec:app-I-robust-multi-period-ordering}

In this section, we apply our future-imposed conditions under the minimax-MDP framework to a variation of the newsvendor problem that involves multiple preparation days followed by a final operating day. On each \emph{preparation day}, the decision-maker can place orders in anticipation of demand on the final \emph{operating day}. The goal is to maximize profit by balancing the overage cost (incurred when total orders exceed demand) and the underage cost (incurred when demand exceeds total orders). Each preparation day, the decision-maker receives a prediction interval for the final demand, typically generated by a machine learning algorithm. These intervals become progressively narrower as the operating day approaches, reflecting increasingly accurate predictions. The lengths of these prediction intervals, which are usually determined by the machine learning model's design, are assumed to be known to the decision-maker. However, no additional structural assumptions are made about the intervals, as if they were selected by an adversary. The \emph{robust multi-period ordering with predictions (RMOwP)} problem aims to develop a computationally efficient online algorithm that remains competitive with the optimal hindsight ordering decisions, under the worst-case (adversarial) sequence of prediction intervals. This problem is relevant to practical scenarios such as staff planning, where workforce demand remains uncertain until the operating day. On each preparation day, the decision-maker recruits labor with the help of prediction intervals, striving for near-optimal decisions across all possible prediction interval sequences.

We consider two metrics to measure the competitiveness of the online decision algorithm. The first is \emph{regret}, defined as the difference between the profit achieved by the optimal hindsight decision and that of the online algorithm. The second metric is the \emph{competitive ratio}, which is the ratio of the profit obtained by the online algorithm to that of the optimal hindsight decision. In Section~\ref{sec:app-I-robust-regret} and Section~\ref{sec:app-I-robust-competitive-ratio}, we apply our minimax-MDP framework and future-imposed conditions to address both metrics respectively. 

Finally, we may extend our minimax-MDP framework to a more general setting, namely the \emph{scale-dependent prediction sequences}, where the upper bounds on the length of prediction intervals may depend on the predicted demand scale and be non-uniform within a single preparation day. This extension is particularly relevant in practical scenarios, where larger-scale quantities are inherently more difficult to predict accurately within fixed-length intervals. The scale-dependent formulation allows the model to capture prediction interval lengths that correlate with the magnitude of the predicted values. We defer the detailed discussion and derivation of this extension to Section~\ref{sec:app-I-non-uniform-interval-lengths}.

\subsection{Problem Formulation}
\label{sec:app-I-problem-formulation}
The problem consists of $T+1$ days. On each of the first $T$ days, the decision-maker may place orders to meet the demand that will be realized on day $T+1$. The demand $d$ is not accessible until day $T+1$. However, the retailer has access to a forecast $\mathcal{P}_t$ for $d$ on each day $t \in [T]$, which is represented as an interval: $d \in \mathcal{P}_t  = [\unld_t,\ovld_t]$. Similar to \citet{feng2024robust}, we make the following regularity assumption about the sequence of prediction intervals $\mathcal{P} = \{\mathcal{P}_t\}_{t \in [T]}$ (\citet{feng2024robust} slightly generalized the consistency property to $\epsilon$-consistency, i.e., $d \in [\unld_t - \epsilon, \ovld_t +\epsilon]$; our algorithm may also be adapted to this generalization straightforwardly):
\begin{assumption}[Regularity of prediction sequences]
\label{asp:regularityofpredictionsequence}
Given an interval $[\unld_0,\ovld_0]$ and a sequence $\{\triangle_{t}\}_{t\in[T]}$, we call an interval sequence $\mathcal{P}=\{\mathcal{P}_t\}_{t\in[T]}$ a \emph{regular prediction sequence} under $\{[\unld_0,\ovld_0],\{\triangle_{t}\}_{t\in[T]}\}$ if for each day $t\in [T]$, the prediction interval $\mathcal{P}_{t}=[\unld_t,\ovld_t]$ satisfies:
\begin{itemize}
\item \underline{Consistency}: the unknown demand $d$ is in the interval $[\unld_{t},\ovld_{t}]$ and $[\unld_{t},\ovld_{t}]\subseteq [\unld_{t-1},\ovld_{t-1}]$. 
\item \underline{Bounded error}: the prediction error is bounded, i.e., $\ovld_{t}-\unld_{t}\leq \triangle_{t}$ for $\triangle_{t}\geq 0$. $\{\triangle_{t}\}_{t\in [T]}$ is called the sequence of ``prediction error upper bounds'', and is known to the decision-maker.  
\end{itemize}
\end{assumption}

Without loss of generality, we further assume that $\{\triangle_{t}\}_{t\in [T]}$ is a non-decreasing and non-negative sequence and  $\triangle_{1}\leq \ovld_0-\unld_0$. We formally define the problem below.

\begin{definition}\label{def:definition-of-robust-multi-period}
An instance of the robust multi-period ordering with predictions (RMOwP) problem is defined as a tuple:
\begin{align}\label{eq:definition-of-robust-multi-period}
\mathcal{J}\defeq\{T,[\unld_{0},\ovld_{0}],\{\triangle_{t}\}_{t\in [T]},c,p,\{V_{t}\}_{t\in [T]}\} ,
\end{align}
where $T$, $\unld_{0}$, $\ovld_{0}$, $\{\triangle_{t}\}_{t\in [T]}$ are as previously defined. Additionally, we define the following parameters:
\begin{enumerate}
\item \emph{Ordering cost} ($c$): the cost per unit for placing an order on the preparation days. 
\item \emph{Unit revenue} ($p$): the revenue generated per unit of demand met on the operating day ($T+1)$. W.l.o.g., we assume $p>c$, as otherwise, the optimal policy would be to place no orders.
\item \emph{Supply capacity} ($V_{t}, t\in [T])$): the maximum allowable order quantity on day $t$, imposing an upper bound constraint on the ordering amount $a_t$: 
$    0\leq a_t\leq V_t$.
\end{enumerate}
\end{definition}

Given an instance $\mathcal{J}$, a feasible policy $\pi=(\pi_{1}, \pi_2, \dots,\pi_{T})$ maps the prediction interval and the inventory level at any time $t \in \{1, 2, \dots, T\}$ to an ordering decision:
\begin{align}
    \pi_{t}: \{[a,b]: 0 \leq b - a \leq \triangle_t\} \times \mathbb{R} \to [0,V_{t}], \qquad \forall  t \in \{1, 2, \dots, T\}.
\end{align}

\medskip
\noindent\uline{{Timeline.}} We formalize the timeline of the problem as follows.
\begin{itemize}
\item On day $0$, the problem parameters are revealed to the decision-maker. The initial inventory level is $x_1 = 0$.
\item On each day $t\in \{1, 2, \dots, T\}$:
\begin{itemize}
\item the prediction $\mathcal{P}_{t}=[\unld_{t},\ovld_{t}]$ is revealed to the decision-maker;
\item the decision-maker orders quantity $a_{t}=\pi_{t}([\unld_t,\ovld_t],x_{t})$ and updates the inventory level:
\begin{align}
    x_{t+1}=x_t+a_t=x_t+\pi_{t}([\unld_t,\ovld_t],x_{t}) .
\end{align}
\end{itemize} 
\item On day $T+1$: the demand $d$ is revealed, and the profit is
\begin{align}\label{eq:profit-of-policy}
\mathcal{R}_{\pi}(\mathcal{P}, d; \mathcal{J}) \defeq \text{(revenue)} -\text{(total ordering cost)}=p\cdot\min(d,x_{T+1})-c\cdot x_{T+1} .
\end{align} 
\end{itemize}

Let $\pi^\sharp$ be the hindsight optimal policy with the knowledge of $d$. It is straightforward to verify 
\begin{align}\label{eq:profit-offline-optimal}
\mathcal{R}^\sharp (\mathcal{P}, d; \mathcal{J}) \defeq \mathcal{R}_{\pi^\sharp}(\mathcal{P}, d; \mathcal{J})=(p-c)\cdot \min\left\{\sum_{t=1}^{T} V_t, d\right\}.
\end{align}

\medskip
\noindent\uline{{Competitiveness metrics.}} Given $\mathcal{P}$ and $d$, we respectively define the \emph{regret} and the \emph{competitive ratio} of a policy $\pi$ as
\begin{align} 
\mathrm{Reg}_\pi (\mathcal{P}, d; \mathcal{J}) \defeq \mathcal{R}^\sharp (\mathcal{P}, d; \mathcal{J}) - \mathcal{R}_\pi (\mathcal{P}, d; \mathcal{J}),  \text{~~~~and~~~~} \varphi_\pi (\mathcal{P}, d; \mathcal{J}) \defeq \frac{\mathcal{R}_\pi (\mathcal{P}, d; \mathcal{J})}{ \mathcal{R}^\sharp (\mathcal{P}, d; \mathcal{J})}. 
\end{align}
The robust (worst-case) regret and competitive ratio of $\pi$ are respectively defined as
\begin{align} \label{eq:app-I-competitive-metrics}
\mathrm{Reg}_\pi(\mathcal{J}) \defeq \sup_{\mathcal{P}, d} \{\mathrm{Reg}_\pi (\mathcal{P}, d; \mathcal{J})\},  \text{~~~~and~~~~} \varphi_\pi(\mathcal{J}) \defeq \inf_{\mathcal{P}, d} \{\varphi_\pi (\mathcal{P}, d; \mathcal{J})\},
\end{align}
where the supremum and the infimum are taken over all $(\mathcal{P}, d)$ pairs such that $\mathcal{P}$ is a regular prediction sequence with respect to $\{[\unld_0,\ovld_0], \{\triangle_t\}_{t\in[T]}\}$, and $d \in [\unld_T,\ovld_T]$.
In the following Section~\ref{sec:app-I-robust-regret} and Section~\ref{sec:app-I-robust-competitive-ratio}, we will apply our future-imposed conditions under the minimax-MDP framework to derive efficient algorithms that achieve the optimal robust regret and robust competitive ratio, respectively.

\subsection{Optimizing Robust Regret}
\label{sec:app-I-robust-regret}
In this section, we derive an online policy to achieve the optimal robust regret under any problem instance $\mathcal{J}$. We will first work with the decision problem of whether there exists an online policy achieving a robust regret at most a given threshold $\Gamma$, i.e., deciding whether
\begin{align}\label{eq:decision-problem-app-I-robust-regret}
\inf_{\pi} \{ \mathrm{Reg}_\pi (\mathcal{J}) \} \leq \Gamma.
\end{align}

We will assume w.l.o.g.~that $\ovld_0 \leq V = \sum_{t=1}^T V_t$. This is because once we obtain an optimal online policy $\pi^*$ under this assumption, we may construct a policy $\tilde{\pi}^*$ that simulates $\pi^*$ with $\tilde{\ovld}_0 = V$ and $\tilde{\mathcal{P}}_t = \mathcal{P}_t \cap (-\infty, V]$ as the prediction intervals, and show that $\tilde{\pi}^*$ is robust-regret-optimal even without the assumption.

\medskip
\noindent\uline{{The reduction.}} To apply the future-imposed conditions, we reduce the decision problem Eq.~\eqref{eq:decision-problem-app-I-robust-regret} to a minimax-MDP $\mathcal{I}$ with time horizon $T+2$. We label the time periods by $0, 1, 2, \dots, T+1$, corresponding to the days in the timeline of $\mathcal{J}$. We let the environment states in $\mathcal{I}$ be the prediction intervals. Specifically, we let
\begin{align*}
S_{t}=\{[a,b];~0\leq a\leq b\},~~~~ \forall 0\leq t\leq T,\qquad \text{and}\qquad  S_{T+1}=\{d;~d\geq 0\}.
\end{align*}
Note that on day $T+1$, the demand is realized and therefore the intervals collapse to a single real number, represented by our definition of $S_{T+1}$. Initially, in $\mathcal{I}$, we set
\begin{align*}
s_0 = [\unld_0,\ovld_0], \qquad \text{and}\qquad x_0 = 0.
\end{align*}
The environment state transition functions are set as follows:
\begin{align*}
F_{t}([a,b])&=\{[e,f];~0\leq f-e\leq\triangle_{t+1},~[e,f]\subseteq [a,b]\}, ~~~~ \forall t\in\{0,1,\dots,T-1\},~\\
F_{T}([a,b])&=\{d;~d\in[a,b]\},
\end{align*}
We then naturally define the environment-dependent action constraints (where $V_t$ is the supply capacity in $\mathcal{J}$):
\begin{align}
&U_{t}(\cdot)\equiv 0, ~~~~ \forall t\in\{0,1,\dots,T+1\}, \\
&V_{t}(\cdot)\equiv V_{t}, ~~~~ \forall t\in\{1,2,\dots,T\}, \qquad \text{and} \qquad V_{0}(\cdot)\equiv V_{T+1}(\cdot)\equiv 0. \label{eq:app-1-reduction-enviroment-dependent-action-constraint-V}
\end{align} 

The environment-dependent inventory constraints before time $T+1$ are trivially set as
\begin{align}
    \label{eq:left-side-from-0-to-T}
    L_0(\cdot) \equiv 0, ~~~~ R_0(\cdot) \equiv 0, ~~~~ L_t(\cdot)\equiv -\infty, ~~~~ R_t(\cdot)\equiv +\infty, \qquad t\in\{1,2, \dots,T\}.
\end{align}
We finally define the nontrivial constraints $L_{T+1}(\cdot)$ and $R_{T+1}(\cdot)$, which are related to the threshold $\Gamma$ in Eq.~\eqref{eq:decision-problem-app-I-robust-regret}:
\begin{align} \label{eq:app-I-robust-regret-L-R-T-plus-1}
L_{T+1}(d)=d-\frac{\Gamma}{p-c}, \qquad \text{and} \qquad R_{T+1}(d)=d+\frac{\Gamma}{c}, \qquad \forall d \geq 0.
\end{align}

We prove the following lemma about our reduction.
\begin{lemma}\label{lem:app-I-reduction}
Given $\mathcal{J}$, let the minimax-MDP $\mathcal{I}$ be constructed by our reduction. Then Eq.~\eqref{eq:decision-problem-app-I-robust-regret} holds if and only if $\mathcal{I}$ admits a feasible policy.
\end{lemma}
\proof{Proof.}
For any valid ordering policy $\pi^\mathcal{J}$ for $\mathcal{J}$, consider the mapping 
\[
\sigma : \pi^\mathcal{J} \mapsto \pi^\mathcal{I} = (\pi^\mathcal{I}_0(\cdot, \cdot) \equiv 0, \pi^\mathcal{I}_1 = \pi^\mathcal{J}_1, \pi^\mathcal{I}_2 = \pi^\mathcal{J}_2, \dots, \pi^\mathcal{I}_T = \pi^\mathcal{J}_T)
\]
such that $\pi^\mathcal{I} = \sigma(\pi^\mathcal{J})$ is policy for $\mathcal{I}$. Conversely, if $\pi^\mathcal{I}$ is a feasible policy for $\mathcal{I}$, one may verify the existence of $\sigma^{-1}(\pi^\mathcal{I})$.

To prove the lemma, we only need to show that $\pi^\mathcal{I} = \sigma(\pi^\mathcal{J})$ is a feasible policy for $\mathcal{I}$ if and only if $\pi^\mathcal{J}$ is a valid ordering policy with $\mathrm{Reg}_\pi(\mathcal{J}) \leq \Gamma$.  

For convenience, let $x_{T+1}^{\mathcal{I}}$ denote the inventory level at time period $T+1$ when running policy $\pi^\mathcal{I}$ in $\mathcal{I}$ (which can be naturally defined even if the policy does not meet the environment-dependent action constraints). Similarly let $x_{T+1}^{\mathcal{J}}$ be the total ordering amount before day $T+1$ under the ordering policy $\pi^\mathcal{J}$ in $\mathcal{J}$ (which can also be naturally defined even if the policy does not adhere to the supply capacity constraints). If $\pi^\mathcal{I} = \sigma(\pi^\mathcal{J})$, we directly have $x_{T+1}^{\mathcal{I}} = x_{T+1}^{\mathcal{J}}$, denoted by $x$. 

Note that the only non-trivial constraint in $\mathcal{I}$ is the environment-dependent inventory constraint at time period $T+1$. Hence, our proof further reduces to demonstrating that for any demand $d$, $L_{T+1}(d) \leq x \leq R_{T+1}(d)$ if and only if the regret of $\pi^{\mathcal{J}}$, given demand $d$ and total ordering amount $x$, does not exceed $\Gamma$, i.e.,
\[
x \in \left[d - \frac{\Gamma}{p-c}, d + \frac{\Gamma}{c}\right]  \Leftrightarrow (p-c) \cdot d - [p \cdot \min(d, x) - c \cdot x] \leq \Gamma,
\]
which can be easily verified by considering the two cases: $x \leq d$ and $x > d$.
\hfill\Halmos
\endproof

By Lemma~\ref{lem:app-I-reduction}, we invoke Theorem~\ref{thm:main-equivalent-conditions} with $\mathcal{I}$ to derive the following equivalent conditions of Eq.~\eqref{eq:decision-problem-app-I-robust-regret}, proof details of which will be provided in Section~\ref{sec:proofofAssessingbyDifferenceinwarmup}.
\begin{proposition}
\label{prop:conclusion-app-I-optimal-robust-regret}
Eq.~\eqref{eq:decision-problem-app-I-robust-regret} holds if and only if
\begin{align*}
\Gamma \geq \Gamma^{*} = \frac{c\cdot (p-c)}{p}\cdot \mathop{\max}\limits_{t\in [T]}\left\{\triangle_{t}-\sum\limits_{s=t+1}^{T}V_{s}\right\}.
\end{align*}
Moreover, the optimal ordering policy to achieve the robust regret $\Gamma^*$ is
\begin{align*}
\pi_{t}([\unld_t,\ovld_t],x_t)=\min\left\{V_t,\frac{\Gamma^*}{c}+\unld_t-x_t\right\}, \qquad \forall t \in \{1, 2, \dots, T\}.
\end{align*}
\end{proposition}

While the solution \cite{feng2024robust} to computing the optimal robust regret is based on solving a linear program of size $\Omega(T)$, our Proposition~\ref{prop:conclusion-app-I-optimal-robust-regret} provides an explicit closed-form formula that can be computed in $O(T)$ (linear) time given the sequences $\{\triangle_t\}_t$ and $\{V_t\}_t$. This formula also enables the fast identification of ``bottleneck'' days --- those where improving prediction accuracy would lead to a reduction in robust regret.

\subsection{Optimizing Robust Competitive Ratio}
\label{sec:app-I-robust-competitive-ratio}
In this section, we derive an online policy that achieves the optimal robust competitive ratio for any problem instance $\mathcal{J}$. Compared to regret, the competitive ratio metric may offer a more meaningful comparison to the optimal hindsight policy due to its normalization, particularly in scenarios where demand exhibits high variability. On the other hand, it is not clear how the method by \cite{feng2024robust} for the optimal robust regret can be applied to competitive ratio. Specifically, their method crucially relies on the key observation that the adversary does not lose any power in inflating the decision-maker's regret when restricted to a small subset of prediction interval sequences, referred to as the ``single-switching sequences''. However, in the context of the competitive ratios metric, such a restriction could substantially hurt the adversary's power. In Section~\ref{sec:counterexample-for-app-I-robust--competitive-ratio}, we provide an example where the adversary has to utilize non-single-switching prediction sequences to minimize the decision-maker's robust competitive ratio.

Now we turn to our minimax-MDP-based solution. Similarly to Section~\ref{sec:app-I-robust-regret}, we  assume w.l.o.g.~that $\ovld_0 \leq V = \sum_{t=1}^T V_t$, and first work with the decision problem of whether there exists an online policy achieving a robust competitive ratio at least a given threshold $\Phi \leq 1$, i.e., whether
\begin{align}\label{eq:decision-problem-app-I-robust-comepetive-ratio}
\sup_{\pi} \{ \varphi_\pi (\mathcal{J}) \} \geq \Phi.
\end{align}

The reduction from $\mathcal{J}$ to a minimax-MDP $\mathcal{I}$ is almost the same as that described in Section~\ref{sec:app-I-robust-regret}, where the only difference lies in the environment-dependent inventory constraint at time period $T+1$ (Eq.~\eqref{eq:app-I-robust-regret-L-R-T-plus-1}), which is now given by
\begin{align}\label{eq:app-I-robust-competitive-ratio-L-R-T-plus-1}
    L_{T+1}(d) = d \cdot \Phi, \qquad \text{and} \qquad  R_{T+1}(d) = d \cdot \frac{(1-\Phi) \cdot  p + \Phi \cdot c}{c}, \qquad \forall d \geq 0.
\end{align}

We then have the following lemma about our reduction.

\begin{lemma}\label{lem:app-I-reduction-competitive-ratio}
Given $\mathcal{J}$, let the minimax-MDP $\mathcal{I}$ be constructed by the reduction in this subsection. Then Eq.~\eqref{eq:decision-problem-app-I-robust-comepetive-ratio} holds if and only if $\mathcal{I}$ admits a feasible policy.
\end{lemma}
\proof{Proof.}
Following the proof of Lemma~\ref{lem:app-I-reduction}, we define the same $\sigma$ mapping and only need to show that $\pi^\mathcal{I} = \sigma(\pi^\mathcal{J})$ is a feasible policy for $\mathcal{I}$ if and only if $\pi^\mathcal{J}$ is a valid ordering policy with $\varphi_\pi(\mathcal{J}) \geq \Phi$.  

Define $x_{T+1}^{\mathcal{I}}$ and $x_{T+1}^{\mathcal{J}}$ in the same way and denote their (identical) value by $x$. Along the same lines, our proof reduces to demonstrating that for any demand $d$, $L_{T+1}(d) \leq x \leq R_{T+1}(d)$ if and only if the competitive ratio of $\pi^{\mathcal{J}}$, given demand $d$ and total ordering amount $x$, does not drop below $\Phi$, i.e.,
\[
x \in \left[d \cdot \Phi,  d \cdot \frac{(1-\Phi) \cdot  p + \Phi \cdot c}{c}\right]  \Leftrightarrow \frac{p \cdot \min(d, x) - c \cdot x}{(p-c) \cdot d} \geq \Phi,
\]
which can also be easily verified by considering the two cases: $x \leq d$ and $x > d$.
\hfill\Halmos
\endproof

By Lemma~\ref{lem:app-I-reduction-competitive-ratio}, we invoke Theorem~\ref{thm:main-equivalent-conditions} with $\mathcal{I}$ to derive the following equivalent conditions of Eq.~\eqref{eq:decision-problem-app-I-robust-comepetive-ratio}, proof details of which will be provided in Section~\ref{sec:proof-of-app-I-optimal-competitive-ratio}.
\begin{proposition}
\label{prop:conclusion-app-I-optimal-robust-competitive-ratio}
Eq.~\eqref{eq:decision-problem-app-I-robust-comepetive-ratio} holds if and only if
\begin{align*}
\Phi \leq \Phi^{*} = \mathop{\min}\limits_{t\in [T]}\left\{\frac{p \cdot \unld_0 + c \cdot \sum_{s=t+1}^{T} V_s}{p \cdot \unld_0 + c \cdot \triangle_t}\right\}.
\end{align*}
Moreover, the optimal ordering policy to achieve the robust competitive ratio $\Phi^*$ is
\begin{align*}
\pi_{t}([\unld_t,\ovld_t],x_t)=\min\left\{V_t,\unld_t \cdot \frac{(1 - \Phi^*) \cdot p + \Phi^* \cdot c}{c} - x_t\right\}, \qquad \forall t \in \{1, 2, \dots, T\}.
\end{align*}
\end{proposition}

\section{Application II: Robust Resource Allocation with Predictions}
\label{sec:robust-resource-allocation-with-predictions}

In this section, we study the problem of strategically allocating resources over multiple periods to two sectors in order to maximize overall utility. In each period, the decision-maker distributes a fixed amount of resources between the two sectors, subject to that period’s resource constraint. The objective is to maximize the overall utility $u(x, y; \bm{\theta})$  at the end of the time horizon, where $x$ and $y$ denote the cumulative resources allocated to each sector, and $u(\cdot, \cdot; \bm{\theta})$ is a parametric utility function. The parameter vector $\bm{\theta}$ is initially unknown to the decision-maker. However, as in the problem considered in Section~\ref{sec:app-I-robust-multi-period-ordering}, the decision-maker receives a prediction interval for each parameter in $\bm{\theta}$ at the beginning of every period, with the prediction intervals becoming increasingly accurate over time. The goal of the \emph{robust resource allocation with predictions (RRAwP)} problem is to design a computationally efficient online algorithm that achieves competitive overall utility compared to the optimal allocation in hindsight, while maintaining robustness against the worst-case sequence of prediction intervals.

The RRAwP modeling framework captures practical scenarios such as the investment planning of a manufacturing firm, where the decision-maker allocates the budget in each investment phase between capital $K$ (e.g., machinery, tools, and factory space) and labor $L$ (e.g., workers).  At the end of the investment horizon, the firm’s productivity is evaluated via the Leontief production function $u_{\mathrm{Leontief}}(K, L; \theta_1, \theta_2) = \min(K/\theta_1,L/\theta_2)$. The output elasticity parameters $\bm{\theta} = (\theta_1, \theta_2)$ are estimated from data (using, for example, the methods of \cite{aigner1977formulation}), and the resulting prediction intervals serve as the input to the RRAwP problem. As more data become available over time, the parameter estimates (and thus the predictions) become more accurate. This modeling framework also extends to other investment allocation problems, such as distributing the annual government budgets between education and healthcare to maximize long-term social welfare.

Similar to Section~\ref{sec:app-I-robust-multi-period-ordering}, our minimax-MDP framework efficiently addresses the RRAwP problem under both competitiveness metrics: \emph{regret} and \emph{competitive ratio}. For conciseness, we focus on demonstrating how to achieve the optimal robust competitive ratio using our minimax-MDP framework, noting that the algorithm for the optimal robust regret can be derived in a similar manner.

\subsection{Problem Formulation}
\label{sec:app-II-problem-formulation}
The RRAwP problem consists of $T$ periods. In each $T$ period, the decision-maker allocates resources to two sectors, A and B. The parameter vector $\bm{\theta} = (\theta_1, \theta_2, \dots, \theta_d)$ of the parametric utility function $u(\cdot, \cdot, \bm{\theta})$ is unknown. However, at the beginning of each period $t\in[T]$, the decision-maker receives a forecast $\bm{\mathcal{P}}_t = (\mathcal{P}_{t,1}, \mathcal{P}_{t,2}, \dots, \mathcal{P}_{t,d})$, where each $\mathcal{P}_{t,i} = [\underline{\theta}_{t, i}, \overline{\theta}_{t, i}]$ is a prediction interval for $\theta_i$. We denote the prediction sequences as $\bm{\mathcal{P}} = \{\bm{\mathcal{P}}_t\}_{t \in [T]}$. We extend Assumption~\ref{asp:regularityofpredictionsequence} to  $\bm{\mathcal{P}}$ as follows.

\begin{assumption}[Regularity of multivariate prediction sequences]
\label{asp:regularity-multivariate-prediction-sequence}
Given the intervals $\bm{\mathcal{P}}_0 = ([\underline{\theta}_{0, i}, \overline{\theta}_{0, i}])_{i \in [d]}$ and a sequence $\{\bm{\Delta}_{t} = (\triangle_{t, 1}, \triangle_{t,2}, \dots, \triangle_{t,d})\}_{t\in[T]}$, we say $\bm{\mathcal{P}}=\{\bm{\mathcal{P}}_t\}_{t\in[T]}$ is a \emph{regular multivariate prediction sequence} under $\{\bm{\mathcal{P}}_0 ,\{\bm{\Delta}_{t}\}_{t\in[T]}\}$ if for each day $i\in [d]$, the prediction sequence $\{\mathcal{P}_{t,i}\}_{t \in [T]}$ is regular with respect to $[\underline{\theta}_{0, i}, \overline{\theta}_{0, i}]$ and $\{\triangle_{t, i}\}_{t \in [T]}$.
\end{assumption}

\begin{definition}
An instance of the robust resource allocation with prediction problem is defined by the tuple:
\[
    \mathcal{M}\defeq \{T,\bm{\mathcal{P}}_0,\{\bm{\Delta}_t\}_{t\in[T]},\{V_t\}_{t\in[T]},u(\cdot, \cdot ;\cdot )\},
 \]
where $T,\bm{\mathcal{P}}_0,\{\bm{\Delta}_t\}_{t\in[T]},u(\cdot, \cdot ;\cdot )$ are as previously defined, and the \emph{resource supply} $V_t$ ($t \in [T]$) denotes the total amount of resources available in period $t$.

Without loss of generality, let $\triangle_{0, i}$ denote the length of the initial prediction interval $\mathcal{P}_{0,i}$, and assume that $\{\triangle_{t, i}\}_{t\in \{0, 1, \dots, T\}}$ is non-increasing for each $i \in [d]$. We further assume that, for all possible $x, y$, and $\bm{\theta}$, both $u(\cdot, y; \bm{\theta})$ and $u(x, \cdot; \bm{\theta})$ are non-decreasing and $u(\cdot,\cdot;\bm{\theta})$ is concave, representing nonnegative but diminishing returns to investment in each sector.
\end{definition}

Given an instance $\mathcal{M}$, a feasible policy $\pi=(\pi_1,\dots,\pi_T)$  is in the following form:
\begin{align}
    \pi_t: \mathop{\bigtimes}_{i=1}^{d} \{[a_i,b_i]:0\leq b_i-a_i\leq \triangle_{t,i}\} \times \mathbb{R}\to [0,V_t] ,\qquad \forall t\in[T].
\end{align}
In period $t$, $\pi_t$ decides the amount of resource allocated for project A, based on the total resource investment in sector A before period $t$ (the final argument of the function) and the prediction intervals for $\bm{\theta}$ received in period $t$ (the first $d$ arguments of the function). Note that since the utility function is non-decreasing, all resources should be fully utilized to maximize the total profit. Therefore, we may assume that the decision-maker allocates the remaining resources $V_t-\pi_t(\cdot)$, in the period, to sector B.

\medskip
\noindent\uline{Timeline.} We formalize the timeline of the model as follows:
\begin{itemize}
\item Initially: the problem parameters are revealed to the decision-maker. The initial investment for sector A is $x_1=0$, while the investment for sector B is $y_1=0$.
\item In each period $t\in[T]$: 
\begin{itemize}
\item prediction intervals $\bm{\mathcal{P}}_t$ are revealed to the decision-maker;
\item the decision-maker allocates 
$a_{t}=\pi_t(\mathcal{P}_{t,1}, \mathcal{P}_{t,2}, \dots, \mathcal{P}_{t,d}, x_t)$ amount of
resource investment to sector A, and the rest to sector B, updating the cumulative resource investments:
$x_{t+1}=x_t+a_t, y_{t+1}=y_t+(V_t-a_t)$.
\end{itemize}
\item  At the end of the horizon: the parameter vector $\bm{\theta}$ is revealed, and the overall utility is 
\begin{align}
\mathcal{R}_\pi(\bm{\mathcal{P}},\bm{\theta};\mathcal{M})=u(x_{T+1}, y_{T+1}; \bm{\theta}).
\end{align}
\end{itemize}
The utility of the hindsight optimal policy with the knowledge of $\bm{\theta}$ is
\begin{align}\label{eq:app-II-profit-offline-optimal}
\mathcal{R}^\sharp (\bm{\mathcal{P}}, \bm{\theta}; \mathcal{M}) = \mathrm{opt}(\bm{\theta}; \mathcal{M}) \defeq \sup_{0\leq x\leq V} \{u(x, V-x; \bm{\theta})\},
\end{align}
where $V\defeq\sum_{t=1}^{T}V_t$ is the total amount of available resources over the development periods. 

\medskip
\noindent\uline{Competitive ratio.} Given $\bm{\mathcal{P}},\bm{\theta}$, the \emph{competitive ratio} of a feasible policy $\pi$ is
\begin{align}
     \varphi_\pi (\bm{\mathcal{P}},\bm{\theta}; \mathcal{M}) \defeq \frac{\mathcal{R}_\pi (\bm{\mathcal{P}},\bm{\theta}; \mathcal{M})}{ \mathcal{R}^\sharp (\bm{\mathcal{P}},\bm{\theta}; \mathcal{M})}. 
\end{align}
The robust (worst-case) competitive ratio of $\pi$ is defined as
\begin{align}
\varphi_\pi(\mathcal{M}) \defeq \inf_{\bm{\mathcal{P}},\bm{\theta}} \{\varphi_\pi (\bm{\mathcal{P}},\bm{\theta}; \mathcal{M})\},
\end{align}
where the infimum is taken over all $(\bm{\mathcal{P}},\bm{\theta})$ tuples such that $\bm{\mathcal{P}}$ is a regular multivariate prediction sequence with respect to $\{\bm{\mathcal{P}}_0, \{\bm{\Delta}_t\}_{t\in[T]}\}$, and $\bm{\theta}$ is consistent with the final prediction intervals $\bm{\mathcal{P}}_T$.

\subsection{The Minimax-MDP-based Solution}
\label{sec:app-II-reducing-to-minimax-MDP}
We now work with the decision problem of whether there exists an online policy that achieves a robust competitive ratio at least a given threshold $\Phi \leq 1$, i.e., deciding whether
\begin{align}\label{eq:decision-problem-app-II-robust-comepetive-ratio}
\sup_{\pi} \{ \varphi_\pi (\mathcal{M}) \} \geq \Phi.
\end{align}
\noindent\uline{{The reduction.}} To apply the minimax-MDP framework, we reduce the decision problem Eq.~\eqref{eq:decision-problem-app-II-robust-comepetive-ratio} to a minimax-MDP $\mathcal{I}$ with time horizon $T+2$. Similarly to the reductions in Section~\ref{sec:app-I-robust-multi-period-ordering}, we label the time periods by $0, 1, 2, \dots, T+1$, corresponding to the periods in $\mathcal{M}$, while we define the environment states in $\mathcal{I}$ in a slightly different way --- the environment states should correspond to the pair of prediction intervals for the demands of both projects, i.e., we let
\begin{align*}
S_{t}=\mathop{\bigtimes}_{i=1}^{d} \{[a_i,b_i]:0\leq b_i-a_i\leq \triangle_{t,i}\},~~~~ \forall 0\leq t\leq T,\qquad \text{and}\qquad  S_{T+1}=\{(\theta_i)_{i\in[d]}; \theta_i \in \mathbb{R}, \forall i \in [d]\}.
\end{align*}
Note that in period $T+1$, the parameter vector is realized and therefore the intervals collapse to $d$ numbers, represented by our definition of $S_{T+1}$. Initially, in $\mathcal{I}$, we set 
\begin{align*}
s_0 \defeq \bm{\mathcal{P}}_0=([\underline{\theta}_{0, i}, \overline{\theta}_{0, i}])_{i \in [d]}, \qquad \text{and}\qquad x_0 = 0.
\end{align*}
For each $t\in\{0,1,\dots,T-1\}$, we set the environment state transition functions as follows:
\begin{align*}
F_{t}(([a_i,b_i])_{i\in[d]})=\{([e_i,f_i])_{i\in[d]};~\forall i\in[d],~0\leq f_i-e_i\leq\triangle_{t+1,i},[e_i,f_i]\subseteq [a_i,b_i]\},
\end{align*}
while we define the transition functions in time period $T$ as 
\begin{align*}
    F_{T}(([a_i,b_i])_{i\in[d]})=\{(\theta_i)_{i\in[d]};~\forall i\in[d],~\theta_i\in[a_i,b_i]\}.
\end{align*}
Similar to the reductions in Section~\ref{sec:app-I-robust-multi-period-ordering}, we define the trivial environment-dependent action  and inventory constraints (where $V_t$ is the supply capacity in $\mathcal{M}$):
\begin{align}
& U_{t}(\cdot)\equiv 0, ~~~~V_{t}(\cdot)\equiv V_{t}, ~~~~ L_t(\cdot)\equiv 0, ~~~~ R_t(\cdot)\equiv +\infty, & & \forall t\in\{1,2,\dots,T\}, \\
& U_0(\cdot) \equiv V_{0}(\cdot)\equiv  L_0(\cdot) \equiv  R_0(\cdot) \equiv U_{T+1}(\cdot) \equiv V_{T+1}(\cdot)\equiv 0. 
\end{align} 
The only non-trivial constraints $L_{T+1}(\cdot)$ and $R_{T+1}(\cdot)$ are related to the threshold $\Phi$ in Eq.~\eqref{eq:decision-problem-app-II-robust-comepetive-ratio} and the hindsight optimum $\mathrm{opt}(\bm{\theta}; \mathcal{M})$ defined in Eq.~\eqref{eq:app-II-profit-offline-optimal}: 
\begin{align} 
&L_{T+1}(\bm{\theta};\mathcal{M})=\inf\{x\in[0,V];~u(x,V-x;\bm{\theta})\geq \Phi\cdot \mathrm{opt}(\bm{\theta};\mathcal{M})\}, \\
&R_{T+1}(\bm{\theta};\mathcal{M})=\sup\{x\in[0,V];~u(x,V-x;\bm{\theta})\geq \Phi\cdot \mathrm{opt}(\bm{\theta};\mathcal{M})\}.
\end{align}
\begin{observation}\label{obs:app-II-reduce-equivalent-in-selling-stage}
    For any $\bm{\theta}\in \mathbb{R}_{\geq 0}^d$ and $\Phi\leq 1$, we have
    \[
    \emptyset \subsetneqq \{x\in[0,V];~u(x,V-x;\bm{\theta})\geq \Phi\cdot \mathrm{opt}(\bm{\theta};\mathcal{M})\}=[L_{T+1}(\bm{\theta};\mathcal{M}),~R_{T+1}(\bm{\theta};\mathcal{M})]\subset [0,V] .
    \]
\end{observation}
\proof{Proof.} 
Due to the fact that $u(\cdot,\cdot;\bm{\theta})$ is concave, we prove that $u(x,V-x;\bm{\theta})$ is concave with respect to $x$ by noticing that for all $\lambda \in [0, 1]$, it holds that
\[
\lambda\cdot u(x,V-x;\bm{\theta})+(1-\lambda)\cdot u(x',V-x';\bm{\theta})\geq u(\lambda\cdot x+(1-\lambda)\cdot x',V-\lambda\cdot x-(1-\lambda)\cdot x';\bm{\theta}).
\]
Thus we have $u(x,V-x;\bm{\theta})$ is concave with respect to $x$. Therefore, the set $\{x;~u(x,V-x;\bm{\theta})\geq \Phi\cdot \mathrm{opt}(\bm{\theta};\mathcal{M})\}$ is convex and equals $[L_{T+1}(\bm{\theta};\mathcal{M}),~R_{T+1}(\bm{\theta};\mathcal{M})]$. Moreover, we have 
\[
\arg\max_{x\in[0,V]} u(x,V-x;\bm{\theta}) \in \{x\in[0,V];~u(x,V-x;\bm{\theta})\geq \Phi\cdot \mathrm{opt}(\bm{\theta};\mathcal{M})\},
\]
which indicates the non-empty property of the set.
\hfill\Halmos
\endproof

We prove the following lemma about our reduction.
\begin{lemma}\label{lem:app-II-reduction}
    Given $\mathcal{M}$, let the minimax-MDP $\mathcal{I}$ be constructed by our reduction. Then Eq.~\eqref{eq:decision-problem-app-II-robust-comepetive-ratio} holds if and only if $\mathcal{I}$ admits a feasible policy.
\end{lemma}
\proof{Proof.}
For any valid allocation policy $\pi^\mathcal{M}$ for $\mathcal{M}$, consider the mapping 
\[
\sigma : \pi^\mathcal{M} \mapsto \pi^\mathcal{I} = (\pi^\mathcal{I}_0(\cdot, \cdot) \equiv 0, \pi^\mathcal{I}_1 = \pi^\mathcal{M}_1, \pi^\mathcal{I}_2 = \pi^\mathcal{M}_2, \dots, \pi^\mathcal{I}_T = \pi^\mathcal{M}_T)
\]
such that $\pi^\mathcal{I} = \sigma(\pi^\mathcal{M})$ is a policy for $\mathcal{I}$. Conversely, if $\pi^\mathcal{I}$ is a feasible policy for $\mathcal{I}$, one may verify the existence of $\sigma^{-1}(\pi^\mathcal{I})$.

To prove the lemma, we only need to show that $\pi^\mathcal{I} = \sigma(\pi^\mathcal{M})$ is a feasible policy for $\mathcal{I}$ if and only if $\pi^\mathcal{M}$ is a valid allocating policy with $\varphi_\pi(\mathcal{M}) \geq \Phi$.  

For convenience, let $x_{T+1}^{\mathcal{I}}$ denote the inventory level at time period $T+1$ when running policy $\pi^\mathcal{I}$ in $\mathcal{I}$ (which can be naturally defined even if the policy does not meet the environment-dependent action constraints). Similarly, let $x_{T+1}^{\mathcal{M}}$ be the total resource investment for project A over the development periods under the allocation policy $\pi^\mathcal{M}$ in $\mathcal{M}$. If $\pi^\mathcal{I} = \sigma(\pi^\mathcal{M})$, we directly have $x_{T+1}^{\mathcal{I}} = x_{T+1}^{\mathcal{M}}$, denoted by $x$. 

Note that the only non-trivial constraint in $\mathcal{I}$ is the environment-dependent inventory constraint at time period $T+1$. Hence, our proof further reduces to demonstrating that for any $\bm{\theta}\in \mathbb{R}_{\geq 0}^d$, $L_{T+1}(\bm{\theta};\mathcal{M}) \leq x \leq R_{T+1}(\bm{\theta};\mathcal{M})$ if and only if the competitive ratio of $\pi^{\mathcal{M}}$, given $\bm{\theta}$ and project A's total resource investment $x$, is at least $\Phi$, i.e.,
$x\in[L_{T+1}(\bm{\theta};\mathcal{M}),~R_{T+1}(\bm{\theta};\mathcal{M})]
\Leftrightarrow u(x,V-x;\bm{\theta})\geq \Phi\cdot \mathrm{opt}(\bm{\theta};\mathcal{M})$,
which can be directly derived from Observation~\ref{obs:app-II-reduce-equivalent-in-selling-stage}.
\hfill\Halmos
\endproof

By Lemma~\ref{lem:app-II-reduction}, we invoke Theorem~\ref{thm:main-equivalent-conditions} with $\mathcal{I}$ to derive the following equivalent conditions of Eq.~\eqref{eq:decision-problem-app-II-robust-comepetive-ratio}, proof details of which will be provided in Section~\ref{sec:proof-of-robust-resource-allocation-with-predictions}.

\begin{proposition}
\label{prop:app-II-robust-competitive-ratio}
Eq.~\eqref{eq:decision-problem-app-II-robust-comepetive-ratio} holds if and only if for any $\tau,\bm{\theta}=(\theta_i)_{i\in[d]},\bm{\theta}'=(\theta'_i)_{i\in[d]}$ such that $1\leq\tau\leq T$, $\theta_i,\theta'_i\in[\underline{\theta}_{0, i}, \overline{\theta}_{0, i}]$ and $|\theta_i-\theta'_i|\leq \triangle_{\tau,i}$ for each $i$, we have 
\begin{align}
R_{T+1}(\bm{\theta};\mathcal{M})+\sum_{t=\tau+1}^{T}V_t\geq L_{T+1}(\bm{\theta}';\mathcal{M}). \label{eq:app-II-R-plus-sum-bigger-than-L}
\end{align}
          

Moreover, the corresponding feasible policy for each $t\in[T]$ is 
\[
\pi_t(\{[a_i,b_i]\}_{i\in[d]},x_t)=\min\left\{V_t,\min_{\forall i\in[d],\theta_i\in[a_i,b_i]}[R_{T+1}((\theta_i)_{i\in[d]};\mathcal{M})]-x_t\right\}.
\]
\end{proposition}

\subsection{Application to the Leotief Production Utility Function}
Applying Proposition~\ref{prop:app-II-robust-competitive-ratio} to the Leontief production utility function example introduced at the beginning of this section, we have the following closed-form solution for the optimal robust competitive ratio, the proof of which will be provided in Section~\ref{sec:proof-of-best-robust-competitive-ratio-of-Leontief-utility}.
\begin{proposition}
\label{prop:best-robust-competitive-ratio-of-Leontief-utility}
    Consider the Leontief utility function $u_{\mathrm{Leontief}}(x,y;\bm{\theta})=\min(x/\theta_1,y/\theta_2)$, then the optimal robust competitive ratio is 
    $\Phi^*=\min_{1\leq \tau \leq T} \left\{ f_\tau^{-1}\right\}$,
    where  
    \[
        f_\tau = \max_{\theta_1\in[\underline{\theta}_{0,1},\overline{\theta}_{0,1}-\triangle_{\tau,1}] \atop \theta_2\in[\underline{\theta}_{0,2}+\triangle_{\tau,2},\overline{\theta}_{0,2}]}\frac{V}{V+\sum_{t=\tau+1}^{T}V_t} \cdot \left( \frac{\theta_2}{\theta_1+\theta_2}+\frac{\theta_1+\triangle_{\tau,1}}{\theta_1+\theta_2+\triangle_{\tau,1}-\triangle_{\tau,2}} \right).
    \]
    Moreover, both $f_\tau$ and $\Phi^*$ admit closed-form expressions, as detailed in Section~\ref{sec:proof-of-best-robust-competitive-ratio-of-Leontief-utility}.
\end{proposition}

\section{Multi-phase Minimax-MDP}
\label{sec:multi-phase-minimax-mdp}
In this section, we generalize the minimax-MDP to a more complex version called multi-phase minimax-MDP, which can also be solved by extending results in minimax-MDP. 

\subsection{Formulation of the Multi-phase Minimax-MDP}
In a $K$-phase minimax-MDP, the $T$ time periods are divided into $K$ consecutive periods according to the grid vector $\bm{\tau} = (\tau_0 = 1, \tau_1, \tau_2, \dots, \tau_K = T)$, where $\tau_0 < \tau_1 < \dots < \tau_K$ and phase $v$ consists of the time periods $\tau_{v-1}, \tau_{v-1} + 1, \dots, \tau_v - 1$. The system state at time $t$ is represented as a vector $(s_t,\bm{x}_t)$, where $s_t$ denotes the environment state and $\bm{x}_t = (x_{t, 1}, x_{t, 2}, \dots, x_{t, K})$ denotes the cumulative inventory levels before time $t$, with the initial inventory levels given by $\bm{x}_1 = \bm{0}$.

For each phase $v$ and during each time period $t\in [\tau_{v-1}, \tau_v)$ in the phase, the decision-maker needs to select an ordering action $a_t\in \mathbb{R}$ and the inventory levels at the next time period become
$\bm{x}_{t+1}=\bm{x}_t+a_t \bm{e}_v$,
where $\bm{e}_v$ is the $v$-th canonical basis vector. 



\begin{definition}
An instance of a multi-phase minimax-MDP is defined as a tuple:
\[
    \mathcal{W}=\{T,K,\bm{\tau},\{S_t\}_{t\in[T]},s_1,\bm{x}_1,\{F_t\}_{t \in [T-1]},\{U_t, V_t\}_{t \in [T-1]},\{n_v, W_v\}_{v \in [K]}\},
\]
where the \emph{time horizon} $T$, \emph{environment state set} $\{S_t\}_{t\in[T]}$, \emph{initial environment state} $s_1$, \emph{environment state transition functions} $\{F_t\}_{t \in [T-1]}$, \emph{environment-dependent action constraints} $\{U_t, V_t\}_{t \in [T-1]}$ are defined in the same way as in Definition~\ref{def:minimax-MDP}, the \emph{phase count} $K$, the \emph{grid vector} $\bm{\tau}$, and the \emph{initial inventory level vector} $\bm{x}_1$ are described as above. In addition, the multi-phase minimax-MDP imposes the \emph{environment-dependent inventory constraints} via the \emph{constraint matrices} $\{W_v : S_{\tau_v} \to \mathbb{R}^{n_v \times (\{0\} \cup [K])}\}$ as follows: at the end of phase $v \in [K]$, the inventory level vector $\bm{x}_{\tau_v}$ has to satisfy
\begin{align} \label{eq:multi-phase-inventory-constraints}
    W_v(s_{\tau_v}) \left[
    \begin{array}{c}
    1 \\
    \bm{x}_{\tau_v}
    \end{array}
    \right] \leq \bm{0}, 
    \text{~i.e.,}\qquad
    [W_v(s_{\tau_v})]_{i, 0} \times 1 + \sum_{j=1}^{K} [W_v(s_{\tau_v})]_{i, j} \times x_{\tau_v, j} \leq 0, \qquad \forall i \in [n_v].
\end{align}
\end{definition}
In other words, Eq.~\eqref{eq:multi-phase-inventory-constraints} consists of $n_v$ environment-state-dependent linear constraints on the inventory levels at the end of phase $v$. The environment-dependent inventory constraints in the original minimax-MDP formulation (Eq.~\eqref{eq:environment-dependent-inventory-constraint}) can be expressed by two such linear constraints. We may further assume w.l.o.g.~that $[W_v(s_{\tau_v})]_{i,v} \neq 0$ for each $i \in [n_v]$, because otherwise the corresponding constraint can be moved to the previous phase.

A policy $\pi = (\pi_1, \pi_2, \dots, \pi_{T-1})$ maps the system state at any time $t \in \{1, 2, \dots, T - 1\}$ to an action: 
\begin{align}\label{eq:definition-of-policy-in-multi-phase}
\pi_t : S_t \times \mathbb{R}^K \to \mathbb{R}, \qquad \forall t \in \{1, 2, \dots, T - 1\}.
\end{align}
We naturally extend Definition~\ref{def:pi-compatible-trajectory} to the multi-phase setting --- say that a (partial) trajectory $\phi = (s_\alpha, \bm{x}_\alpha, s_{\alpha + 1}, \bm{x}_{\alpha+1}, \dots, s_\beta, \bm{x}_\beta)$ is \emph{$\pi$-compatible} if it can occur under the system $\mathcal{M}$ following the policy $\pi$, and it is \emph{feasible} if it satisfies all environment-dependent action and inventory constraints.

\medskip
\noindent \uline{\textbf{The Feasible Policy Problem for Multi-phase Minimax-MDPs.}} Similar to the single-phase minimax-MDP (Section~\ref{sec:minimax-MDP-formulation}), given a multi-phase minimax-MDP $\mathcal{M}$, a policy $\pi$ is said to be \emph{feasible} if every $\pi$-compatible trajectory under $\mathcal{M}$ is also feasible. The feasible policy problem in the multi-phase setting is to efficiently decide the existence of and compute the feasible policy. The key ingredient of our solution to the problem is a careful application of our future-imposed conditions for the single-phase minimax-MDP to reduce a $K$-phase minimax-MDP to a $(K-1)$-phase minimax-MDP. In the following section, we provide such a phase reduction theorem and solve the feasible policy problem in the multi-phase setting.

\subsection{Solving the Multi-phase Minimax-MDP via Phase Reduction Theorem}
\label{sec:solving-multi-phase-minimax-MDP-via-phase-reduction}
Consider a $K$-phase ($K \geq 2$ is a constant) minimax-MDP finite-state instance 
\[
    \mathcal{W}=\{T,K,\bm{\tau},\{S_t\}_{t\in[T]},s_1,\bm{x}_1 = \bm{0},\{F_t\}_{t \in [T-1]},\{U_t, V_t\}_{t \in [T-1]},\{n_v, W_v\}_{v \in [K]}\}.
\]
For convenience, denote $S = \max_{t \in [T]} |S_t|$ and $N = \max_{v \in [K]} \{n_K\}$.

We begin by focusing on the final phase. Suppose we follow a policy $\pi$ up to time $\tau_{K-1}$, such that the system reaches state $(s_{\tau_{K-1}}, \bm{x}_{\tau_{K-1}})$ and all constraints up to and including time $\tau_{K-1}$ are satisfied. To ensure a feasible trajectory under all environment transitions in the final phase, the task becomes a single-phase minimax-MDP
\[
\mathcal{I}_K = \{\tau_{K} - \tau_{K-1}+1, \{S_t\}_{t=\tau_{K-1}}^{\tau_{K}}, s_{\tau_{K-1}}, x_{\tau_{K-1},K} = 0, \{F_t\}_{t=\tau_{K-1}}^{\tau_{K}-1},\{U_t, V_t\}_{t=\tau_{K-1}}^{\tau_{K}-1}, \{\tilde{L}_t, \tilde{R}_t\}_{t=\tau_{K-1}}^{\tau_{K}}\},
\]
where $\tilde{L}_{\tau_{K-1}}(s_{\tau_{K-1}}) = \tilde{R}_{\tau_{K-1}}(s_{\tau_{K-1}}) = 0$, $\tilde{L}_t(\cdot) \equiv -\infty$ and $\tilde{R}_t(\cdot) \equiv +\infty$ for each time $t \in (\tau_{K-1}, \tau_{K})$, and at time $\tau_{K}$, for each $s \in S_{\tau_K}$,
\begin{align}
\tilde{L}_{\tau_{K}}(s) & = \max_{i: [W_K(s)]_{i,K} < 0} \left\{-\frac{1}{[W_K(s)]_{i,K}}\left([W_K(s)]_{i,0} + \sum_{j=1}^{K-1} [W_K(s)]_{i,j} \times x_{\tau_{K-1},j}\right)\right\}, \label{eq:multi-phase-inventory-constraint-left-final-phase}\\
\tilde{R}_{\tau_{K}}(s) & = \min_{i: [W_K(s)]_{i,K} > 0} \left\{-\frac{1}{[W_K(s)]_{i,K}}\left([W_K(s)]_{i,0} + \sum_{j=1}^{K-1} [W_K(s)]_{i,j} \times x_{\tau_{K-1},j}\right)\right\} , \label{eq:multi-phase-inventory-constraint-right-final-phase}
\end{align}
so that the inventory constraint in $\mathcal{I}_K$ ($\tilde{L}_{\tau_{K}}(s) \leq x_{\tau_K,K}\leq \tilde{R}_{\tau_{K}}(s)$), is equivalent to the inventory constraint in $\mathcal{W}$ (Eq.~\eqref{eq:multi-phase-inventory-constraints} with $v = K$) at time $\tau_K$. 

By Theorem~\ref{thm:main-equivalent-conditions}, there exists a feasible policy for $\mathcal{I}_K$ if and only if the following future-imposed conditions hold: 
\begin{align}
 &\tilde{R}_{\star \leadsto \tau_{K-1}}(s_{\tau_{K-1}}) \geq \tilde{L}_{\star \leadsto \tau_{K-1}}(s_{\tau_{K-1}}), \label{eq:multi-phase-reduction-final-phase-two-point-0} \\
& \tilde{R}_{\star \leadsto \alpha - 1 \to [\alpha]} (s_{\alpha-1})\geq \tilde{L}_{\star \leadsto \alpha - 1 \to [\alpha]} (s_{\alpha-1}), \quad \forall \alpha \in (\tau_{K-1}, \tau_K],  s_{\alpha - 1} \in F_{\tau_{K-1}\to\alpha - 1}(s_{\tau_{K-1}}), \label{eq:multi-phase-reduction-final-phase-two-point-1} 
\end{align}
where $\tilde{R}_{\star \leadsto \cdot}$, $\tilde{R}_{\star \leadsto \cdot \to [\cdot]}$, $\tilde{L}_{\star\leadsto \cdot}$ and $\tilde{L}_{\star \leadsto \cdot \to [\cdot]}$ are defined using Eq.~\eqref{eq:def-R-star} and Eq.~\eqref{eq:def-R-star-bracket} based on $\tilde{R}$ and $\tilde{L}$. In Lemma~\ref{lem:phase-reduction-in-the-last-phase}, we demonstrate that Eq.~\eqref{eq:multi-phase-reduction-final-phase-two-point-0} and Eq.~\eqref{eq:multi-phase-reduction-final-phase-two-point-1} can be equivalently expressed as a set of 
$  \widehat{n}_{K-1}^+ = 3\cdot (\tau_K - \tau_{K-1}) \cdot n_K^2 \cdot S^4
$
linear constraints on $\bm{x}_{\tau_{K-1}}$. Combining with the initial $n_{K-1}$ linear constraints at time $\tau_{K-1}$, we denote these constraints as the matrix $\widehat{W}_{K-1}(s_{\tau_{K-1}}) \in \mathbb{R}^{\hat{n}_{K-1} \times (\{0\} \cup [K-1])}$ in the form of Eq.~\eqref{eq:multi-phase-inventory-constraints}, where $\widehat{n}_{K-1} = n_{K-1}+\widehat{n}_{K-1}^+$. In this way, we define $\widehat{W}_{K-1}(s)$ for every $s \in S_{\tau_{K-1}}$. Let $\widehat{T} = \tau_{K-1}$, we finally construct a $(K-1)$-phase minimax-MDP instance 
\[
\widehat{\mathcal{W}} = \{\widehat{T}, K-1, \bm{\tau}, \bm{x}_1 = 0,  \{S_t\}_{t \in [\widehat{T}]}, \{F_t\}_{t \in [\widehat{T}-1]},\{U_t, V_t\}_{t \in [\widehat{T}-1]},\{n_v, W_v\}_{v \in [K-2]} \cup \{\widehat{n}_{K-1}, \widehat{W}_{K-1}\}\} .
\]
The following phase reduction theorem, whose proof is deferred to Section~\ref{sec:proof-of-phase-reduction}, is the main ingredient of our solution to the multi-phase minimax-MDP,
\begin{theorem}[Phase reduction theorem]
\label{thm:phase-reduction}
The $K$-phase minimax-MDP $\mathcal{W}$ admits a feasible policy if and only if the corresponding $(K-1)$-phase minimax-MDP $\widehat{\mathcal{W}}$ admits a feasible policy.    
\end{theorem}

We may repeatedly invoke Theorem~\ref{thm:phase-reduction} until a multi-phase minimax-MDP is reduced to a single-phase minimax-MDP with $\exp(O(\log (NTS) \times 2^K)$ constraints, whose feasibility can be decided by verifying the corresponding future-imposed conditions in $\exp(O(\log (NTS) \times 2^K)$ time, which is polynomial in $N$, $T$, $S$ when $K$ is a constant.

\section{Application III: Robust Ordering Decisions under Changing Costs}
\label{sec:Multi-period-Ordering-Decisions-with-Multi-phase-costs-and-Predictions}
We now examine a generalized version of the application described in Section~\ref{sec:app-I-robust-multi-period-ordering}, and demonstrate how to use the multi-phase minimax-MDP framework to solve the problem. We inherit the setting from the multi-period ordering decision with predictions problem in Section~\ref{sec:app-I-problem-formulation} where the only difference is that the ordering costs may be different across the preparation days. Specifically, the $T$ preparation days are divided into $K$ phases based on a \emph{grid vector}  $\bm{\tau}=(\tau_0 = 1, \tau_1, \tau_2, \dots, \tau_K = T+1)$ where $\tau_0 < \tau_1 < \dots < \tau_K$ and phase $v$ consists of days $\tau_{v-1}$ to $\tau_v - 1$. On each day $t$ in phase $v$, the per unit ordering cost is $c_t = \gamma_v$, where $\bm{\gamma} = (\gamma_1, \gamma_2, \dots, \gamma_K)$ is known to the decision-maker. We assume that $\gamma_1 \leq \gamma_2 \leq \dots \leq \gamma_K$, which is natural in many practical scenarios, as the earlier an order is placed, the longer the production lead time is available, resulting a lower contracted price. We also assume $\gamma_K < p$ where $p$ is the revenue generated per unit of fulfilled demand.

Given a problem instance
$\mathcal{V}=\{T,K,\bm{\tau}, [\unld_0,\ovld_0],\{\triangle_t\}_{t\in[T]}, \bm{\gamma}, p, \{V_t\}_{t\in[T]}\}$,
along the same timeline as described in Section~\ref{sec:app-I-problem-formulation}, the decision-maker makes daily ordering decisions based on a policy $\pi = (\pi_1, \pi_2, \dots, \pi_T)$, to meet the demand $d$ that will be realized on day $T+1$, with the access of a regular prediction sequence for $d$ under the parameters $\{[\unld_0,\ovld_0],\{\triangle_{t}\}_{t\in[T]}\}$, and subject to the supply capacity constraints $\{V_t\}_{t\in [T]}$. On day $T+1$, when the demand $d$ is revealed, the profit is
\begin{align}
\mathcal{R}_{\pi}(\mathcal{P}, d; \mathcal{V}) \defeq \text{(revenue)} -\text{(total ordering cost)}=p\cdot\min(d,x_{T+1})-\sum\limits_{t=1}^{T}c_{t}\cdot a_t .    
\end{align} 
It is also straightforward to derive the hindsight optimal revenue (of the optimal policy with the knowledge of $d$) as
\begin{align}
\mathcal{R}^\sharp (\mathcal{P}, d; \mathcal{V}) = \mathcal{R}^\sharp (\cdot, d; \mathcal{V})  = \sum_{t=1}^{T} (p - c_t) \times \min\left\{V_t, \max\left\{d - \sum_{i=1}^{t-1} V_i, 0\right\}\right\} .
\end{align}

For illustration purposes, we consider the task of finding a robust ordering policy $\pi$ to maximize its competitive ratio, $\varphi_\pi (\mathcal{V})$, which is defined in the same way as in Eq.~\eqref{eq:app-I-competitive-metrics}. To decide whether the optimal competitive ratio $\sup_\pi \{\varphi_\pi\}$ is at least a given threshold $\Phi$, we reduce the problem to a multi-phase minimax-MDP following the natural extension of the reduction presented for Application I: the $K$ cost phases correspond to the $K$ phases in the minimax-MDP under the same grid vector $\bm{\tau}$, the prediction intervals and the finally revealed demand correspond to the environment states, and the inventory vector in the minimax-MDP keeps the amount of orders made during each cost phase. The only two non-trivial environment-dependent inventory constraints are imposed at time $\tau_K = T+1$, representing the competitive ratio guarantee:
$\langle \bm{\gamma}, \bm{x}_{T+1}\rangle \leq p\cdot d - \Phi\cdot \mathcal{R}^\sharp (\cdot, d; \mathcal{V})$, and
 $\langle - (p \cdot \bm{1} - \bm{\gamma}), \bm{x}_{T+1}\rangle \leq - \Phi\cdot \mathcal{R}^\sharp (\cdot, d; \mathcal{V})$.
Repeatedly invoking the phase reduction theorem (Theorem~\ref{thm:phase-reduction}) for multi-phase minimax-MDPs, we may derive an algorithm to decide whether $\sup_\pi \{\varphi_\pi\} \geq \Phi$ in time $\exp(O(\log T) \times 2^K)$, which is polynomial in $T$ when $K$ is a constant.

In Section~\ref{sec:optimizing-robust-competitive-ratio-in-T-O-K}, we refine the phase reduction analysis by leveraging the special structural property of the multi-phase minimax-MDP derived in this problem --- namely, environment states are indexed by intervals and the environment-dependent inventory constraints are piecewise-linear functions of the left endpoints of these intervals. Based on this refined analysis, Proposition~\ref{prop:multi-phase-cost-competitive-ratio} in the supplementary materials establishes that there is an algorithm that computes and achieves the optimal robust competitive ratio in time $T^{O(K)}$.

In general, the number of cost phases $K$ can be as large as $T$ and the running time of our improved algorithm still becomes super-polynomial in $T$. In Section~\ref{sec:a-ptas-for-robust-ordering-decisions-under-changing-costs}, we present an algorithm to reduce the number of phases to $O(1/\epsilon)$ by rounding the daily ordering costs to an exponentially spaced grid, while the loss in the competitive ratio introduced by the rounding error is upper bounded by $\epsilon$. Consequently, the algorithm achieves a robust competitive ratio that is at least the optimum minus $\epsilon$ in time $T^{O(1/\epsilon)}$, yielding a PTAS for general $K$.

\section{Conclusion and Future Directions}
\label{sec:summaryandfuturedirections}
In this paper, we introduce a minimax-MDP framework that encompasses a variety of learning-augmented problems. We further develop a set of \emph{future-imposed conditions} that characterizes the feasibility of a minimax-MDP, leading to efficient, often closed-form solutions for the learning-augmented applications. For future research, it would be promising to extend the application of our framework to more general prediction structures beyond the interval-based predictions illustrated here. Another worthwhile and technically interesting direction is to generalize the one-dimensional internal state to a high-dimensional setting, which could enable the further application of our framework to multi-item inventory control and other complex resource allocation problems.

\bibliographystyle{informs2014} 
\bibliography{references} 

\ECSwitch

\ECHead{Electronic Companion}

\section{Omitted Proofs in Section~\ref{sec:mainthm}}
\label{sec:proofanddiscussionsformainthm}

Note that the domain of $F_{t}(\cdot)$ is $S_{t}$, and we may naturally extend its domain to $\mathcal{P}(S_{t})$ by
$F_{t}(A)\defeq \{y;~\exists x\in A,~\text{s.t.}~y\in F_{t}(x)\}$. 
It is also easy to deduce the following claim which will be used later.
\begin{claim}
\label{claim:dpforV}
For any $a, b \in [T]$ satisfying $a\leq b<T$, and any environment states $s_{a}\in S_{a}$, $s_{b+1}\in F_{a \to (b+1)}(s_{a})$, it holds that
$V_{a\to b+1}(s_{a},s_{b+1})=\mathop{\min}\limits_{s_{b}'\in F_{a\to b}(s_{a}): s_{b+1}\in F_{b}(s_{b}')}[V_{a\to b}(s_{a},s_{b}')+V_{b}(s_{b}')]$.
\end{claim}

\subsection{Proof of Lemma~\ref{lem:future-imposed-upper-bound}}
\label{sec:proof-of-lemma-future-imposed-upper-bound}
We only prove the first inequality ($L_{\star \leadsto t}(s_t) \leq x_t$). The second inequality ($x_t \leq R_{\star \leadsto t}(s_t)$) can be similarly derived. To prove the inequality, it suffices to show that for each $\alpha\geq t$, we have $x_t\geq L_{\alpha\leadsto t}(s_t)$. Let $(s_t,\dots,s_{\alpha})$ be a compatible partial environment state trajectory such that 
$
L_{\alpha\leadsto t}(s_t)=L_{\alpha}(s_\alpha)-\sum_{i=t}^{\alpha-1}V_i(s_i)$.
Note $(s_1,\dots,s_\alpha)$ is a compatible partial environment. According to Observation~\ref{obs:compatible-trajectory-environment-state-trajectory}, we can derive a $\pi$-compatible partial trajectory $(s_1,x_1,\dots,s_\beta,x_\beta)$ where $x_{i+1}=x_{i}+\pi_{i}(s_i,x_i)$ for $i\in\{1,\dots,\beta-1\}$. We have $L_{\alpha \leadsto t}(s_t)=L_{\alpha}(s_{\alpha})-\sum_{i=t}^{\alpha-1}V_{i}(s_i) \leq x_{\alpha}-\sum_{i=t}^{\alpha-1}(x_{i+1}-x_{i}) =x_t$, where the inequality is due to environment-dependent action constraints and environment-dependent inventory constraints (Eq.~\eqref{eq:environment-dependent-action-constraint} and Eq.~\eqref{eq:environment-dependent-inventory-constraint}).
\hfill\Halmos

\subsection{Proof of Lemma~\ref{lem:future-imposed-upper-bound-strengthened}}
\label{sec:proof-of-lemma-future-imposed-upper-bound-strengthened}
We only prove the second inequality ($x_t \leq R_{\star\leadsto t-1 \to [t]}(s_{t-1})$) and the first one can be similarly derived.
Let $s^*_t\in F_{t-1}(s_{t-1})$ be such that $R_{\star\leadsto t-1 \to [t]}(s_{t-1}) = R_{\star\leadsto t}(s^*_{t})$ (because of Eq.~\eqref{eq:def-R-star-bracket}). We have that $(s_1,x_1,\dots,s_{t-1},x_{t-1},s^*_t,x^*_t)$ a $\pi$-compatible trajectory, where 
$x^*_t=x_{t-1}+\pi_{t-1}(s_{t-1},x_{t-1})$.
We then deduce that 
$
x_t=x_{t-1}+\pi_{t-1}(s_{t-1},x_{t-1})=x^*_t\leq R_{\star\leadsto t}(s^*_{t})=R_{\star\leadsto t-1 \to [t]}(s_{t-1})$,
where the first equality is due to the condition that $(s_1,x_1,\dots,s_t,x_t)$ is a $\pi$-compatible trajectory and the inequality is due to Lemma~\ref{lem:future-imposed-upper-bound}.
\hfill\Halmos

\subsection{Proof of Claim~\ref{claim:tilde-L-larger-L}}
\label{sec:proof-of-tilde-L-larger-L}
By the definition of $\tilde{L}_\beta(s_\beta)$, we discuss the following two cases:

\noindent\underline{Case 1:} There exist $1<\alpha \leq \beta$, $s_{\alpha-1} \in F_{1 \to (\alpha-1)}(s_1)$, $s_\alpha \in F_{\alpha-1}(s_{\alpha-1})$, $s_\beta \in F_{\alpha\to\beta}(s_\alpha)$ such that
$\tilde{L}_\beta(s_\beta) = R_{\star \leadsto \alpha-1\to [\alpha]}(s_{\alpha-1}) + V_{\alpha \to \beta}(s_\alpha, s_\beta)$.
In this case, by the future-imposed conditions, we have 
$R_{\star \leadsto \alpha-1 \to [\alpha]}(s_{\alpha-1})\geq L_{\star\leadsto \alpha-1 \to [\alpha]}(s_{\alpha-1})\geq L_{\star \leadsto \alpha}(s_\alpha)\geq L_\beta(s_\beta)- V_{\alpha \to \beta}(s_\alpha, s_\beta)$.
Thus, we conclude that $\tilde{L}_\beta(s_\beta)\geq L_\beta(s_\beta)$.

\noindent\underline{Case 2:} There exists $s_\beta\in F_{1\to \beta}(s_1)$ such that
$\tilde{L}_\beta(s_\beta) =R_{\star \leadsto 1}(s_1) + V_{1 \to \beta}(s_1, s_\beta)$. 
In this case, by the future-imposed conditions, we have 
$R_{\star \leadsto 1}(s_1)\geq L_{\star\leadsto 1}(s_1)\geq L_\beta(s_\beta)- V_{1 \to \beta}(s_1, s_\beta)$.
We also conclude that $\tilde{L}_\beta(s_\beta)\geq L_\beta(s_\beta)$.
\hfill\Halmos

\subsection{Proof of Claim~\ref{claim:tilde-L-beta-property}}
\label{sec:proof-of-claim-tilde-L-beta-property}
By the definition of $\tilde{L}_\beta(s_\beta)$, we discuss the following two cases:

\noindent\underline{Case 1:} There exist $1<\alpha \leq \beta$, $s_{\alpha-1} \in F_{1 \to (\alpha-1)}(s_1)$, $s_\alpha \in F_{\alpha-1}(s_{\alpha-1})$, $s_\beta \in F_{\alpha\to\beta}(s_\alpha)$ such that
$\tilde{L}_\beta(s_\beta) = R_{\star \leadsto \alpha-1\to [\alpha]}(s_{\alpha-1}) + V_{\alpha \to \beta}(s_\alpha, s_\beta)$.
In this case, we have that
$\tilde{L}_\beta(s_\beta)+V_{\beta}(s_\beta)=R_{\star \leadsto \alpha-1\to [\alpha]}(s_{\alpha-1}) + V_{\alpha \to \beta}(s_\alpha, s_\beta)+V_{\beta}(s_\beta) 
    \geq R_{\star \leadsto \alpha-1\to [\alpha]}(s_{\alpha-1}) + V_{\alpha \to (\beta+1)}(s_\alpha, s_{\beta+1}) 
    \geq \tilde{L}_{\beta + 1}(s_{\beta+1})$, 
where the first inequality is due to Claim~\ref{claim:dpforV} and the second inequality is due to Eq.~\eqref{eq:def-tilde-L-beta}.

\noindent\underline{Case 2:} There exists $s_\beta\in F_{1\to \beta}(s_1)$ such that
$\tilde{L}_\beta(s_\beta) =R_{\star \leadsto 1}(s_1) + V_{1 \to \beta}(s_1, s_\beta)$. 
In this case, we have
$    \tilde{L}_\beta(s_\beta)+V_{\beta}(s_\beta)=R_{\star \leadsto 1}(s_1) + V_{1 \to \beta}(s_1, s_\beta)+V_{\beta}(s_\beta) 
    \geq R_{\star \leadsto 1}(s_1) + V_{1 \to (\beta+1)}(s_1, s_{\beta+1})
    \geq \tilde{L}_{\beta + 1}(s_{\beta+1})$,
where the first inequality is due to Claim~\ref{claim:dpforV} and the second one is due to Eq.~\eqref{eq:def-tilde-L-beta}.
\hfill\Halmos

\section{Omitted Proofs and Discussions in Section~\ref{sec:app-I-robust-multi-period-ordering}}
\label{app:proofofUnknowndemandanddynamicdemandforecasts}

\subsection{Proof of Proposition~\ref{prop:conclusion-app-I-optimal-robust-regret}}
\label{sec:proofofAssessingbyDifferenceinwarmup}
Using the construction in Section~\ref{sec:app-I-robust-regret}, we compute the following objects:
\begin{enumerate}
\item Multi-time-period environment state transition: for any $\alpha,\beta\in\{0,\dots,T+1\}$ with $\alpha<\beta$, and any environment state $s_\alpha=[a,b]\in S_\alpha$, we have 
\begin{equation*}
    F_{\alpha\to\beta}(s_\alpha) = \begin{cases}
        \{[e,f];~0\leq f-e\leq \triangle_\beta \wedge [e,f]\subset [a,b]\},\qquad &\beta<T+1, \\[14pt]
        \{d:~d\in[a,b]\}, \qquad &\beta=T+1.
    \end{cases}
\end{equation*}
\item Multi-time-period action constraints: for any $\alpha,\beta\in\{0,\dots,T+1\}$ with $\alpha<\beta$, we have 
\begin{align*}
    U_{\alpha\to\beta}(\cdot,\cdot)\equiv 0, \qquad V_{\alpha\to\beta}(\cdot,\cdot)\equiv \sum_{t=\alpha}^{\beta-1}V_t .
\end{align*}
\item Future-imposed inventory level constraints: for any $\alpha,\beta\in\{0,\dots,T+1\}$ with $\beta\geq\alpha$, and any environment state $s_\alpha = [a,b] \in S_\alpha$, we have
\begin{equation*}
   R_{\beta \leadsto \alpha}(s_\alpha) = \left\{
    \begin{array}{lllll}
        0, \qquad &\text{when}~\beta=0, &\qquad \qquad  & a+\frac{\Gamma}{c}, \qquad &\text{when}~\beta=T+1,~\alpha< T+1,\\[5pt]
        +\infty,\qquad &\text{when}~ 0<\beta<T+1, & & d+\frac{\Gamma}{c}, \qquad &\text{when}~\beta=\alpha=T+1,~s_{\alpha}=d. 
    \end{array}
    \right.
\end{equation*}
The case when $\beta=T+1$ and $\alpha<T+1$ is because  
$    R_{(T+1) \leadsto \alpha}([a,b])=\min_{d\in F_{\alpha\to (T+1)}([a,b])}(d+\frac{\Gamma}{c}-0) 
    =a+\frac{\Gamma}{c}$,
where both equalities hold directly by definition (the case with $\alpha=\beta=T+1$ is similar). Similarly, we have
\begin{equation*}
   L_{\beta\leadsto\alpha}(s_\alpha) = \left\{
    \begin{array}{lllll}
        0, \qquad &\text{when}~\beta=0,  &&b-\frac{\Gamma}{p-c}-\sum_{t=\alpha}^{T}V_t, \qquad &\text{when}~\beta=T+1,~\alpha<T+1,\\[5pt]
        -\infty,\qquad &\text{when}~0<\beta<T+1, && d-\frac{\Gamma}{p-c},\qquad &\text{when}~\alpha=\beta=T+1,~s_\alpha=d.
    \end{array} \right.
\end{equation*}

We can also derive that 
\begin{align*}
    &R_{\star \leadsto 0}([a,b]) = 0,\qquad 
    L_{\star \leadsto 0}([a,b]) = 0,\qquad 
    R_{\star \leadsto (T{+}1)}(d) = d + \frac{\Gamma}{c},\qquad 
    L_{\star \leadsto (T{+}1)}(d) = d - \frac{\Gamma}{p{-}c}, \\
    &R_{\star \leadsto \alpha}([a,b]) = a + \frac{\Gamma}{c},\qquad 
    L_{\star \leadsto \alpha}([a,b]) = b - \frac{\Gamma}{p{-}c} - \sum_{t=\alpha}^T V_t, 
    \qquad \forall \alpha \in \{1,\dots,T\}.
\end{align*}

\item Strengthened future-imposed inventory bounds: directly from the definition and the discussion for future-imposed inventory upper bounds, we can derive that for any $\alpha\in\{0,\dots,T\}$ and $[a,b]\subset[\unld_0,\ovld_0]$ with $b-a\leq \triangle_{\alpha}$, it holds that $
R_{\star\leadsto\alpha\to[\alpha+1]}([a,b])=a+\frac{\Gamma}{c}$ and $
L_{\star\leadsto\alpha\to[\alpha+1]}([a,b])=b-\frac{\Gamma}{p-c}-\sum_{t=\alpha+1}^T V_t$.
\end{enumerate}

By Lemma~\ref{lem:app-I-reduction} and Theorem~\ref{thm:main-equivalent-conditions}, Eq.~\eqref{eq:decision-problem-app-I-robust-regret} holds if and only if the following future-imposed conditions are satisfied: for any time period $\alpha \in \{1,\dots,T+1\}$, and each $s_{\alpha-1} \in F_{0 \to \alpha-1}(s_0)$, it holds that
\begin{align} \label{eq:two-point-condition-in-multi-period}
R_{\star \leadsto \alpha -1 \to [\alpha]}(s_{\alpha-1}) \geq L_{\star \leadsto \alpha -1 \to [\alpha]}(s_{\alpha-1});
\end{align} 
also, for $\alpha = 0$, the future-imposed condition about time period $\alpha$ requires that
\begin{align}\label{eq:two-point-condition-special-case-in-multi-period}
R_{\star \leadsto 0}(s_0)  \geq L_{\star\leadsto 0}(s_0).
\end{align}
We discuss the above conditions according to the following cases.
\begin{enumerate}
\item $\alpha=0$: Eq.~\eqref{eq:two-point-condition-special-case-in-multi-period} automatically holds.
\item $\alpha\in[T]$: Denote $\triangle_0\defeq\ovld_0-\unld_0$ for convenience. Eq.~\eqref{eq:two-point-condition-in-multi-period} corresponds to that for any interval $[a,b]\subset[\unld_0,\ovld_0]$ with $b-a\leq\triangle_{\alpha-1}$ (referring to $s_{\alpha-1}\in F_{0 \to \alpha-1}(s_0)$), it holds that
$    (a+\frac{\Gamma}{c}) \geq b-\frac{\Gamma}{p-c}-\sum_{t=\alpha}^{T} V_t$.
This condition is equivalent to 
$    \Gamma\geq \frac{c\cdot(p-c)}{p}(\triangle_{\alpha-1}-\sum_{t=\alpha}^{T}V_t)$.
(Note that when $\alpha = 1$, because 
$\triangle_0\leq\ovld_0\leq\sum\limits_{t=1}^{T}V_t$, the corresponding condition automatically holds.)
\item $\alpha=T+1$: Eq.~\eqref{eq:two-point-condition-in-multi-period} corresponds to that for any interval $[a,b]\subset[\unld_0,\ovld_0]$ with $b-a\leq\triangle_{T}$ (referring to $s_{T}\in F_{0\to T}(s_0)$), it holds that 
$
(a+\frac{\Gamma}{c})+0\geq b-\frac{\Gamma}{p-c}$. This condition is equivalent to 
$\Gamma\geq \frac{c\cdot(p-c)}{p}\cdot \triangle_T$.
\end{enumerate}
Summarizing the above discussion, we conclude that Eq.~\eqref{eq:decision-problem-app-I-robust-regret} holds if and only if 
\[
\Gamma\geq\Gamma^*\defeq \frac{c\cdot (p-c)}{p}\cdot\mathop{\max}\limits_{t\in [T]}\left\{\triangle_{t}-\sum\limits_{s=t+1}^{T}V_{s}\right\}.
\]
Thus the optimal robust regret equals $\Gamma^*$. Moreover, from Corollary~\ref{cor:feasible-policy}, we can derive that for each $t\in\{0,\dots,T\}$, using previous calculation, one may directly verify that the corresponding feasible policy is $\pi_{t}([\unld_t,\ovld_t],x_t)=\min(V_t,\frac{\Gamma^*}{c}+\unld_t-x_t)$.

\subsection{Proof of Proposition~\ref{prop:conclusion-app-I-optimal-robust-competitive-ratio}}
\label{sec:proof-of-app-I-optimal-competitive-ratio}
We denote $\tilde{\Phi} \defeq \frac{(1-\Phi) \cdot  p + \Phi \cdot c}{c}$ and may directly verify that $\Phi\leq 1$ and $\tilde{\Phi}\geq 1$. Thus, the environment-dependent inventory constraints are $L_{T+1}(d)=d\cdot\Phi$ and $R_{T+1}(d)=d\cdot\tilde{\Phi}$. We first calculate the objects introduced in Section~\ref{sec:additional-notations}. An observation is that the multi-time-period environment state transition and the multi-time-period action constraints are the same as in Section~\ref{sec:proofofAssessingbyDifferenceinwarmup}. We compute the other objects as follows:
\begin{enumerate}
\item Future-imposed inventory level constraints: for any $\alpha,\beta\in\{0,\dots,T+1\}$ with $\beta\geq\alpha$, and an environment state $s_\alpha=[a,b]\in S_\alpha$, we have 
\begin{equation*}
   R_{\beta \leadsto \alpha}(s_\alpha) = \left\{
    \begin{array}{lllll}
        0, \qquad &\text{when}~\beta=0, &\qquad \qquad  & a\cdot\tilde{\Phi}, \qquad &\text{when}~\beta=T+1,~\alpha< T+1,\\[5pt]
        +\infty,\qquad &\text{when}~ 0<\beta<T+1, & & d\cdot\tilde{\Phi}, \qquad &\text{when}~\beta=\alpha=T+1,~s_{\alpha}=d. 
    \end{array}
    \right.
\end{equation*}
The case with $\beta=T+1$ and $\alpha<T+1$ is because 
$  
R_{(T+1) \leadsto \alpha}([a,b])=\min_{d\in F_{\alpha\to (T+1)}([a,b])}(d\cdot\tilde{\Phi}-0)=a\cdot\tilde{\Phi},
$
where both equalities hold directly from the definition (the case with $\alpha=\beta=T+1$ is similar). Similarly, we have
\begin{equation*}
   L_{\beta\leadsto\alpha}(s_\alpha) = \left\{
    \begin{array}{lllll}
        0, \qquad &\text{when}~\beta=0,  &&b\cdot \Phi-\sum_{t=\alpha}^{T}V_t, \qquad &\text{when}~\beta=T+1,~\alpha<T+1,\\[5pt]
        -\infty,\qquad &\text{when}~0<\beta<T+1, && d\cdot \Phi,\qquad &\text{when}~\alpha=\beta=T+1,~s_\alpha=d.
    \end{array} \right.
\end{equation*}
We can also derive that 
\begin{align*}
    &R_{\star \leadsto 0}([a,b]) = 0,\qquad 
    L_{\star \leadsto 0}([a,b]) = 0,\qquad 
    R_{\star \leadsto (T{+}1)}(d) = d \cdot \tilde{\Phi},\qquad 
    L_{\star \leadsto (T{+}1)}(d) = d \cdot \Phi, \\
    &R_{\star \leadsto \alpha}([a,b]) = a \cdot \tilde{\Phi},\qquad 
    L_{\star \leadsto \alpha}([a,b]) = b \cdot \Phi - \sum_{t = \alpha}^T V_t, 
    \qquad \forall \alpha \in \{1,\dots,T\}.
\end{align*}
\item Strengthened future-imposed inventory bound: directly from the definition and the discussion for future-imposed inventory bounds, we can derive that for any $\alpha\in\{0,\dots,T\}$ and $[a,b]\subset[\unld_0,\ovld_0]$ with $b-a\leq \triangle_{\alpha}$, it holds that $R_{\star\leadsto\alpha\to[\alpha+1]}([a,b])=a\cdot\tilde{\Phi}$ and $L_{\star\leadsto\alpha\to[\alpha+1]}([a,b])=b\cdot\Phi-\sum_{t=\alpha+1}^T V_t$.
\end{enumerate}

By Lemma~\ref{lem:app-I-reduction-competitive-ratio} and Theorem~\ref{thm:main-equivalent-conditions}, Eq.~\eqref{eq:decision-problem-app-I-robust-comepetive-ratio} holds if and only if the following future-imposed conditions are satisfied: for any two time periods $\alpha\in \{1,\dots,T+1\}$, the \emph{future-imposed conditions} require that for each $s_{\alpha-1} \in F_{0 \to \alpha-1}(s_0)$, it holds that
\begin{align} \label{eq:two-point-condition-in-multi-period-new}
R_{\star \leadsto \alpha -1 \to [\alpha]}(s_{\alpha-1}) \geq L_{\star \leadsto \alpha -1 \to [\alpha]}(s_{\alpha-1}); 
\end{align} 
also, for $\alpha = 0$, the future-imposed condition about time periods $\alpha$ requires that
\begin{align}\label{eq:two-point-condition-special-case-in-multi-period-new}
R_{\star \leadsto 0}(s_0)  \geq L_{\star\leadsto 0}(s_0).
\end{align}
We discuss the above conditions according to the following cases.
\begin{enumerate}
\item $\alpha=0$: Eq.~\eqref{eq:two-point-condition-special-case-in-multi-period-new} automatically holds.
\item $\alpha\in[T]$: Eq.~\eqref{eq:two-point-condition-in-multi-period-new} corresponds to that for any interval $[a,b]\subset[\unld_0,\ovld_0]$ with $b-a\leq\triangle_{\alpha-1}$ (referring to $s_{\alpha-1}\in F_{0 \to \alpha-1}(s_0)$), it holds that 
$
    a\cdot\tilde{\Phi}\geq b\cdot\Phi-\sum_{t=\alpha}^{T} V_t.
$
Noticing $\tilde{\Phi}\geq 1\geq\Phi$, this condition is equivalent to 
$
    \unld_0\cdot\tilde{\Phi}+\sum_{t=\alpha}^{T} V_t \geq (\unld_0+\triangle_{\alpha-1})\cdot\Phi.
$
Combining the definition that $\tilde{\Phi}\defeq[(1-\Phi)\cdot p+\Phi\cdot c)]/c$, we have 
$
\Phi\leq \frac{p \cdot \unld_0 + c \cdot \sum_{t=\alpha}^{T} V_t}{p \cdot \unld_0 + c \cdot \triangle_{\alpha-1}}.    
$
(Note that when $\alpha = 1$, because 
$\triangle_0\leq\ovld_0\leq\sum\limits_{t=1}^{T}V_t$, the corresponding condition automatically holds.)
\item $\alpha=T+1$: Eq.~\eqref{eq:two-point-condition-in-multi-period-new} corresponds to that for any interval $[a,b]\subset[\unld_0,\ovld_0]$ with $b-a\leq\triangle_{T}$ (referring to $s_{T}\in F_{0\to T}(s_0)$), it holds that 
$
a\cdot\tilde{\Phi}+0 \geq b\cdot\Phi.
$
This condition is equivalent to 
$
\Phi\leq \frac{p \cdot \unld_0 }{p \cdot \unld_0 + c \cdot \triangle_T}$.
\end{enumerate}
Summarizing the above discussion, we conclude that Eq.~\eqref{eq:decision-problem-app-I-robust-comepetive-ratio} holds if and only if 
\[
\Phi \leq \Phi^{*}\defeq \mathop{\min}\limits_{t\in [T]}\left\{\frac{p \cdot \unld_0 + c \cdot \sum_{s=t+1}^{T} V_s}{p \cdot \unld_0 + c \cdot \triangle_t}\right\}.
\]
Thus the optimal robust competitive ratio equals $\Phi^*$. Moreover, from Corollary~\ref{cor:feasible-policy}, using previous computation, one may directly verify that for each $t\in\{0,\dots,T\}$, the corresponding feasible policy is $\pi_{t}([\unld_t,\ovld_t],x_t)=\min\left\{V_t,\unld_t \cdot \frac{(1 - \Phi^*) \cdot p + \Phi^* \cdot c}{c} - x_t\right\}$.

\subsection{An Example where Single-switching Sequences Fail to Minimize the Competitive Ratio}
\label{sec:counterexample-for-app-I-robust--competitive-ratio}
We consider the following instance of the robust multi-period ordering with predictions problem:
\begin{enumerate}
\item $T=3$, with total time horizon $T+1=4$.
\item The initial prediction interval is $[\unld_0,\ovld_0]=[1,7]$.
\item The daily upper bounds on the lengths of prediction intervals are $\triangle_1=5$, $\triangle_2=4$, and $\triangle_3=1$.
\item The ordering cost $c=1$ and the unit revenue $p=2$.
\item The daily supply capacities are $V_1=4$, $V_2=2$, and $V_3=1$.
\end{enumerate}

The ``single-switching sequences'' idea proposed in ~\citet{feng2024robust} restricts the adversary to a small subset of single-switching prediction sequences $\{\mathcal{P}^{(k)}\}_{k\in\{0,\dots,T\}}$, where $\mathcal{P}^{(k)}=\{[\unld_t^{(k)},\ovld_{t}^{(k)}]\}_{t\in[T]}$ is defined as follows:
\begin{align*}
    &\text{when~} t\in[k]:\qquad &&[\unld_t^{(k)},\ovld_t^{(k)}]= [\ovld_{t-1}^{(k)}-\triangle_t, \ovld_{t-1}^{(k)}], \\
    &\text{when~} t\in[k+1,T]: \qquad &&[\unld_t^{(k)},\ovld_t^{(k)}]= [\unld_{t-1}^{(k)}, \unld_{t-1}^{(k)}+\triangle_t],
\end{align*}
where we denote $[\unld_0^{(k)},\ovld_0^{(k)}]=[\unld_0,\ovld_0]$. In our instance, the set of single-switching prediction sequences consists of 
\begin{align*}
P^{(0)}=\{[1,6],~[1,5],~[1,2]\},&\qquad & P^{(1)}=\{[2,7],~[2,6],~[2,3]\}, \\
P^{(2)}=\{[2,7],~[3,7],~[3,4]\}, &\qquad & P^{(3)}=\{[2,7],~[3,7],~[6,7]\}.
\end{align*}

\begin{figure}[t]
\begin{center}
\includegraphics[width=0.6\linewidth]{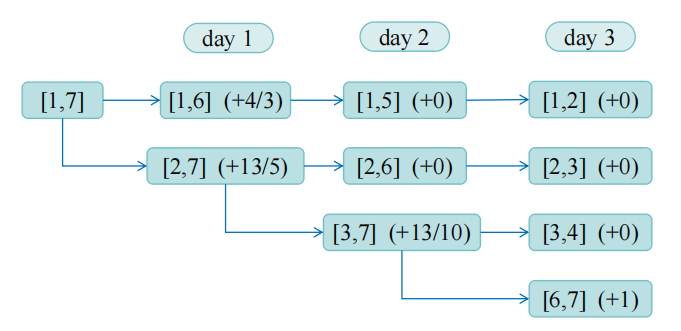}
\end{center}
\caption{Illustration of an ordering policy under the single-switching prediction sequences.}
\label{fig:single-switching-counter-example-policy}
\end{figure}

When the adversary is restricted to the single-switching prediction sequences, we show that there is an ordering policy achieving a robust competitive ratio of $2/3$, which is illustrated in Figure~\ref{fig:single-switching-counter-example-policy}. In the figure, the intervals represent the prediction intervals revealed to the decision-maker on each preparation day, and the numbers in parentheses represent the policy's action (the ordering quantity).

 Below, we discuss the four possible single-switching prediction sequences chosen by the adversary, and show that in each case, the competitive ratio of the proposed policy is at least $2/3$.
\begin{itemize}
\item When the adversary chooses $\mathcal{P}^{(0)}$: the final inventory level is $4/3$, and the competitive ratio is at least 
$
\min_{d \in [1, 2]} \left\{\frac{2\cdot\min(4/3,d)-4/3}{d}\right\} = \frac{2}{3}$.
\item When the adversary chooses $\mathcal{P}^{(1)}$: the final inventory level is $13/5$, and the competitive ratio is at least 
$\min_{d \in [2, 3]} \left\{\frac{2\cdot\min(13/5,d)-13/5}{d}\right\} = \frac{7}{10} \geq \frac{2}{3}$.
\item When the adversary chooses $\mathcal{P}^{(2)}$: the final inventory level is $39/10$, and the competitive ratio is at least 
$
\min_{d \in [3, 4]} \left\{\frac{2\cdot\min(39/10,d)-39/10}{d}\right\} = \frac{7}{10} \geq \frac{2}{3}$.
\item When the adversary chooses $\mathcal{P}^{(3)}$: the final inventory level is $49/10$, and the competitive ratio is at least 
$\min_{d \in [6, 7]} \left\{\frac{2\cdot\min(49/10,d)-49/10}{d}\right\} = \frac{7}{10} \geq \frac{2}{3}$.
\end{itemize}

On the other hand, by Proposition~\ref{prop:conclusion-app-I-optimal-robust-competitive-ratio}, we directly calculate that the robust competitive ratio of this problem instance is $\Phi^* = 1/3 < 2/3$, implying that restricting the adversary to single-switching prediction sequences hurts the adversary's power to minimize the competitive ratio.

\subsection{Handling Scale-dependent Prediction Sequences}
\label{sec:app-I-non-uniform-interval-lengths}

Our Assumption~\ref{asp:regularityofpredictionsequence}, inherited from \citet{feng2024robust}, imposes a uniform length upper bound ($\triangle_t$) on the lengths of all potential prediction intervals on day $t$. However, in practice, these intervals may have non-uniform lengths. Relaxing Assumption~\ref{asp:regularityofpredictionsequence} to allow more refined, non-uniform upper bounds within a day could enhance the performance of our robust algorithms. Our minimax-MDP framework naturally accommodates this non-uniformity, as the environment state can encode any prediction interval without a fixed length constraint. In this subsection, we introduce one such practically motivated relaxation --- namely the \emph{scale-dependent prediction sequences} --- and demonstrate how our minimax-MDP handles this scenario efficiently.

\medskip
\noindent\uline{The scale-dependent prediction sequences.} A larger-scaled quantity is inherently more difficult to predict within a fixed-length prediction interval. For example, during a rocket launch, measurements taken within the first few seconds may be accurate to the nearest centimeter, whereas after a few minutes, as the distance increases, the accuracy may be limited to meters. Similarly, predicting the demand of a wealthier group with fixed precision is more difficult than for a less affluent group, as higher expenditure levels often correspond to greater variability in demand. 

In light of this, in the RMOwP problem, if on a given day, the demand is predicted to be relatively smaller in scale, it is reasonable to assume that the decision-maker will receive a more accurate prediction, corresponding to a shorter prediction interval. Formally, we introduce the following assumption about the prediction interval sequences.

\begin{assumption}\label{asp:scale-dependent-prediction-sequence}
Given a sequence of functions $\{\delta_t : \mathbb{R}_{\geq 0} \times \mathbb{R}_{\geq 0} \to \mathbb{R}_{\geq 0}\}_{t\in[T]}$ and an interval $[\unld_0,\ovld_0]$, an interval sequence $\mathcal{P}=\{\mathcal{P}_t\}_{t\in[T]}$ is a \emph{scale-dependent prediction sequence} with respect to $\{[\unld_0,\ovld_0],\{\delta_t\}_{t\in[T]}\}$, if it satisfies the consistency property in Assumption~\ref{asp:regularityofpredictionsequence}, and for each $t\in[T]$, the prediction interval $\mathcal{P}_t=[\unld_t,\ovld_t]$ satisfies the \emph{scale-dependent prediction error} property: the prediction interval length $\ovld_t-\unld_t\leq \delta_t(\unld_{t-1}, \ovld_{t-1})$. 
\end{assumption}
The regular prediction sequence (Assumption~\ref{asp:regularityofpredictionsequence}) is a special case of the scale-dependent prediction sequence when $\delta_t(\cdot, \cdot)\equiv \triangle_t$.  The scale-dependent prediction error in Assumption~\ref{asp:scale-dependent-prediction-sequence} characterizes the relation between the prediction error and the historically known bounds about the demand. 

As an illustration of Assumption~\ref{asp:scale-dependent-prediction-sequence}, consider the scenario where the final demand $d$ is estimated as the median of independent samples drawn from a log-normal distribution $\mathrm{Lognormal}(d, \sigma^2)$, representing the outcome of independent surveys (see, e.g., \citep{aitchison1969lognormal}). For simplicity, suppose $\sigma$ (or an upper bound of $\sigma$) is known. Let $\hat{d}$ be the maximum likelihood estimate based on $n$ samples. The confidence interval for $d$ takes the form 
\[
d  \in \left[\exp\left(\hat{d} - c \cdot \sigma  / {\sqrt{n}}\right), \exp\left(\hat{d} + c \cdot {\sigma} / {\sqrt{n}}\right)\right],
\]
where $c$ is an appropriate $\mathcal{N}(0, 1)$ percentage point. Given the historical prediction $d \leq \ovld_{t-1}$, we may set 
\begin{align}\label{eq:lognormal-CI-length}
\delta_t(\unld_{t-1}, \ovld_{t-1}) = \ovld_{t-1} \times \left[\exp\left(c \sigma / \sqrt{n_t}\right) - \exp\left(-c \sigma /\sqrt{n_t}\right)\right],
\end{align}
where $n_t$ denotes the number of training samples available to the decision-maker by day $t$.

\medskip
\noindent\uline{Reduction to minimax-MDPs.} Under Assumption~\ref{asp:scale-dependent-prediction-sequence}, an instance of the RMOwP problem can be defined as
\begin{align} \label{eq:robust-multi-period-ordering-decision-predictions-scale-dependent-defn}
    \mathcal{K}\defeq\{T,[\unld_0,\ovld_0],\{\delta_t(\cdot,\cdot)\}_{t\in[T]},c,p,\{V_t\}_{t\in[T]}\},
\end{align}
where the only difference from the original definition is that $\{\triangle_t\}_{t \in [T]}$ in Eq.~\eqref{eq:definition-of-robust-multi-period} is replaced with $\{\delta_t(\cdot,\cdot)\}_{t\in[T]}$ as the parameter for the scale-dependent prediction sequence $\mathcal{P}$. For a policy $\pi$, we can similarly define the robust regret $\mathrm{Reg}_\pi(\mathcal{K})$ and robust competitive ratio $\phi_\pi(\mathcal{K})$.

We then reduce the decision problems $\inf_\pi\{\mathrm{Reg}_\pi(\mathcal{K})\} \leq \Gamma$ and $\sup_\pi\{\phi_\pi(\mathcal{K})\} \geq \Phi$ to deciding the feasibility of minimax-MDPs. The reductions are similar to that described in Section~\ref{sec:app-I-robust-regret} and Section~\ref{sec:app-I-robust-competitive-ratio}, where the only difference lies in the environment state transition functions, which are set as follows:
\begin{align*}
F_{t}([a,b])&=\{[e,f];~0\leq f-e\leq \delta_t(a, b),~[e,f]\subseteq [a,b]\}, ~~~~ \forall t\in\{0,1,\dots,T-1\},~\\
F_{T}([a,b])&=\{d;~d\in[a,b]\}.
\end{align*}

Following the same lines as in the proofs of Lemmas~\ref{lem:app-I-reduction} and \ref{lem:app-I-reduction-competitive-ratio}, we establish that $\inf_\pi\{\mathrm{Reg}_\pi(\mathcal{K})\} \leq \Gamma$ and $\sup_\pi\{\phi_\pi(\mathcal{K})\} \geq \Phi$ hold if and only if the corresponding minimax-MDP obtained through the reduction is feasible.

\medskip
\noindent\uline{Example of scale-dependent sequences: scale-linear prediction intervals.} To illustrate how our minimax-MDP framework efficiently addresses the scale-dependent prediction sequences, we consider the case where the prediction interval lengths are linear with the scale of the demand:
\begin{align}\label{eq:RMOwP-scal-linear-defn}
\delta_t(\unld_{t-1}, \ovld_{t-1}) = a_t \ovld_{t-1} + b_t,
\end{align}
where $\{a_t, b_t\}_{t \in [T]}$ are given as problem parameters. These \emph{scale-linear prediction intervals} encompass the lognormal-based estimation example described in Eq.~\eqref{eq:lognormal-CI-length}, where we may set $a_t = \exp\left(c \sigma / \sqrt{n_t}\right) - \exp\left(-c \sigma /\sqrt{n_t}\right)$ and $b_t = 0$. The following proposition shows how our future-imposed conditions derive a closed-form solution to the optimal robust competitive ratio.

\begin{proposition}
\label{prop:competitive-ratio-of-scale-linear-prediction-intervals}
Consider the scale-dependent RMOwP problem in Eq.~\eqref{eq:robust-multi-period-ordering-decision-predictions-scale-dependent-defn} with scale-linear interval lengths (Eq.~\eqref{eq:RMOwP-scal-linear-defn}). Without loss of generality, we assume $\{a_t\}$ and $\{b_t\}$ are non-negative, non-increasing sequences, and $\ovld_0-\unld_0\geq a_1\cdot \ovld_0+b_1$.  The optimal robust competitive ratio equals to 
\begin{align*}
    \min_{\tau \in [T], e\in[\unld_0,\ovld_0]} \left\{ \frac{c\cdot \sum_{i=\tau+1}^T V_i+p\cdot e}{(p-c)\cdot e+c\cdot g_\tau(e)} \right\},
\end{align*}
where $g_\tau(e)$ denotes the largest possible right endpoint of a prediction interval with left endpoint $e$ at time $\tau$, i.e., $g_\tau(e) = \max \left\{f: [e, f] \in F_{0 \to \tau}([\unld_0,\ovld_0])\right\}$, where, recall that $F_{0 \to \tau}(\cdot)$ is defined in Eq.~\eqref{eq:def-F-a-to-b} based on Eq.~\eqref{eq:robust-multi-period-ordering-decision-predictions-scale-dependent-defn}.
Moreover, the function $g_\tau(\cdot)$ is piecewise linear with at most $T+1$ pieces. Both $g_\tau(\cdot)$ and the resulting optimal robust competitive ratio admit closed-form expressions, which are detailed in Section~\ref{sec:proof-of-competitive-ratio--of-scale-linear-prediction-intervals}.
\end{proposition}

\subsection{Proof of Proposition~\ref{prop:competitive-ratio-of-scale-linear-prediction-intervals}}
\label{sec:proof-of-competitive-ratio--of-scale-linear-prediction-intervals}
Notice that $F_{\alpha\to T+1}([a,b])=\{d;d\in[a,b]\}$. Following the same proof in Section~\ref{sec:proof-of-app-I-optimal-competitive-ratio}, the feasible policy exists if and only if for any $1\leq\tau\leq T$ and $[e,f]\in F_{0\to \tau}([\unld_0,\ovld_0])$, we have
\begin{align}
    e\cdot \tilde{\Phi}+\sum_{i=\tau+1}^T V_i \geq f\cdot \Phi. \label{eq:scale-dependent-decision-phi}
\end{align}
Next, we calculate $F_{0 \to \tau }([\unld_0,\ovld_0])$. We may directly verify that
\[
F_{0 \to \tau}([\unld_0,\ovld_0])=\{[e,f];~\unld_0\leq e\leq \ovld_0 \wedge e\leq f \leq g_\tau(e)\}.
\]
We then calculate the explicit expression of $g_\tau(e)$. Consider the sequence $\{f_{i}(e)\}_{i\in[\tau]}$ defined as 
\[
    f_{1}(e)=\min(\ovld_0, e+a_1\cdot\ovld_0 + b_1), \qquad 
    f_{i}(e)=\min(\ovld_0, e+a_i\cdot f_{i-1}(e)+ b_i), \qquad i\in\{2,\dots,\tau\}.
\]
One may directly verify that $[a,f_{\tau}(e)]\in F_{0 \to \tau}([\unld_0,\ovld_0])$ by noticing that $\{[e,f_{i}(e)]\}_{i\in[\tau]}$ is a scale-dependent prediction sequence, which indicates that $g_\tau(e)\geq f_{\tau}(e)$.

Then we prove that $g_\tau(e)\leq f_{\tau}(e)$: consider any regular sequence $\{e'_i,f'_i\}_{i=0}^\tau$ with $[e'_0,f'_0]=[\unld_0,\ovld_0]$ and $e'_\tau=a$, it suffices to show that $f_{\tau}(e) \geq f'_\tau$. We prove this inequality by showing that $f_{i}(e)\geq f'_i, \forall i\in[\tau]$ via the following induction:
\begin{itemize}
    \item Induction basis: when $i=1$, we have $f'_1\leq \ovld_0$ and 
    $
    f'_1 \leq e'_1+a_1\cdot\ovld_0+b_1 \leq e+a_1\cdot\ovld_0+b_1$,
    which indicates that $f'_1\leq f_{1}(e) $.
    \item Induction step: if $f_{i}(e)\geq f'_i$ holds when $i=j-1$, then we have $f'_{j}\leq \ovld_0$ and 
    $
    f'_j \leq e'_j+a_j\cdot f'_{j-1}+b_j \leq e+ a_j\cdot f_{j-1}(e)+b_j$,
    which indicates that $f'_i \leq f_{i}(e)$ holds for $i = j$.
\end{itemize}

Combining the above discussions, we have $g_\tau(e)=f_{\tau}(e)$. Before calculating $f_{\tau}(e)$, we prove a property that $f_{i}(e)\leq f_{i-1}(e), \forall i : 2\leq i\leq\tau$ via the following induction:
\begin{itemize}
    \item Induction basis: when $i=2$, we have $f_{2}(e)\leq \ovld_0$ and 
    $
    f_{2}(e) \leq e+ a_2\cdot f_{1}(e)+b_2 \leq e+ a_1\cdot\ovld_0+b_1$, 
    which indicates that $f_{2}(e)\geq f_{1}(e)$.
    \item Induction step: if $f_{i}(e)\leq f_{i-1}(e)$ holds for some $i \geq 2$, we have $f_{i+1}(e)\leq \ovld_0$ and 
    $
    f_{i+1}(e)\leq e+ a_{i+1}\cdot f_{i}(e)+b_{i+1}\leq e+a_i\cdot f_{i-1}(e)+b_i,
    $
    which indicates that $f_{i+1}(e)\leq f_{i}(e)$.
\end{itemize}

Finally, we demonstrate how to calculate the value $f_{\tau}(e)$. We set $a_0=1$ and $b_0=0$ for convenience, and consider the set $A_\tau(e)\defeq\{t\in[\tau];e+a_t\cdot\ovld_0+b_t\leq \ovld_0\}$. We discuss two situations when $A_\tau(e)\neq\emptyset$ and $A_\tau(e)=\emptyset$ as follows:
\begin{enumerate}
    \item $A_\tau(e)\neq\emptyset$: then there exists $t\in[\tau]$ such that 
$
    e+a_{t-1}\cdot \ovld_0+b_{t-1} \geq \ovld_0$ and $e+a_t\cdot\ovld_0+b_t\leq \ovld_0. 
$
Due to the condition that $\{a_i\}$ and $\{b_i\}$ are both decreasing and non-negative, one may directly verify that 
\[
    f_{i}(e)=\ovld_0,\qquad 1\leq i\leq t-1,\qquad\qquad
    f_{j+1}(e)=e+a_{j+1}\cdot f_j(e)+b_{j+1},\qquad t-1\leq j \leq \tau-1. 
\]
We may directly verify that 
$
f_\tau=e\cdot D_1(\tau,t+1)+D_2(\tau,t),
$
where
\begin{align*}
    &D_1(\tau,t+1)=
    \begin{cases}
        1+\sum_{i=t+1}^\tau \prod_{j=i}^\tau a_j &\qquad t\leq \tau-1 \\[10pt]
        1 &\qquad t=\tau
    \end{cases},
    \\
    &D_2(\tau,t)=
    \begin{cases}
        \ovld_0\cdot \prod_{i=t}^{\tau} a_i  + \sum_{i=t}^{\tau-1}\left( b_i\cdot\prod_{j=i+1}^\tau a_i \right)+ b_\tau &\qquad t\leq \tau-1\\[10pt]
        \ovld_0\cdot a_\tau + b_\tau &\qquad t=\tau
    \end{cases}.
\end{align*}
\item $A_\tau(e)=\emptyset$: then $e+a_\tau\cdot \ovld_0+b_\tau > \ovld_0$, and we have $f_i(e)=\ovld_0$ for all $i\in[\tau]$.
\end{enumerate}

Summarizing the above discussion, we can derive the explicit form of $g_\tau(e)$ as
\begin{align*}
    g_\tau(e)=
    \begin{cases}
        e\cdot D_1(\tau,2)+D_2(\tau,1),\qquad &\unld_0\leq e\leq \ovld_0\cdot(1-a_1)-b_1, \\[5pt]
        e\cdot D_1(\tau,t+1)+D_2(\tau,t),\qquad &~2\leq t\leq \tau,~\ovld_0\cdot(1-a_{t-1})-b_{t-1}\leq e\leq \ovld_0\cdot(1-a_t)-b_t, \\[5pt]
        \ovld_0,\qquad & e\geq \ovld_0\cdot(1-a_\tau)-b_\tau.
    \end{cases}
\end{align*}
We then rewrite Eq.~\eqref{eq:scale-dependent-decision-phi} as
\begin{align}\label{eq:scale-dependent-decision-phi-g-tau-e}
    \sum_{i=\tau+1}^T V_i \geq \Phi\cdot \left[ e\cdot \frac{p-c}{c} + g_\tau(e) \right]-e\cdot\frac{p}{c}.    
\end{align}
Note that $g_\tau(\cdot)$ is a continuous piecewise-linear function, thus Eq.~\eqref{eq:scale-dependent-decision-phi-g-tau-e} holds for all $e\in[\unld_0,\ovld_0]$ if and only if Eq.~\eqref{eq:scale-dependent-decision-phi-g-tau-e} holds on all breakpoints of $g_\tau(\cdot)$. Another fact is that the right-hand side of Eq.~\eqref{eq:scale-dependent-decision-phi-g-tau-e} is strictly decreasing when $e\geq \ovld_0\cdot(1-a_\tau)-b_\tau$, thus we only need to verify the following segment points: $\unld_0$ and $\ovld_0\cdot (1-a_t)-b_t$ for all $t\in[\tau]$.
Inputting these values, we may directly verify that Eq.~\eqref{eq:scale-dependent-decision-phi} holds if and only if for any $1\leq \tau\leq T$, we have that $\Phi$ is smaller than all the following values: 
\begin{align}
    &\frac{c\cdot\sum_{i=\tau+1}^T V_i+p\cdot\left[ \ovld_0\cdot (1-a_t) -b_t \right]}{ \left[ \ovld_0\cdot (1-a_t) -b_t \right]\cdot \left[ p-c + c\cdot D_1(\tau,t+1) \right] + c\cdot D_2(\tau,t) }, \qquad \forall 1\leq t\leq \tau, \\
    &\frac{c\cdot\sum_{i=\tau+1}^T V_i + p \cdot \unld_0}{\unld_0\cdot \left[ p - c + c\cdot D_1(\tau,2) \right] + c\cdot D_2(\tau,1)}.
\end{align}

\section{Omitted Proofs in Section~\ref{sec:robust-resource-allocation-with-predictions}}
\subsection{Proof of Proposition~\ref{prop:app-II-robust-competitive-ratio}}
\label{sec:proof-of-robust-resource-allocation-with-predictions}
Using the construction in Section~\ref{sec:app-II-reducing-to-minimax-MDP}, we compute the following objects:
\begin{enumerate}
\item Multi-time-period environment state transition: for any $\alpha,\beta\in\{0,\dots,T+1\}$ with $\alpha<\beta$, and an environment state $s_\alpha=([a_i,b_i])_{i\in[d]}\in S_\alpha$, the value of $F_{\alpha\to\beta}(s_\alpha)$ is 
\begin{equation*}
    \begin{cases}
        \{([e_i,f_i])_{i\in[d]};~\forall i\in[d],~0\leq f_i-e_i\leq \triangle_{\beta,i}\},\qquad &\beta<T+1, \\[5pt]
        \{(\theta_i)_{i\in[d]}; ~\forall i\in[d],~ \theta_i\in[a_i,b_i]\}, \qquad &\beta=T+1.
    \end{cases}
\end{equation*}
\item Multi-time-period action constraints: for any $\alpha,\beta\in\{0,\dots,T+1\}$ with $\alpha<\beta$, we have 
\begin{align*}
    U_{\alpha\to\beta}(\cdot,\cdot)\equiv 0,\qquad\qquad
    V_{\alpha\to\beta}(\cdot,\cdot)\equiv \sum_{t=\alpha}^{\beta-1}V_t .
\end{align*}
\item Future-imposed inventory level constraints: for any $\alpha,\beta\in\{0,\dots,T+1\}$ with $\beta\geq\alpha$, and an environment state $s_\alpha\in S_\alpha$, we have 
\begin{equation*}
   R_{\beta \leadsto \alpha}(s_\alpha)= \begin{cases}
        0, \qquad &\beta=0, \\[5pt]
        +\infty,\qquad &0<\beta<T+1, \\[5pt]
        \min\limits_{\forall i\in[d], \theta_i\in[a_i,b_i]}R_{T+1}((\theta_i)_{i\in[d]};\mathcal{M}), \qquad &\beta=T+1,~\alpha< T+1, ~s_\alpha=([a_i,b_i])_{i\in[d]}, \\[5pt]
        R_{T+1}((\theta_i)_{i\in[d]};\mathcal{M}), \qquad &\beta=\alpha=T+1,~s_{\alpha}=(\theta_i)_{i\in[d]}.
    \end{cases}
\end{equation*}
Similarly, we have that $L_{\beta\leadsto\alpha}(s_\alpha) = $
\begin{equation*}
     \begin{cases}
        0, \qquad &\beta=0, \\[5pt]
        -\infty, \qquad &0<\beta<T+1, \\[5pt]
        \max\limits_{\forall i\in[d], \theta_i\in[a_i,b_i]}L_{T+1}((\theta_i)_{i\in[d]};\mathcal{M})-\sum_{t=\alpha}^T V_t, \qquad &\beta=T+1,~\alpha< T+1, ~s_\alpha=([a_i,b_i])_{i\in[d]}, \\[5pt]
        L_{T+1}((\theta_i)_{i\in[d]};\mathcal{M}), \qquad &\beta=\alpha=T+1,~s_{\alpha}=(\theta_i)_{i\in[d]}.
    \end{cases}
\end{equation*}
We then derive that 
\begin{align*}
    &R_{\star \leadsto 0}(([a_i,b_i])_{i\in[d]})=0,&& L_{\star \leadsto 0}(([a_i,b_i])_{i\in[d]})=0,\\
    &R_{\star \leadsto (T+1)}((\theta_i)_{li\in[d]})=R_{T+1}((\theta_i)_{i\in[d]};\mathcal{M}),&& L_{\star \leadsto (T+1)}((\theta_i)_{i\in[d]})=L_{T+1}((\theta_i)_{i\in[d]};\mathcal{M});
\end{align*}
also, for $\alpha\in[T]$, we have
\begin{align*}    
    &R_{\star \leadsto \alpha}(([a_i,b_i])_{i\in[d]})=\min\limits_{\forall i\in[d], \theta_i\in[a_i,b_i]}R_{T+1}((\theta_i)_{i\in[d]};\mathcal{M}), \\
    &L_{\star \leadsto \alpha}(([a_i,b_i])_{i\in[d]})=\max\limits_{\forall i\in[d], \theta_i\in[a_i,b_i]}L_{T+1}((\theta_i)_{i\in[d]};\mathcal{M})-\sum_{t=\alpha}^T V_t. 
\end{align*}
\item Strengthened future-imposed inventory bound: directly from the definition and the discussion for future-imposed inventory upper bounds, we can derive that for any $\alpha\in\{0,\dots,T\}$, $[a_i,b_i]\subset[\underline{\theta}_{0, i}, \overline{\theta}_{0, i}]$ and $0\leq b_i-a_i\leq \triangle_{\alpha,i}$ for each $i$, it holds that
\begin{align}
    &R_{\star\leadsto\alpha\to[\alpha+1]}(([a_i,b_i])_{i\in[d]})=\min\limits_{\forall i\in[d], \theta_i\in[a_i,b_i]}R_{T+1}((\theta_i)_{i\in[d]};\mathcal{M}), \notag \\
    &L_{\star\leadsto\alpha\to[\alpha+1]}(([a_i,b_i])_{i\in[d]})=\max\limits_{\forall i\in[d], \theta_i\in[a_i,b_i]}L_{T+1}((\theta_i)_{i\in[d]};\mathcal{M})-\sum_{t=\alpha+1}^T V_t. \notag
\end{align}
\end{enumerate}

By Lemma~\ref{lem:app-II-reduction} and Theorem~\ref{thm:main-equivalent-conditions}, Eq.~\eqref{eq:decision-problem-app-II-robust-comepetive-ratio} holds if and only if the following future-imposed conditions are satisfied: for any two time periods $\alpha\in \{1,\dots,T+1\}$, the \emph{future-imposed condition} requires that for each $s_{\alpha-1} \in F_{0 \to \alpha-1}(s_0)$, it holds that
\begin{align} \label{eq:two-point-condition-in-app-II}
R_{\star \leadsto \alpha -1 \to [\alpha]}(s_{\alpha-1})  \geq L_{\star \leadsto \alpha -1 \to [\alpha]}(s_{\alpha-1}); 
\end{align} 
also, for $\alpha = 0$, the future-imposed condition about time periods $\alpha$ requires that
\begin{align}\label{eq:two-point-condition-special-case-in-app-II}
R_{\star \leadsto 0}(s_0) \geq L_{\star\leadsto\alpha}(s_\alpha).
\end{align}
We discuss the above conditions according to the following cases.
\begin{enumerate}
\item $\alpha=0$: Eq.~\eqref{eq:two-point-condition-special-case-in-app-II} automatically holds.

\item $\alpha\in[T]$: Eq.~\eqref{eq:two-point-condition-in-app-II} correspond to that for each $i$, for any $[a_i,b_i]\subset[\underline{\theta}_{0, i}, \overline{\theta}_{0, i}]$ with $b_i-a_i\leq\triangle_{\alpha-1,i}$ (referring to $s_{\alpha-1}\in F_{0 \to \alpha-1}(s_0)$), it holds that
\begin{align}\label{eq:app-II-proof-min-R-plus-sum-bigger-L}
    \min\limits_{\forall i\in[d],\theta'_i\in[a_i,b_i]}R_{T+1}((\theta'_i)_{i\in[d]};\mathcal{M}) \geq  \max\limits_{\forall i\in[d],\theta''_i\in[a_i,b_i]}L_{T+1}((\theta''_i)_{i\in[d]};\mathcal{M})-\sum_{t=\alpha}^{T} V_t,
\end{align}
which is equivalent to the condition that for any $\bm{\theta}=(\theta_i)_{i\in[d]},\bm{\theta}'=(\theta'_i)_{i\in[d]}$ such that $\theta_i,\theta'_i\in[\underline{\theta}_{0, i}, \overline{\theta}_{0, i}]$ and $|\theta_i-\theta'_i|\leq \triangle_{\alpha-1,i}$ for each $i$, we have 
$
R_{T+1}(\bm{\theta};\mathcal{M})+\sum_{t=\alpha}^{T}V_t\geq L_{T+1}(\bm{\theta}';\mathcal{M}).
$
(Note that when $\alpha=1$, due to Observation~\ref{obs:app-II-reduce-equivalent-in-selling-stage}, the corresponding condition automatically holds.)
\item $\alpha=T+1$: Eq.~\eqref{eq:two-point-condition-in-app-II} corresponds to that for each $i$, for any interval $[a_i,b_i]\subset[\underline{\theta}_{0, i}, \overline{\theta}_{0, i}]$ with $b_i-a_i\leq\triangle_{T,i}$ (referring to $s_{T}\in F_{0\to T}(s_0)$), it holds that 
$
\min\limits_{\forall i\in[d],\theta'_i\in[a_i,b_i]}R_{T+1}((\theta'_i)_{i\in[d]};\mathcal{M})\geq \max\limits_{\forall i\in[d],\theta''_i\in[a_i,b_i]}L_{T+1}((\theta''_i)_{i\in[d]};\mathcal{M}),
$
which is equivalent to the condition that for any $\bm{\theta}=(\theta_i)_{i\in[d]}$ and $\bm{\theta}'=(\theta'_i)_{i\in[d]}$ such that $\theta_i,\theta'_i\in[\underline{\theta}_{0, i}, \overline{\theta}_{0, i}]$ and $|\theta_i-\theta'_i|\leq \triangle_{T,i}$ for each $i$, we have 
$
R_{T+1}(\bm{\theta};\mathcal{M})\geq L_{T+1}(\bm{\theta}';\mathcal{M}).
$
\end{enumerate}
Summarizing the above discussion, we complete the proof of Proposition~\ref{prop:app-II-robust-competitive-ratio}.
\hfill\Halmos

\subsection{Proof of Proposition~\ref{prop:best-robust-competitive-ratio-of-Leontief-utility} }
\label{sec:proof-of-best-robust-competitive-ratio-of-Leontief-utility}
For convenience, let us denote $  u_{\mathrm{Leontief}}(x, y; \bm{\theta})$ by $u(x, y; \bm{\theta})$. One may derive that the $\mathrm{opt}(\bm{\theta};\mathcal{M})=V/(\theta_1+\theta_2)$ and the set $\{x\in[0,V];u(x,V-x;\bm{\theta}) \geq \Phi\cdot \mathrm{opt}(\bm{\theta};\mathcal{M})\}$ equals $[V\cdot \Phi\cdot \frac{\theta_1}{\theta_1+\theta_2}, V - V\cdot\Phi \cdot \frac{\theta_2}{\theta_1+\theta_2}]$. Invoking Proposition~\ref{prop:app-II-robust-competitive-ratio}, Eq.~\eqref{eq:decision-problem-app-I-robust-comepetive-ratio} holds if and only if for any $\tau,\theta_1,\theta_2,\theta'_1,\theta'_2$ such that $1\leq\tau\leq T$, $\theta_1,\theta'_1\in[\underline{\theta}_{0, 1}, \overline{\theta}_{0, 1}]$, $\theta_2,\theta'_2\in[\underline{\theta}_{0, 2}, \overline{\theta}_{0, 2}]$ with $|\theta_1-\theta'_1|\leq \triangle_{\tau,1}$ and $|\theta_2-\theta'_2|\leq \triangle_{\tau,2}$, we have 
\begin{align}
V-V\cdot \Phi\cdot \frac{\theta_2}{\theta_1+\theta_2}+\sum_{t=\tau+1}^{T}V_t\geq V\cdot \Phi\cdot \frac{\theta'_1}{\theta'_1+\theta'_2}, \notag
\end{align}
which is equivalent to
\begin{align}
\Phi^{-1}\geq \frac{V}{V+\sum_{t=\tau+1}^{T}V_t} \cdot \left( \frac{\theta_2}{\theta_1+\theta_2} + \frac{\theta'_1}{\theta'_1+\theta'_2} \right). \label{eq:RRAwP-two-point-condition-leontief}
\end{align}

The rest of discussions will focus on the maximum value of right-hand side of Eq.~\eqref{eq:RRAwP-two-point-condition-leontief}. Notice the right-hand side of Eq.~\eqref{eq:RRAwP-two-point-condition-leontief} increases with the decrease of $\theta_1$ and $\theta'_2$. Thus, to obtain the maximum value of the right-hand side, we can assume $\theta_1$ to be $\max(\theta'_1-\triangle_{\tau,1},\underline{\theta}_{0,1})$ and $\theta'_2$ to be $\max(\theta_2-\triangle_{\tau,2},\underline{\theta}_{0,2})$, and it becomes
\[
\frac{V}{V+\sum_{t=\tau+1}^{T}V_t} \cdot \left( \frac{\theta_2}{\max(\theta'_1-\triangle_{\tau,1},\underline{\theta}_{0,1})+\theta_2} + \frac{\theta'_1}{\theta'_1+\max(\theta_2-\triangle_{\tau,2},\underline{\theta}_{0,2})} \right).
\]
This function increases with $\theta_2\in[\underline{\theta}_{0,2},\underline{\theta}_{0,2}+\triangle_{\tau,2}]$ and $\theta'_1\in[\underline{\theta}_{0,1},\underline{\theta}_{0,1}+\triangle_{\tau,1}]$. Thus, we can assume $\theta_2\in[\underline{\theta}_{0,2}+\triangle_{\tau,2},\overline{\theta}_{0,2}]$ and $\theta_1\in[\underline{\theta}_{0,1}+\triangle_{\tau,1},\overline{\theta}_{0,1}]$. Combining the above discussions, we rewrite the right-hand side of Eq.~\eqref{eq:RRAwP-two-point-condition-leontief} as the function of $\theta_1\in[\underline{\theta}_{0,1},\overline{\theta}_{0,1}-\triangle_{\tau,1}]$ and $\theta_2\in[\underline{\theta}_{0,2}+\triangle_{\tau,2},\overline{\theta}_{0,2}]$ as
\begin{align}\label{eq:RRAwP-two-point-condition-leontief-theta-1-2}
    \frac{V}{V+\sum_{t=\tau+1}^{T}V_t} \cdot \left( \frac{\theta_2}{\theta_1+\theta_2}+\frac{\theta_1+\triangle_{\tau,1}}{\theta_1+\theta_2+\triangle_{\tau,1}-\triangle_{\tau,2}} \right).
\end{align}

Next, we prove that the maximum of the right-hand side of Eq.~\eqref{eq:RRAwP-two-point-condition-leontief-theta-1-2} can be obtained only when $\theta_1=\underline{\theta}_{0,1}$ or $\theta_2=\underline{\theta}_{0,2}+\triangle_{\tau,2}$. We consider the function $g(\lambda)$ as 
\[
g(\lambda)=\frac{\lambda\cdot\theta_2}{\lambda\cdot\theta_1+\lambda\cdot\theta_2}+\frac{\lambda\cdot\theta_1+\triangle_{\tau,1}}{\lambda\cdot\theta_1+\lambda\cdot\theta_2+\triangle_{\tau,1}-\triangle_{\tau,2}},
\]
and the derivative of $g$ is
\[
g'(\lambda)= \frac{-\triangle_{\tau,1}\cdot\theta_2-\triangle_{\tau,2}\cdot\theta_1}{\left( \lambda\cdot\theta_1+\lambda\cdot\theta_2-\triangle_{\tau,2}+\triangle_{\tau,1}\right)^2}\leq 0.
\]
Thus, shrinking both variables by the same ratio results in the increase of the value. 

We assume that $\theta_1=\underline{\theta}_{0,1}$ or $\theta_2=\underline{\theta}_{0,2}+\triangle_{\tau,2}$ and discuss these two cases as follows.
\begin{enumerate}
\item $\theta_1=\underline{\theta}_{0,1}$: consider the function $f(\theta_2)$ as 
\[
f(\theta_2)=\frac{\theta_2}{\underline{\theta}_{0,1}+\theta_2}+\frac{\underline{\theta}_{0,1}+\triangle_{\tau,1}}{\underline{\theta}_{0,1}+\theta_2+\triangle_{\tau,1}-\triangle_{\tau,2}}.
\]
To obtain the maximum value of $f(\cdot)$, we consider its derivative as
\begin{align}
    f'(\theta_2)=\frac{\underline{\theta}_{0,1}}{\left( \underline{\theta}_{0,1}+\theta_2 \right)^2}-\frac{\underline{\theta}_{0,1}+\triangle_{\tau,1}}{\left(\underline{\theta}_{0,1}+\triangle_{\tau,1}+\theta_2-\triangle_{\tau,2}\right)^2}. \notag
\end{align}
One may directly verify that $f'(\theta_2)$ has the same sign with respect to
\[
\frac{\triangle_{\tau,1}-\triangle_{\tau,2}}{\underline{\theta}_{0,1}+\theta_2}-\left( \sqrt{1+\frac{\triangle_{\tau,1}}{\underline{\theta}_{0,1}}}-1\right).
\]
Thus, if we have
\[
 \frac{\triangle_{\tau,1}-\triangle_{\tau,2}}{\sqrt{1+\frac{\triangle_{\tau,1}}{\underline{\theta}_{0,1}}}-1}-\underline{\theta}_{0,1} \in [\underline{\theta}_{0,2}+\triangle_{\tau,2},\overline{\theta}_{0,2}],
\]
then the maximum value of $f$ is obtained when 
\[
\theta_2 = \frac{\triangle_{\tau,1}-\triangle_{\tau,2}}{\sqrt{1+\frac{\triangle_{\tau,1}}{\underline{\theta}_{0,1}}}-1}-\underline{\theta}_{0,1}.
\]
The corresponding value of $f$ equals
\[
1 +  \frac{\underline{\theta}_{0,1}+\triangle_{\tau,1}}{\triangle_{\tau,1}-\triangle_{\tau,2}} \cdot \left( 1-\sqrt{\frac{\underline{\theta}_{0,1}}{\underline{\theta}_{0,1}+\triangle_{\tau,1}}} \right)-\frac{\underline{\theta}_{0,1}}{\triangle_{\tau,1}-\triangle_{\tau,2}}\cdot  \left( \sqrt{1+\frac{\triangle_{\tau,1}}{\underline{\theta}_{0,1}}}-1 \right) .
\]
For the rest cases, the maximum of $f(\cdot)$ is obtained when $\theta_2$ equals $\underline{\theta}_{0,2}+\triangle_{\tau,2}$ or $\overline{\theta}_{0,2}$, whose corresponding values will be presented in the end of the proof.
\item $\theta_2=\underline{\theta}_{0,2}+\triangle_{\tau,2}$: we consider the function $f(\theta_1)$ as 
\[
f(\theta_1)=\frac{\underline{\theta}_{0,2}+\triangle_{\tau,2}}{\theta_1+\underline{\theta}_{0,2}+\triangle_{\tau,2}}+\frac{\theta_1+\triangle_{\tau,1}}{\theta_1+\triangle_{\tau,1}+\underline{\theta}_{0,2}}
\]
We may obtain the maximum value of $f(\cdot)$ along the similar lines with the previous case (when $\theta_1=\underline{\theta}_{0,1}$). One may verify that when
\[
\frac{\triangle_{\tau,2}-\triangle_{\tau,1}}{1-\sqrt{\frac{\underline{\theta}_{0,2}}{\underline{\theta}_{0,2}+\triangle_{\tau,2}}}} - \underline{\theta}_{0,2}-\triangle_{\tau,2} \in [\underline{\theta}_{0,1},\overline{\theta}_{0,1}-\triangle_{\tau,1}],
\]
then the maximum value of $f$ equals to
\[
1+\frac{\underline{\theta}_{0,2}+\triangle_{\tau,2}}{\triangle_{\tau,2}-\triangle_{\tau,1}} \cdot \left( 1-\sqrt{\frac{\underline{\theta}_{0,2}}{\underline{\theta}_{0,2}+\triangle_{\tau,2}}} \right) - \frac{\underline{\theta}_{0,2}}{\triangle_{\tau,2}-\triangle_{\tau,1}}\cdot \left( \sqrt{1+\frac{\triangle_{\tau,2}}{\underline{\theta}_{0,2}}}-1 \right)
\]
For the rest cases, the maximum of $f(\cdot)$ is obtained when $\theta_1$ equals $\underline{\theta}_{0,1}$ or $\overline{\theta}_{0,1}-\triangle_{\tau,1}$, whose corresponding values will be presented in the end of the proof.
\end{enumerate}
Finally, one may also derive the right-hand side value of Eq.~\eqref{eq:RRAwP-two-point-condition-leontief} in the following cases:
\begin{enumerate}
    \item $\theta_1=\underline{\theta}_{0,1}$ and $\theta_2=\underline{\theta}_{0,2}+\triangle_{\tau,2}$: the right-hand side value of Eq.~\eqref{eq:RRAwP-two-point-condition-leontief} equals
    \[
    \frac{V}{V+\sum_{t=\tau+1}^{T}V_t}\cdot \left( \frac{\underline{\theta}_{0,2}+\triangle_{\tau,2}}{\underline{\theta}_{0,1}+\underline{\theta}_{0,2}+\triangle_{\tau,2}}+\frac{\underline{\theta}_{0,1}+\triangle_{\tau,1}}{\underline{\theta}_{0,1}+\underline{\theta}_{0,2}+\triangle_{\tau,1}} \right).
    \]
    \item $\theta_1=\underline{\theta}_{0,1}$ and $\theta_2=\overline{\theta}_{0,2}$: the right-hand side value of Eq.~\eqref{eq:RRAwP-two-point-condition-leontief} equals
    \[
    \frac{V}{V+\sum_{t=\tau+1}^{T}V_t}\cdot \left( \frac{\overline{\theta}_{0,2}}{\underline{\theta}_{0,1}+\overline{\theta}_{0,2}} + \frac{\underline{\theta}_{0,1}+\triangle_{\tau,1}}{\underline{\theta}_{0,1}+\overline{\theta}_{0,2}+\triangle_{\tau,1}-\triangle_{\tau,2}} \right).
    \]
    \item $\theta_1=\overline{\theta}_{0,1}-\triangle_{\tau,1}$ and $\theta_2=\underline{\theta}_{0,2}+\triangle_{\tau,2}$: the right-hand side value of Eq.~\eqref{eq:RRAwP-two-point-condition-leontief} equals
    \[
    \frac{V}{V+\sum_{t=\tau+1}^{T}V_t}\cdot \left( \frac{\underline{\theta}_{0,2}+\triangle_{\tau,2}}{\overline{\theta}_{0,1}+\underline{\theta}_{0,2}+\triangle_{\tau,2}-\triangle_{\tau,1}} + \frac{\overline{\theta}_{0,1}}{\overline{\theta}_{0,1} + \underline{\theta}_{0,2}} \right).
    \]
\end{enumerate}

\section{Proof of Theorem~\ref{thm:phase-reduction}}
\label{sec:proof-of-phase-reduction}

Invoking the future-imposed conditions with instance $\mathcal{I}_K$, by Theorem~\ref{thm:main-equivalent-conditions}, we derive the following lemma. The details of the proof are deferred to Section~\ref{sec:proof-of-lemma-phase-reduction-in-the-last-phase}.
\begin{lemma}\label{lem:phase-reduction-in-the-last-phase}
    The minimax-MDP $\mathcal{I}_K$ admits a feasible policy if and only if Eq.~\eqref{eq:multi-phase-reduction-final-phase-two-point-0} and Eq.~\eqref{eq:multi-phase-reduction-final-phase-two-point-1} hold. These conditions are equivalent to the following $3\cdot (\tau_K - \tau_{K-1}) \cdot n_K^2 \cdot S^4$ environment-dependent linear constraints on $\bm{x}_{\tau_{K-1}}$:
    \begin{enumerate}
        \item For each $s_{\tau_{K}}\in F_{\tau_{K-1}\to \tau_{K}}(s_{\tau_{K-1}})$ and $i\in[n_{K}]$ such that $[W_K(s_{\tau
        _{K}})]_{i,K} > 0$, we have
        \begin{align}\label{eq:last-phase-minimax-MDP-w-U-w-w-x}
            [W_K(s_{\tau_{K}})]_{i,K}\cdot U_{\tau_{K-1}\to \tau_{K}}(s_{\tau_{K-1}},s_{\tau_{K}})+[W_K(s_{\tau_{K}})]_{i,0}+\sum_{j=1}^{K-1} [W_K(s_{\tau_{K}})]_{i,j}\cdot x_{\tau_{K-1},j}  \leq 0
        \end{align}
        \item For each $s_{\tau_{K}}\in F_{\tau_{K-1}\to \tau_{K}}(s_{\tau_{K-1}})$ and $i\in[n_{K}]$ such that $[W_K(s_{\tau_{K}})]_{i,K} < 0$, we have
        \begin{align}\label{eq:last-phase-minimax-MDP-w-V-w-w-x}
            [W_K(s_{\tau_{K}})]_{i,K}\cdot V_{\tau_{K-1}\to \tau_{K}}(s_{\tau_{K-1}},s_{\tau_{K}})+[W_K(s_{\tau_{K}})]_{i,0}+\sum_{j=1}^{K-1} [W_K(s_{\tau_{K}})]_{i,j}\cdot x_{\tau_{K-1},j}  \leq 0
        \end{align}
        \item For each $\alpha\in\{\tau_{K-1}+1,\dots,\tau_K\}$, $i,i'\in [n_{K}]$, $s_\alpha,s'_\alpha\in S_{\alpha}$, and $s_{\tau_{K}},s'_{\tau_{K}}\in 
S_{\tau_{K}}$ such that $s_{\tau_{K}}\in F_{\alpha\to \tau_{K}}(s_\alpha)$, $s'_{\tau_{K}}\in F_{\alpha\to \tau_{K}}(s'_\alpha)$, $[W_K(s_{\tau_{K}})]_{i,K}<0$, $[W_K(s_{\tau_{K}})]_{i',K}\geq 0$, and $\exists s_{\alpha-1}\in F_{\tau_{K-1}\to \alpha-1}(s_{\tau_{K-1}})$ with $s_\alpha, s'_\alpha\in F_{\alpha-1}(s_{\alpha-1})$,
\begin{align}\label{eq:last-phase-minimax-MDP-w-w-V-U-w-w-w-w-w-w-w-w-x}
    & [W_K(s'_{\tau_{K}})]_{i',K}\cdot [W_K(s_{\tau_{K}})]_{i,K} \cdot \Big[V_{\alpha\to \tau_{K}}(s_\alpha, s_{\tau_{K}})-U_{\alpha\to \tau_{K}}(s'_\alpha, s'_{\tau_{K}})\Big] \notag \\
    &+[W_K(s_{\tau_{K}})]_{i,0}\cdot [W_K(s'_{\tau_{K}})]_{i',K}-[W_K(s_{\tau_{K}})]_{i,K}\cdot [W_K(s'_{\tau_{K}})]_{i',0} \notag \\
    &+ \sum_{j=1}^{K-1} \bigg\{[W_K(s_{\tau_{K}})]_{i,j}\cdot [W_K(s'_{\tau_{K}})]_{i',K}-[W_K(s_{\tau_{K}})]_{i,K}\cdot [W_K(s'_{\tau_{K}})]_{i',j}\bigg\}\cdot x_{\tau_{K-1},j} \leq 0,
\end{align}
\end{enumerate}
\end{lemma}
We denote the corresponding feasible policy as $\pi^{\mathcal{I}_K}=(\pi^{\mathcal{I}_K}_{\tau_{K-1}},\dots,\pi^{\mathcal{I}_K}_{\tau_K-1})$, where for each $t\in\{\tau_{K-1},\dots,\tau_K-1\}$, $\pi_t^{\mathcal{I}_K}$ maps the present environment state and the inventory level to an action:
\[
\pi_t^{\mathcal{I}_K}(\cdot,x_{t,1},\dots,x_{t,K-1},\cdot): S_t \times \mathbb{R} \to \mathbb{R}, \qquad \forall t\in\{\tau_{K-1},\dots,\tau_K-1\},
\]
where $x_{t,i}$ for each $i\in[K-1]$ is actually a constant during $\mathcal{I}_K$, equaling $x_{\tau_{K-1},i}$.

Now we first prove that if $\widehat{\mathcal{W}}$ admits a feasible policy $\pi^{\widehat{\mathcal{W}}}$, then $\mathcal{W}$ admits a feasible policy $\pi^{\mathcal{W}}$. If $\widehat{\mathcal{W}}$ admits a feasible policy $\pi^{\widehat{\mathcal{W}}}$, then under policy $\pi^{\widehat{\mathcal{W}}}$, for any possible system state $(s_{\tau_{K-1}},\bm{x}_{\tau_{K-1}})$ at time $\tau_{K-1}$ under policy $\pi^{\widehat{\mathcal{W}}}$, we have that Eq.~\eqref{eq:multi-phase-reduction-final-phase-two-point-0} and Eq.~\eqref{eq:multi-phase-reduction-final-phase-two-point-1} hold for $(s_{\tau_{K-1}},\bm{x}_{\tau_{K-1}})$. Applying Lemma~\ref{lem:phase-reduction-in-the-last-phase}, we derive that $\mathcal{I}_K$ admits a policy $\pi^{\mathcal{I}_K}$. We then construct $\pi^{\mathcal{W}}$ as follows:
\begin{enumerate}
    \item For each $t\in[\tau_{K-1}-1]$ and $s_t\in S_t$,
    $
    \pi^{\mathcal{W}}_t(s_t,\bm{x_t})\defeq \pi_t^{\widehat{\mathcal{W}}}(s_t,\bm{x_t}).
    $
    \item For each $t\in\{\tau_{K-1},\dots,\tau_K-1\}$ and $s_t\in S_t$,
    $
    \pi^{\mathcal{W}}_t(s_t,\bm{x_t})\defeq \pi_t^{\mathcal{I}_K}(s_t,\bm{x_t}).
    $
\end{enumerate}
We may directly verify the environment-dependent inventory constraints before phase $K$ and the environment-dependent action constraints. It suffices to verify the environment-dependent inventory constraints on day $\tau_K$. We denote the $\bm{x}_{\tau_K}(\mathcal{I}_K)$ as the final inventory level at time $\tau_K$ of instance $\mathcal{I}_K$ under policy $\pi_t^{\mathcal{I}_K}$, and $\bm{x}_{\tau_K}(\mathcal{W})$ as the inventory level at time $\tau_K$ of instance $\mathcal{W}$ under policy $\pi^\mathcal{W}$. It can be directly verified that these two values are the same, denoted by $\bm{x}_{\tau_K}$. It suffices to show that
\begin{align}\label{eq:multi-phase-last-phase-L-x-R}
    \tilde{L}_{\tau_K}(s_{\tau_K})\leq x_{\tau_K,K}\leq \tilde{R}_{\tau_K}(s_{\tau_K}),
\end{align}
if and only if the environment-dependent inventory constraints at time $\tau_K$ in instance $\mathcal{W}$ are satisfied, i.e., 
\begin{align}\label{eq:multi-phase-last-phase-linear-constraints}
    [W_K(s_{\tau_K})]_{i,0}+\sum_{j=1}^K [W_K(s_{\tau_K})]_{i,j}\cdot x_{\tau_K,j}\leq 0,\qquad\qquad\forall i\in[n_K],
\end{align}
which may be directly verified by the definition (Eq.~\eqref{eq:multi-phase-inventory-constraint-left-final-phase} and Eq.~\eqref{eq:multi-phase-inventory-constraint-right-final-phase}).

Finally we prove that if $\mathcal{W}$ admits a feasible policy $\pi^\mathcal{W}$, then $\widehat{\mathcal{W}}$ admits a feasible policy $\pi^{\widehat{\mathcal{W}}}$. We construct $\pi^{\widehat{\mathcal{W}}}$ as $\pi_t^{\widehat{\mathcal{W}}}(s_t,\bm{x}_t)\defeq \pi^{\mathcal{W}}_t(s_t,\bm{x}_t)$ for any $t\in[\tau_{K-1}-1]$. Along the similar proof previously, it suffices to show that the additional environment-dependent inventory constraints (Eq.~\eqref{eq:multi-phase-reduction-final-phase-two-point-0} and Eq.~\eqref{eq:multi-phase-reduction-final-phase-two-point-1}) at time $\tau_{K-1}$ hold. Consider the system state on day $\tau_{K-1}$, $(s_{\tau_{K-1}},\bm{x}_{\tau_{K-1}})$. We have stated the equivalence between Eq.~\eqref{eq:multi-phase-last-phase-L-x-R} and Eq.~\eqref{eq:multi-phase-last-phase-linear-constraints}, and we can naturally correspond the phase $K$ under instance $\mathcal{W}$ starting from the system state $(s_{\tau_{K-1}},\bm{x}_{\tau_{K-1}})$, to the minimax-MDP instance $\mathcal{I}_K$. The existence of feasible policy $\pi^\mathcal{W}$ can derive the existence of feasible policy for the minimax-MDP $\mathcal{I}_K$, as $(s_{\tau_{K-1}},\bm{x}_{\tau_{K-1}})$ is a possible system state at time $\tau_{K-1}$ under $\pi^\mathcal{W}$. Invoking Lemma~\ref{lem:phase-reduction-in-the-last-phase}, we can derive that $(s_{\tau_{K-1}},\bm{x}_{\tau_{K-1}})$ satisfies Eq.~\eqref{eq:multi-phase-reduction-final-phase-two-point-0} and Eq.~\eqref{eq:multi-phase-reduction-final-phase-two-point-1}.
\hfill\Halmos

\subsection{Proof of Lemma~\ref{lem:phase-reduction-in-the-last-phase}}
\label{sec:proof-of-lemma-phase-reduction-in-the-last-phase}
Using the construction in Section~\ref{sec:solving-multi-phase-minimax-MDP-via-phase-reduction}, we compute the following objects:
\begin{enumerate}
    \item Future-imposed inventory level constraints: for any $\alpha,\beta\in\{\tau_{K-1},\dots,\tau_K\}$ with $\beta\geq\alpha$, and an environment state $s_\alpha\in S_\alpha$, we have
    \begin{equation*}
   \tilde{R}_{\beta \leadsto \alpha}(s_\alpha) = \left\{
    \begin{array}{lllll}
        0, \qquad \text{when}~\beta=\tau_{K-1}; \qquad\qquad \qquad   +\infty, \qquad \text{when}~0<\beta<T+1;\\[8pt]
        \min\limits_{s_{\tau_K}\in F_{\alpha\to \tau_K}(s_\alpha)} [\tilde{R}_{\tau_K}(s_{\tau_K})-U_{\alpha\to\tau_K}(s_\alpha,s_{\tau_K})], \qquad \text{when}~\beta=\tau_K. 
    \end{array}
    \right.
   \end{equation*}
    \begin{equation*}
   \tilde{L}_{\beta \leadsto \alpha}(s_\alpha) = \left\{
    \begin{array}{lllll}
        0, \qquad \text{when}~\beta=\tau_{K-1}; \qquad\qquad \qquad   -\infty, \qquad \text{when}~0<\beta<T+1;\\[8pt]
        \max\limits_{s_{\tau_K}\in F_{\alpha\to \tau_K}(s_\alpha)} [\tilde{L}_{\tau_K}(s_{\tau_K})-V_{\alpha\to\tau_K}(s_\alpha,s_{\tau_K})], \qquad \text{when}~\beta=\tau_K. 
    \end{array}
    \right.
   \end{equation*}
   We can also derive that
   \begin{align*}
    &\tilde{R}_{\star \leadsto \tau_{K-1}}(s_{\tau_{K-1}})=\min\left(0,\min\limits_{s_{\tau_K}\in F_{\tau_{K-1}\to \tau_K}(s_{\tau_{K-1}})} \left[\tilde{R}_{\tau_K}(s_{\tau_K})-U_{\tau_{K-1}\to\tau_K}(s_{\tau_{K-1}},s_{\tau_K})\right]\right), \\ 
    &\tilde{R}_{\star \leadsto \alpha}(s_\alpha)=\min\limits_{s_{\tau_K}\in F_{\alpha\to \tau_K}(s_\alpha)} \left[\tilde{R}_{\tau_K}(s_{\tau_K})-U_{\alpha\to\tau_K}(s_\alpha,s_{\tau_K})\right], \qquad \alpha\in\{\tau_{K-1}+1,\dots,\tau_K\}, \\
    &\tilde{L}_{\star \leadsto \tau_{K-1}}(s_{\tau_{K-1}})=\max\left(0,\max\limits_{s_{\tau_K}\in F_{\tau_{K-1}\to \tau_K}(s_{\tau_{K-1}})} \left[\tilde{L}_{\tau_K}(s_{\tau_K})-V_{\tau_{K-1}\to\tau_K}(s_{\tau_{K-1}},s_{\tau_K})\right]\right), \\ 
    &\tilde{L}_{\star \leadsto \alpha}(s_\alpha)=\max\limits_{s_{\tau_K}\in F_{\alpha\to \tau_K}(s_\alpha)} \left[\tilde{L}_{\tau_K}(s_{\tau_K})-V_{\alpha\to\tau_K}(s_\alpha,s_{\tau_K})\right], \qquad \alpha\in\{\tau_{K-1}+1,\dots,\tau_K\}.
    \end{align*}
   \item Strengthened future-imposed inventory bound: directly from the definition and the discussion for future-imposed inventory upper bounds, we can derive that for any $\alpha\in\{\tau_{K-1}+1,\dots,\tau_K\}$ and $s_{\alpha-1}\in S_{\alpha-1}$, we have 
   \begin{align}
           \tilde{R}_{\star\leadsto\alpha-1\to[\alpha]}(s_{\alpha-1})=\min\limits_{ s_\alpha\in F_{\alpha-1}(s_{\alpha-1}) \atop s_{\tau_K}\in F_{\alpha\to \tau_K}(s_\alpha)} \left[\tilde{R}_{\tau_K}(s_{\tau_K})-U_{\alpha\to\tau_K}(s_\alpha,s_{\tau_K})\right], \notag \\
           \tilde{L}_{\star\leadsto\alpha-1\to[\alpha]}(s_{\alpha-1})=\max\limits_{ s_\alpha\in F_{\alpha-1}(s_{\alpha-1}) \atop s_{\tau_K}\in F_{\alpha\to \tau_K}(s_\alpha)} \left[\tilde{L}_{\tau_K}(s_{\tau_K})-V_{\alpha\to\tau_K}(s_\alpha,s_{\tau_K})\right]. \notag
   \end{align}

\end{enumerate}

By Theorem~\ref{thm:main-equivalent-conditions}, the minimax-MDP $\mathcal{I}_K$ admits a feasible policy if and only if the following future-imposed conditions are satisfied: for any two time periods $\alpha \in \{\tau_{K-1}+1,\dots,\tau_K\}$, the \emph{future-imposed condition} requires that for each $s_{\alpha-1} \in F_{\tau_{K-1} \to \alpha-1}(s_{\tau_{K-1}})$, it holds that
\begin{align} \label{eq:two-point-condition-last-phase-multi-phase}
\tilde{R}_{\star \leadsto \alpha -1 \to [\alpha]}(s_{\alpha-1})   \geq \tilde{L}_{\star \leadsto \alpha -1 \to [\alpha]}(s_{\alpha-1}). 
\end{align} 
also, for $\alpha = \tau_{K-1}$, the future-imposed condition about time periods $\alpha$ requires that
\begin{align}\label{eq:two-point-condition-special-case-last-phase-multi-phase}
\tilde{R}_{\star \leadsto \tau_{K-1}}(s_{\tau_{K-1}}) \geq \tilde{L}_{\star\leadsto\tau_{K-1}}(s_{\tau_{K-1}}).
\end{align}
We discuss the above conditions according to the following cases.
\begin{enumerate}
    \item $\alpha=\tau_{K-1}$: we can derive that Eq.~\eqref{eq:two-point-condition-special-case-last-phase-multi-phase} is equivalent to the following two inequalities: 
    \begin{align}
        \min\limits_{s_{\tau_K}\in F_{\tau_{K-1}\to \tau_K}(s_{\tau_{K-1}})} \left[\tilde{R}_{\tau_K}(s_{\tau_K})-U_{\tau_{K-1}\to\tau_K}(s_{\tau_{K-1}},s_{\tau_K})\right]\geq 0, \label{eq:multi-phase-last-phase-condition-R-U-geq-0} \\
        \max\limits_{s_{\tau_K}\in F_{\tau_{K-1}\to \tau_K}(s_{\tau_{K-1}})} \left[\tilde{L}_{\tau_K}(s_{\tau_K})-V_{\tau_{K-1}\to\tau_K}(s_{\tau_{K-1}},s_{\tau_K})\right]\leq 0. \label{eq:multi-phase-last-phase-condition-L-V-geq-0}
    \end{align}
    Invoking the definition (Eq.~\eqref{eq:multi-phase-inventory-constraint-right-final-phase}), we can derive that Eq.~\eqref{eq:multi-phase-last-phase-condition-R-U-geq-0} correspond to: for each $s_{\tau_{K}}\in F_{\tau_{K-1}\to \tau_{K}}(s_{\tau_{K-1}})$ and $i\in[n_{K}]$ such that $[W_{K}(s_{\tau_{K}})]_{i,K} > 0$, we have
    \begin{align}\label{eq:multi-phase-linear-constraints-explicit-1}
        -\frac{1}{[W_{K}(s_{\tau_{K}})]_{i,K}}\left[[W_{K}(s_{\tau_{K}})]_{i,0}+\sum_{j=1}^{K-1} [W_{K}(s_{\tau_{K}})]_{i,j}\cdot x_{\tau_{K-1},j}\right]-U_{\tau_{K-1}\to \tau_{K}}(s_{\tau_{K-1}},s_{\tau_{K}})  \geq 0.
    \end{align}
    We may directly verify that it corresponds to Eq.~\eqref{eq:last-phase-minimax-MDP-w-U-w-w-x} and the number of such constraints is at most $S\cdot n_K$. Substituting Eq.~\eqref{eq:multi-phase-inventory-constraint-left-final-phase} into Eq.~\eqref{eq:multi-phase-last-phase-condition-L-V-geq-0}, we can derive that Eq.~\eqref{eq:multi-phase-last-phase-condition-L-V-geq-0} correspond to: for each $s_{\tau_{K}}\in F_{\tau_{K-1}\to \tau_{K}}(s_{\tau_{K-1}})$ and $i\in[n_{K}]$ such that $[W_{K}(s_{\tau_{K}})]_{i,K}< 0$,
    \begin{align}\label{eq:multi-phase-linear-constraints-explicit-2}
        V_{\tau_{K-1}\to \tau_{K}}(s_{\tau_{K-1}},s_{\tau_{K}})\geq -\frac{1}{[W_{K}(s_{\tau_{K}})]_{i,K}}\left[[W_{K}(s_{\tau_{K}})]_{i,0}+\sum_{j=1}^{K-1} [W_{K}(s_{\tau_{K}})]_{i,j}\cdot x_{\tau_{K-1},j} \right] .
    \end{align}
     We may directly verify that it corresponds to Eq.~\eqref{eq:last-phase-minimax-MDP-w-V-w-w-x} and the number of such constraints is at most $S\cdot n_K$.
    \item $\alpha>\tau_{K-1}$: invoking the computed (strengthened) future-imposed inventory bounds and the definitions (Eq.~\eqref{eq:multi-phase-inventory-constraint-left-final-phase}, Eq.~\eqref{eq:multi-phase-inventory-constraint-right-final-phase}), we can derive that Eq.~\eqref{eq:two-point-condition-last-phase-multi-phase} correspond to: for each $i,i'\in [n_{K}]$, $s_\alpha,s'_\alpha\in S_{\alpha}$, and $s_{\tau_{K}},s'_{\tau_{K}}\in 
    S_{\tau_{K}}$ such that $s_{\tau_{K}}\in F_{\alpha\to \tau_{K}}(s_\alpha)$, $s'_{\tau_{K}}\in F_{\alpha\to \tau_{K}}(s'_\alpha)$, $[W_{K}(s_{\tau_{K}})]_{i,K}<0$, $[W_{K}(s'_{\tau_{K}})]_{i',K} > 0$, and $\exists s_{\alpha-1}\in F_{\tau_{K-1}\to \alpha-1}(s_{\tau_{K-1}})$ with $s_\alpha, s'_\alpha\in F_{\alpha-1}(s_{\alpha-1})$, we have
    \begin{align}\label{eq:multi-phase-linear-constraints-explicit-3}
        &\Bigg[-\frac{1}{[W_{K}(s'_{\tau_{K}})]_{i',K}}\left([W_{K}(s'_{\tau_{K}})]_{i',0}+\sum_{j=1}^{K-1}[W_{K}(s'_{\tau_{K}})]_{i',j}\cdot x_{\tau_{K-1},j}\right)-U_{\alpha\to\tau_K}(s'_\alpha,s'_{\tau_K})\Bigg]  \notag\\
        &\geq -\frac{1}{[W_{K}(s_{\tau_{K}})]_{i,K}}\left([W_{K}(s_{\tau_{K}})]_{i,0}+\sum_{j=1}^{K-1} [W_{K}(s_{\tau_{K}})]_{i,j} \cdot x_{\tau_{K-1},j} \right)-V_{\alpha\to\tau_K}(s_\alpha,s_{\tau_K}). 
    \end{align}
     We may directly verify that it corresponds to Eq.~\eqref{eq:last-phase-minimax-MDP-w-w-V-U-w-w-w-w-w-w-w-w-x} and the number of such constraints is at most $(\tau_K-\tau_{K-1})\cdot n^2_K \cdot S^4 $.
  \end{enumerate}

\section{Omitted Proofs and Discussions in Section~\ref{sec:Multi-period-Ordering-Decisions-with-Multi-phase-costs-and-Predictions}}
We define the inventory level vector $\bm{x}_t=(x_{t,1},\dots,x_{t,K})\in \mathbb{R}^K$ by
\[
x_{t,v}=\sum_{j=\tau_{v-1}}^{\min(\tau_v-1,t)}a_j,\qquad \forall v\in[K],
\]
The ordering policy $\pi_t$ for each $t\in[T]$ maps an inventory level vector $\bm{x}_t$ with a prediction interval for demand to an ordering action $a_t\in[0,V_t]$. We then derive that for any $v\in[K]$ and $t\in\{\tau_{v-1},\dots, \tau_v-1\}$, 
\[
\bm{x}_{t+1}=\bm{x}_t+a_t\bm{e}_v,
\]
where $\bm{e}_v$ is the $v$-th canonical basis vector. We will use the notation $\bm{x}_t$ to denote the inventory level vector in this section. 

Second, consider $\sup_{\pi} \{ \varphi_\pi (\mathcal{V}) \}$, a crucial fact is that the worst-case ratio is achieved by the regular sequence with each interval length equaling the prediction error upper bound, referring to the intuition that the ``worst'' case is making the prediction interval rough. We formalize it as the observation below, with details of the proof deferred in Section~\ref{sec:proof-of-multi-cost-interval-length-fix}.  
\begin{observation}
\label{obs:multi-cost-interval-length-fix}
    Consider a slightly different robust competitive ratio of $\pi$ 
    \[
    \tilde{\varphi}_\pi(\mathcal{V})\defeq \inf_{\mathcal{P}, d} \{\varphi_\pi (\mathcal{P}, d; \mathcal{V})\},
    \]
    where the infimum is taken over all $(\mathcal{P}, d)$ pairs such that $\mathcal{P}$ is a regular prediction sequence with respect to $\{[\unld_0,\ovld_0], \{\triangle_t\}_{t\in[T]}\}$, and $d \in [\unld_T,\ovld_T]$ with $\ovld_t - \unld_t=\triangle_t$ for each $t\in[T]$. Then we have 
    \[
    \sup_{\pi} \{ \varphi_\pi (\mathcal{V}) \}=\sup_{\pi} \{ \tilde{\varphi}_\pi (\mathcal{V}) \}.
    \]
\end{observation}
We will focus on these special prediction sequences in the rest of this section.

\subsection{Optimizing Robust Competitive Ratio in $T^{O(K)}$}
\label{sec:optimizing-robust-competitive-ratio-in-T-O-K}
In this section, we work with the decision problem of whether there exists an online policy that achieves a robust competitive ratio of at least a given threshold $\Phi \leq 1$, i.e., deciding whether
\begin{align}\label{eq:decision-problem-multi-phase-cost-robust-comepetive-ratio}
\sup_{\pi} \{ \tilde{\varphi}_\pi (\mathcal{V}) \} \geq \Phi.
\end{align}

\noindent\uline{The correspondence.} To apply the multi-phase minimax-MDP, we reduce the decision problem Eq.~\eqref{eq:decision-problem-multi-phase-cost-robust-comepetive-ratio} to a multi-phase minimax-MDP $\mathcal{W}$ with time horizon $T+2$. We label the time periods by $0, 1, 2, \dots, T+1$, corresponding to the days in the timeline of $\mathcal{V}$. The time horizon and environment-dependent action constraints are defined the same way as in Section~\ref{sec:app-I-robust-regret}. The environment state set is defined by $S_t=\mathbb{R}_{\geq 0}$ for any $t\in\{0,\dots,T+1\}$.
The initial environment state is defined as $s_0=\unld_0$.
And the environment state transition functions are defined as 
\begin{align}
    F_t(a)=[a,a+\triangle_t-\triangle_{t+1}],\qquad \forall t\in\{0,\dots,T\},~a\in\mathbb{R}_{\geq 0},    
\end{align}
where $\triangle_0=\ovld_0-\unld_0$ and $\triangle_{T+1}=0$. The phase count of $\mathcal{W}$ equals $K$, which is the dimension of $\bm{\gamma}$, while the grid vector of $\mathcal{W}$ is $\{\tau_i\}_{i=0}^K$ in the instance $\mathcal{V}$. We only have two non-trivial environment-dependent inventory constraints at time $\tau_K$, referring to 
\begin{align}
    [\Phi\cdot \mathcal{R}^\sharp (\mathcal{P}, d; \mathcal{V})-p\cdot d]+\sum\limits_{j=1}^{K}\gamma_{j}\cdot x_{T+1,j}\leq 0, \qquad\qquad
    \Phi\cdot \mathcal{R}^\sharp (\mathcal{P}, d; \mathcal{V})-\sum\limits_{j=1}^{K}(p-\gamma_{j})\cdot x_{T+1,j}\leq 0, \notag
\end{align}
where we note that $\mathcal{R}^\sharp (\mathcal{P}, d; \mathcal{V})$ is a function independent of the prediction sequence $\mathcal{P}$ and is fully determined by $d$ and $\mathcal{V}$. We then have the following lemma about our reduction.
\begin{lemma}\label{lem:app-multi-phase-cost-reduction-competitive-ratio}
Given $\mathcal{V}$, let the multi-phase minimax-MDP $\mathcal{W}$ be constructed by the reduction in this subsection. Then Eq.~\eqref{eq:decision-problem-multi-phase-cost-robust-comepetive-ratio} holds if and only if $\mathcal{W}$ admits a feasible policy.
\end{lemma}
\proof{Proof.}
For any valid ordering policy $\pi^\mathcal{J}$ for $\mathcal{J}$, consider the mapping $\sigma : \pi^\mathcal{V} \mapsto \pi^\mathcal{W}$ defined as
\begin{align}
    \pi^\mathcal{W}_0(\cdot, \cdot) \equiv 0, \qquad \pi^\mathcal{W}_t(a, \bm{x}_t) = \pi^\mathcal{V}_t([a,a+\triangle_t], \bm{x}_t), \qquad \forall t\in[T], a\in\mathbb{R}_{\geq 0}
\end{align}
such that $\pi^\mathcal{W} = \sigma(\pi^\mathcal{V})$ is a policy for $\mathcal{W}$. Conversely, if $\pi^\mathcal{W}$ is a feasible policy for $\mathcal{W}$, one may verify the existence of $\sigma^{-1}(\pi^\mathcal{W})$. And we need to show that $\pi^\mathcal{W} = \sigma(\pi^\mathcal{V})$ is a feasible policy for $\mathcal{W}$ if and only if $\pi^\mathcal{V}$ is a valid ordering policy with $\tilde{\varphi}_\pi(\mathcal{V}) \geq \Phi$.  

We first prove that the environment state transition functions correspond to the regular prediction sequence (i.e., for each $t\in[T]$ , we have $b\in[a,a+\triangle_t-\triangle_{t+1}]$), if and only if $[b,b+\triangle_{t+1}]\subset[a,a+\triangle_t]$. One may directly verify this property. 

For convenience, let $\bm{x}_{T+1}(\mathcal{W})$ denote the inventory level vector at time period $T+1$ when running policy $\pi^\mathcal{W}$ in $\mathcal{W}$ (which can be naturally defined even if the policy does not meet the environment-dependent action constraints). Similarly, let $\bm{x}_{T+1}(\mathcal{V})$ be the inventory level vector at time $T+1$ under the ordering policy $\pi^\mathcal{V}$ in $\mathcal{V}$ (which can also be naturally defined even if the policy does not adhere to the supply capacity constraints). If $\pi^\mathcal{W} = \sigma(\pi^\mathcal{V})$, we directly have $\bm{x}_{T+1}(\mathcal{W}) = \bm{x}_{T+1}(\mathcal{V})$, denoted by $\bm{x}=(x_1,\dots,x_K)$.

Note that the only two non-trivial constraints in $\mathcal{W}$ are the environment-dependent inventory constraints at time period $T+1$. Our proof reduces to demonstrating that for any demand $d$, 
\begin{align*}
    [\Phi\cdot \mathcal{R}^\sharp (\mathcal{P}, d; \mathcal{V})-p\cdot d]+\sum\limits_{j=1}^{K}\gamma_{j}\cdot x_j\leq 0,\qquad\qquad 
    \Phi\cdot \mathcal{R}^\sharp (\mathcal{P}, d; \mathcal{V})-\sum\limits_{j=1}^{K}(p-\gamma_{j})\cdot x_j\leq 0,
\end{align*}
if and only if the competitive ratio of $\pi^{\mathcal{V}}$, given demand $d$ and inventory level vector $\bm{x}$, does not drop below $\Phi$, i.e.,
$
\left[p\cdot\min(d,\sum_{j=1}^{K}x_j)-\sum\limits_{j=1}^{K}\gamma_{j}\cdot x_j\right]/ \mathcal{R}^\sharp (\mathcal{P}, d; \mathcal{V}) \geq \Phi,
$
which can also be easily verified by considering the two cases: $\sum_{j=1}^{K}x_j \leq d$ and $\sum_{j=1}^{K}x_j > d$.
\hfill\Halmos
\endproof

By Lemma~\ref{lem:app-multi-phase-cost-reduction-competitive-ratio}, we apply the phase reduction and Theorem~\ref{thm:phase-reduction} to $\mathcal{W}$ to derive an efficient computation method for the decision problem Eq.~\eqref{eq:decision-problem-multi-phase-cost-robust-comepetive-ratio}, proof details of which will be provided in Section~\ref{sec:a-special-multi-phase-minimax-MDP}, Section~\ref{sec:phase-reduction-for-Q} and Section~\ref{sec:proof-of-multi-phase-cost-competitive-ratio}.

\begin{proposition}
\label{prop:multi-phase-cost-competitive-ratio}
Given $\Phi$ and a multi-period ordering decisions with multi-phase
costs and predictions instance $\mathcal{V}$ with $\{c_{i}\}_{i\in[K]}$ strictly increasing and $c_{K} < p$, after the above correspondence to a multi-phase minimax-MDP $\mathcal{W}$, we can use the phase reduction to compute the correctness of Eq.~\eqref{eq:decision-problem-multi-phase-cost-robust-comepetive-ratio} in $T^{O(K)}$ time. Moreover, we can give a corresponding online feasible efficient algorithm once we finish computing. 
\end{proposition}

\subsection{A PTAS for Robust Ordering Decisions under Changing Costs}
\label{sec:a-ptas-for-robust-ordering-decisions-under-changing-costs}
\begin{proposition}
    Given an instance of multi-period ordering decisions with multi-phase costs and predictions problem $\mathcal{V}$, there exists an algorithm with time complexity of $T^{O(\zeta\cdot\epsilon^{-1})}$ computing the robust competitive ratio 
    $
    \Phi^*\defeq \sup_{\pi}\{\varphi_\pi(\mathcal{V})\}
    $
    with an accuracy of $\epsilon$, i.e., the calculated ratio $\Phi$ satisfies
    $
    |\Phi-\Phi^*|\leq \epsilon$. Furthermore, we may set the constant 
    $
    \zeta = \max \left(\ln\left(\frac{p-\gamma_1}{p-\gamma_K}\right),\ln\left(\frac{\gamma_K}{\gamma_1}\right)\right)$.
\end{proposition}
\proof{Proof.}
Given an instance $\mathcal{V}$ with price $p$ and $\{\gamma_v\}_{v\in[K]}$, we first construct another instance $\widehat{\mathcal{V}}$ which is the same as $\mathcal{V}$ except for the sequence $\{\gamma_v\}_{v\in[K]}$ replaced by $\{\hat{\gamma}_v\}_{v\in[K]}$. We consider the following sequence:
\begin{align}\label{eq:grid-vector-for-cost}
      \left\{\gamma_1\cdot(1+\delta)^{i-1}\right\}_{i\in[N]}\cup\left\{p-(p-\gamma_K)\cdot (1+\delta)^{N+1-i}\right\}_{i\in[N]},
\end{align} 
where
\[
N=3+\lceil \frac{1}{\ln(1+\delta)}\cdot \max \left(\ln\left(\frac{p-\gamma_1}{p-\gamma_K}\right),\ln\left(\frac{\gamma_K}{\gamma_1}\right)\right) \rceil=O(\zeta \cdot \delta^{-1}).
\]
The definition of $N$ ensures that 
\[
\gamma_1\cdot(1+\delta)^{N-2}\geq \gamma_K, \qquad p-(p-\gamma_K)\cdot (1+\delta)^{N-1}\leq \gamma_1.
\]
We sort the sequence Eq.~\eqref{eq:grid-vector-for-cost} in ascending order and label it as $\{\gamma_v^*\}_{v\in[2N]}$. The new price sequence $\{\hat{\gamma}_v\}_{v\in[K]}$ is defined by $\hat{\gamma}_i\defeq \max_{j\in[2N]\wedge \gamma^*_j\leq \gamma_i} \gamma_j^*$ for any $ i\in[K]$.
Using the definition of $\{\gamma_i^*\}_{i\in[2N]}$, one may directly verify that 
\begin{align}\label{eq:grid-relationship-between-cost}
    \hat{\gamma}_i\leq \gamma_i \leq (1+\delta)\cdot \hat{\gamma}_i,\qquad (1+\delta)^{-1}\cdot (p-\hat{\gamma}_i)\leq p-\gamma_i\leq (p-\hat{\gamma}_i).
\end{align}

Second, we construct a near-optimal policy for $\mathcal{V}$ by the optimal policy of $\widehat{\mathcal{V}}$. Invoking Proposition~\ref{prop:multi-phase-cost-competitive-ratio}, with time complexity $T^{O(\zeta\cdot \delta^{-1})}$, we can compute a policy $\pi^{\hat{\mathcal{V}}}$ and a value $\Phi\geq 0$, with the policy having a robust competitive ratio at least $\Phi$, and satisfying
\begin{align}\label{eq:choose-pi-hat-V}
    \Phi\geq \sup_{\pi} \{ \varphi_\pi (\widehat{\mathcal{V}})\}-\delta. 
\end{align} 
Consider a mapping 
\[
\sigma: \pi^{\hat{\mathcal{V}}} \mapsto \pi^{\mathcal{V}}=(\pi^{\mathcal{V}}_1=\pi^{\hat{\mathcal{V}}}_1,\pi^{\mathcal{V}}_2=\pi^{\hat{\mathcal{V}}}_2,\dots,\pi^{\mathcal{V}}_{T}=\pi^{\hat{\mathcal{V}}}_T)
\]
such that $\pi^{\mathcal{V}}=\sigma(\pi^{\hat{\mathcal{V}}})$ is a policy for $\mathcal{V}$. Conversely, if $\pi^{\mathcal{V}}$ is a feasible policy for $\mathcal{V}$, one may verify the existence of $\sigma^{-1}(\pi^{\hat{\mathcal{V}}})$. We denote $\pi^{\mathcal{V}}$ as $\sigma(\pi^{\hat{\mathcal{V}}})$. 

Third, we analyze the difference of robust competitive ratio between $\pi^{\mathcal{V}}$ under $\mathcal{V}$ and $\pi^{\hat{\mathcal{V}}}$ under $\hat{\mathcal{V}}$. We propose the following lemma, the details of whose proof are deferred in Section~\ref{sec:proof-of-competitive-ratio-between-pi-V-pi-hat-V}.

\begin{lemma}\label{lem:competitive-ratio-between-pi-V-pi-hat-V}
    For any regular prediction sequence $\mathcal{P}$ and demand $d$, we have
    \[
        \varphi_{\pi^{\mathcal{V}}} (\mathcal{P}, d; \mathcal{V})\geq \varphi_{\pi^{\hat{\mathcal{V}}}} (\mathcal{P}, d; \widehat{\mathcal{V}})-3\delta.
    \]
    That is, the robust competitive ratio of $\pi^{\mathcal{V}}$ under $\mathcal{V}$ is at least $\Phi-3\delta$.
\end{lemma}  

Fourth, we prove that $\sup_{\pi} \{ \varphi_\pi (\widehat{\mathcal{V}})\}\geq\sup_{\pi} \{ \varphi_\pi (\mathcal{V})\}-\delta$. Consider the policy $\tilde{\pi}^{\mathcal{V}}$ that reaches the optimal robust competitive ratio, i.e.,
\begin{align}\label{eq:choose-tilde-pi}
    \varphi_{\tilde{\pi}^{\mathcal{V}}}(\mathcal{V})=  \sup_{\pi} \{ \varphi_\pi (\mathcal{V}) \},
\end{align}
and we denote $\tilde{\pi}^{\hat{\mathcal{V}}}$ as $\sigma^{-1}(\tilde{\pi}^{\mathcal{V}})$. Consider a regular prediction sequence $\mathcal{P}$ and demand $d$ satisfying
\begin{align}\label{eq:choose-P-d}
    \varphi_{\tilde{\pi}^{\hat{\mathcal{V}}}} (\mathcal{P}, d; \widehat{\mathcal{V}})\leq \sup_{\pi} \{ \varphi_\pi (\widehat{\mathcal{V}})\}.
\end{align}
One may verify that the final inventory level vector of $\mathcal{V}$ under policy $\tilde{\pi}^{\mathcal{V}}$ equals the final inventory level vector of $\hat{\mathcal{V}}$ under policy $\tilde{\pi}^{\hat{\mathcal{V}}}$, denoted by $\tilde{\bm{x}}$. One may also verify that the inventory level vector of $\mathcal{V}$ under hindsight optimal policy equals the inventory level vector of $\hat{\mathcal{V}}$ under hindsight optimal policy, denoted by $\bm{x}^*$. We focus on the relationship between $\varphi_{\tilde{\pi}^{\hat{\mathcal{V}}}} (\mathcal{P}, d; \widehat{\mathcal{V}})$ and $\varphi_{{\tilde{\pi}}^{\mathcal{V}}} (\mathcal{P}, d; \mathcal{V})$. We have 
\begin{align}\label{eq:phi-tilde-pi-V-geq-0}
    \varphi_{{\tilde{\pi}}^{\mathcal{V}}} (\mathcal{P}, d; \mathcal{V})&=\frac{p\cdot\min(d,\sum_{j=1}^K \tilde{x}_i)-\sum_{j=1}^K \gamma_i\cdot \tilde{x}_i}{\sum_{j=1}^K (p-\gamma_i)\cdot x_i^*}\geq \sup_{\pi} \{ \varphi_\pi (\mathcal{V}) \}\geq 0, 
\end{align}
while the inequality holds due to Eq.~\eqref{eq:choose-tilde-pi}. From Eq.~\eqref{eq:phi-tilde-pi-V-geq-0}, we have
\[
p\cdot\min(d,\sum_{j=1}^K \tilde{x}_i)-\sum_{j=1}^K \gamma_i\cdot \tilde{x}_i\geq 0.
\]
Invoking Eq.~\eqref{eq:grid-relationship-between-cost}, one may further verify that
\begin{align}
    \varphi_{\tilde{\pi}^{\hat{\mathcal{V}}}} (\mathcal{P}, d; \widehat{\mathcal{V}})&=\frac{p\cdot\min(d,\sum_{j=1}^K \tilde{x}_i)-\sum_{j=1}^K \hat{\gamma}_i\cdot \tilde{x}_i}{\sum_{j=1}^K (p-\hat{\gamma}_i)\cdot x_i^*} \geq \frac{p\cdot\min(d,\sum_{j=1}^K \tilde{x}_i)-\sum_{j=1}^K \gamma_i\cdot \tilde{x}_i}{\sum_{j=1}^K (p-\hat{\gamma}_i)\cdot x_i^*} \notag \\
    &\geq (1+\delta)^{-1}\cdot\frac{p\cdot\min(d,\sum_{j=1}^K \tilde{x}_i)-\sum_{j=1}^K \gamma_i\cdot \tilde{x}_i}{\sum_{j=1}^K (p-\gamma_i)\cdot x_i^*} =(1+\delta)^{-1}\cdot\varphi_{{\tilde{\pi}}^{\mathcal{V}}} (\mathcal{P}, d; \mathcal{V}),
\end{align}
while the first inequality is due to $\gamma_i\geq \hat{\gamma}_i$ and the second one is from $(p-\gamma_i)\geq (1+\delta)^{-1}(p-\hat{\gamma}_i)$. Combining the above discussions, we have 
\begin{align}\label{eq:compare-phi-tilde-pi-hat-V-phi-tilde-pi-V}
    \sup_{\pi} \{ \varphi_\pi (\widehat{\mathcal{V}})\}\geq\varphi_{\tilde{\pi}^{\hat{\mathcal{V}}}} (\mathcal{P}, d; \widehat{\mathcal{V}})\geq (1+\delta)^{-1}\cdot\varphi_{{\tilde{\pi}}^{\mathcal{V}}} (\mathcal{P}, d; \mathcal{V})\geq \varphi_{{\tilde{\pi}}^{\mathcal{V}}} (\mathcal{P}, d; \mathcal{V})-\delta\geq \sup_{\pi} \{ \varphi_\pi (\mathcal{V})\}-\delta.
\end{align}

Finally, we prove the competitive ratio of our policy $\pi^{\mathcal{V}}$ under instance $\mathcal{V}$ is at least $\sup_{\pi} \{ \varphi_\pi (\mathcal{V})\}-5\delta$. We have
  $\Phi\geq  \sup_{\pi} \{ \varphi_\pi (\widehat{\mathcal{V}})\}-\delta \geq \sup_{\pi} \{ \varphi_\pi (\mathcal{V})\}-2\delta$,
where the first inequality is due to Eq.~\eqref{eq:choose-pi-hat-V}, and the second one is due to Eq.~\eqref{eq:compare-phi-tilde-pi-hat-V-phi-tilde-pi-V}. Invoking Lemma~\ref{lem:competitive-ratio-between-pi-V-pi-hat-V}, we deduce that the robust competitive ratio of $\pi^{\mathcal{V}}$ under instance $\mathcal{V}$ is at least $\sup_{\pi} \{ \varphi_\pi (\mathcal{V})\}-5\delta$. Noticing that we compute the policy $\pi^{\mathcal{V}}$ and the value $\Phi$ within $T^{O(\zeta\cdot \delta^{-1})}$ time, we finish the proof.
\hfill\Halmos
\endproof

\subsection{Proof of Observation~\ref{obs:multi-cost-interval-length-fix}}
\label{sec:proof-of-multi-cost-interval-length-fix}
One may directly verify that $\tilde{\varphi}_\pi(\mathcal{V})\geq \varphi_\pi(\mathcal{V})$. Thus we have 
$ \sup_{\pi} \{ \varphi_\pi (\mathcal{V}) \}\leq\sup_{\pi} \{ \tilde{\varphi}_\pi (\mathcal{V}) \}$.
It suffices to show that $\sup_{\pi'} \{ \varphi_{\pi'} (\mathcal{V}) \} \geq \tilde{\varphi}_\pi (\mathcal{V}) $ for any policy $\pi$. We first construct a desired online policy $\hat{\pi}$ that utilizes all visited states and conclude that there also exists a desired policy that only utilizes the last visited state due to the Markovian property. We construct $\hat{\pi}_t$ for each $t\in[T]$ as follows:
\[
\hat{\pi}_t([\unld_1,\ovld_1],\bm{x}_1,\dots,[\unld_t,\ovld_t],\bm{x}_t)\defeq \pi_t([\unld'_1,\ovld'_1],\bm{x}_1,\dots,[\unld'_t,\ovld'_t],\bm{x}_t),
\]
where we recursively define $[\unld'_i,\ovld'_i]$ for each $i\in[t]$ as 
\begin{align}
    [\unld'_i,\ovld'_i]=
    \begin{cases}
        [\ovld_i-\triangle_i,\ovld_i], & \text{if~}(\ovld_i-\triangle_i \geq \unld'_{i-1}), \\
        [\unld'_{i-1},\unld'_{i-1}+\triangle_i],  & \text{if~}(\ovld_i-\triangle_i < \unld'_{i-1}),
    \end{cases}
\end{align}
where $\unld'_{0}\defeq \unld_0$. 
For any regular prediction sequence $\mathcal{P}=\{[\unld_i,\ovld_i]\}_{i\in[T]}$ and demand $d$, consider another prediction sequence $\mathcal{P}'=\{[\unld'_i,\ovld'_i]\}_{i\in[T]}$ constructed as above. It can be directly verified that $\mathcal{P}'$ satisfies $\ovld'_i - \unld'_i=\triangle_i$ and $ [\unld_i,\ovld_i]\subset [\unld'_i,\ovld'_i]$ for any $i\in[T]$, which indicates that $\mathcal{P}'$ is a regular prediction sequence under $(\{\triangle_t\}_{t\in[T]},d)$. We use $\bm{x}_{T+1}(\mathcal{P})$ to denote the inventory level under policy $\hat{\pi}$ and prediction sequence $\mathcal{P}$, and denote $\bm{x}_{T+1}(\mathcal{P}')$ as the inventory level under policy $\pi$ and prediction sequence $\mathcal{P}'$. We may verify that $\bm{x}_{T+1}(\mathcal{P}')=\bm{x}_{T+1}(\mathcal{P})$, and thus we have 
$
\mathcal{R}_{\hat{\pi}} (\mathcal{P}, d; \mathcal{V})=\mathcal{R}_\pi (\mathcal{P}', d; \mathcal{V})$,
which indicates that 
$
\varphi_{\hat{\pi}} (\mathcal{P}, d; \mathcal{V})=\varphi_\pi (\mathcal{P}', d; \mathcal{V})$.

The above analysis implies that 
$
 \varphi_{\hat{\pi}} (\mathcal{V})  \geq \tilde{\varphi}_\pi (\mathcal{V})$,
which is exactly what we want.
\hfill\Halmos

\subsection{A Special Multi-phase Minimax-MDP}
\label{sec:a-special-multi-phase-minimax-MDP}
We focus on a special multi-phase Minimax-MDP, denoted by a tuple \begin{align*}
\mathcal{Q}\defeq\{T,K,n_A,n_B,\{\tau_{i}\}_{i=0}^{K},A,B,s_{0},\{\triangle_{t}\}_{0\leq t\leq T+1},\{V_{t}\}_{t\in [T]},\{L_{j}(\cdot)\}_{j\in[n_A]},\{R_{j}(\cdot)\}_{j\in[n_B]}\},
\end{align*} 
where $A\in\mathbb{R}^{n_A\times K}$, $B\in\mathbb{R}^{n_B\times K}$. The total time  horizon is $T+2$ and we label the time periods by $0, 1, 2, \dots, T+1$. The phase count is $K$, while the grid vector is $\{\tau_{i}\}_{i=0}^{K}$. The environment-dependent action constraints are defined the same way as in Section~\ref{sec:app-I-robust-regret}. The environment state set is defined by $S_t=\mathbb{R}_{\geq 0}$ for any $t\in\{0,\dots,T+1\}$. The initial environment state is $s_0$. And the environment state transition functions are defined as 
\begin{align}
    F_t(a)=[a,a+\triangle_t-\triangle_{t+1}],\qquad \forall t\in\{0,\dots,T\},~a\in\mathbb{R}_{\geq 0}.    
\end{align}
The constraints total $n_i$ for each $i\in[K]$ is defined by $n_i=(n_A+n_B)\cdot\mathbf{1}\{i=K\}$ for any $i\in[K]$. The constraints total demonstrates that we only have $n_A+n_B$ non-trivial environment-dependent inventory constraints at time $\tau_K$, referring to 
\begin{align}
    \sum_{j=1}^{K}A_{i,j}\cdot x_{T+1,j}\geq L_{i}(s_{T+1}),\qquad\forall i\in[n_A], \qquad\qquad
    \sum_{j=1}^{K}B_{i,j}\cdot x_{T+1,j}\leq R_{i}(s_{T+1}),\qquad\forall i\in[n_B]. \notag
\end{align}

Now, to better characterize the multi-period ordering decisions with multi-phase costs and predictions problem, we introduce the following definition.
\begin{definition}[Monotonous $G$-linear Instance]
\label{def:monotonouslinearinstance}
An instance $\mathcal{Q}$ defined above is called a monotonous $G$-linear instance if and only if: 
\begin{enumerate}
    \item $n_B=1$ and $R_{1}(\cdot)$ is a piecewise linear and increasing function with at most $G$ linear pieces.
    \item For any $i\in[n_A]$, $L_{i}(\cdot)$ is a continuous piecewise linear function with at most $12$ linear pieces.
    \item For any $i\in[n_A]$, $\{A_{i,j}\}_{j=1}^{K}$ is a strictly decreasing sequence satisfying $A_{i,K}>0$.
    \item $\{B_{1,j}\}_{j=1}^{K}$ is a strictly increasing sequence satisfying $B_{1,1}>0$.
\end{enumerate}
\end{definition}

\subsection{Phase Reduction for $\mathcal{Q}$} 
\label{sec:phase-reduction-for-Q}
We first introduce the following lemma, which will be useful when constructing phase reduction, whose proof will be presented in Section~\ref{sec:proofofsplit}.
\begin{lemma}
\label{lem:decomposition-of-range-max-linear-function}
Given a continuous piecewise linear function $g(\cdot)$ with $M$ pieces and a continuous decreasing piecewise linear function $f(\cdot)$ with $N$ pieces, which can be recorded using $O(M)$ and $O(N)$ space complexity, consider $\triangle>0,~h>0$ and the function $w(\cdot)$ which is defined as follows:
\begin{align}
w(x)=\mathop{\max}\limits_{y,z\in[x,x+\triangle] \atop |y-z|\leq h}[g(y)+f(z)].
\end{align}
Then $w(\cdot)$ is continuous and there exist $M\cdot N$ continuous piecewise linear functions $\{w_{i}(\cdot)\}_{i\in[M\cdot N]}$, while each one has at most $12$ linear pieces, satisfying 
\begin{align}
w(x)=\mathop{\max}\limits_{i\in[M\cdot N]}w_{i}(x).
\end{align}
Moreover, $\{w_{i}(\cdot)\}_{i\in[M\cdot N]}$ can be computed completely in $O(M\cdot N)$ time.
\end{lemma}

Given a monotonous $G$-linear instance $\mathcal{Q}$, we use phase reduction to construct another instance:
\begin{align}
\widehat{\mathcal{Q}}\defeq\{\widehat{T},\widehat{K},\widehat{n}_A,n_B=1,\{\tau_{i}\}_{i=0}^{K},\widehat{A},\widehat{B},s_{0},\{\triangle_{t}\}_{0\leq t\leq \hat{T}+1},\{V_{t}\}_{t\in [\hat{T}]},\{\widehat{L}_{j}(\cdot)\}_{j\in[\hat{n}_A]},\{\widehat{R}_{j}(\cdot)\}_{j\in[n_B]}\},
\end{align} 
where $A\in\mathbb{R}^{\hat{n}_A\times \hat{K}}$, $B\in\mathbb{R}^{n_B\times \hat{K}}$, and we only change the value of variables with a hat. The time horizon, phase count and constraints total are defined as
\begin{align}
    \widehat{T}\defeq \tau_{K-1}-1,\qquad \widehat{K}\defeq K-1,\qquad \widehat{n}_A\defeq 12\cdot n_A+12\cdot G\cdot n_A\cdot n_B\cdot (\tau_{K}-\tau_{K-1}).
\end{align}
Then, we define the $\widehat{n}_A+n_B$ nontrivial inventory constraints in the form of linear constraints as follows (referring to Eq.~\eqref{eq:additional-constraint-for-hat-Q-1}, Eq.~\eqref{eq:additional-constraint-for-hat-Q-2}, Eq.~\eqref{eq:additional-constraint-for-hat-Q-3}), which actually correspond to the additional linear constraints of phase reduction in Section~\ref{sec:solving-multi-phase-minimax-MDP-via-phase-reduction} (referring to the Eq.~\eqref{eq:multi-phase-linear-constraints-explicit-1}, Eq.~\eqref{eq:multi-phase-linear-constraints-explicit-2} and Eq.~\eqref{eq:multi-phase-linear-constraints-explicit-3} in Section~\ref{sec:proof-of-lemma-phase-reduction-in-the-last-phase}):
\begin{enumerate}
    \item Eq.~\eqref{eq:multi-phase-linear-constraints-explicit-1}: substituting the parameter in $\mathcal{Q}$ into Eq.~\eqref{eq:multi-phase-linear-constraints-explicit-1}, we have 
    \[
    \sum_{j=1}^{K-1}B_{i,j}\cdot x_{\tau_{K-1},j}\leq R_i(s), \qquad \forall 0\leq s-s_{\tau_{K-1}}\leq \triangle_{\tau_{K-1}}-\triangle_{\tau_K}~\text{and}~i\in[n_B].
    \]
    According to the definition of monotonous $G$-linear instance, we have $R_{i}(\cdot)$ is an increasing function, thus the above constraint corresponds to
    \begin{align}\label{eq:additional-constraint-for-hat-Q-1}
        \sum_{j=1}^{K-1}B_{i,j}\cdot x_{\tau_{K-1},j}\leq R_i(s_{\tau_{K-1}}), \qquad i\in[n_B].
    \end{align}
    \item Eq.~\eqref{eq:multi-phase-linear-constraints-explicit-2}: substituting the parameter in $\mathcal{Q}$ into Eq.~\eqref{eq:multi-phase-linear-constraints-explicit-2}, we have 
    \begin{align}\label{eq:multi-phase-linear-constraints-explicit-2-withou-simplification}
        \sum_{t=\tau_{K-1}}^{\tau_K-1}V_t \geq - \frac{1}{(-A_{i,K})}&\left[L_{i}(s)+\sum_{j=1}^{K-1} (-A_{i,j})\cdot x_{\tau_{K-1},j} \right], \\
        &\forall 0\leq s-s_{\tau_{K-1}}\leq\triangle_{\tau_{K-1}}-\triangle_{\tau_K}~\text{and}~i\in[n_A].
    \end{align}
    Noticing that $L_{i}(\cdot)$ is a continuous piecewise linear function with at most 12 pieces, invoking Lemma~\ref{lem:decomposition-of-range-max-linear-function}, we define $L_i^r$ for each $r\in[12]$ as
    \[
    \max_{y\in[x,x+\triangle_{\tau_{K-1}}-\triangle_{\tau_K}]} L_i(y)=\max_{r\in[12]}L_i^r(x),
    \]
    where $L_i^r$ for each $r\in[12]$ is a continuous piecewise linear function with at most $12$ pieces. And the above linear constraint Eq.~\eqref{eq:multi-phase-linear-constraints-explicit-2-withou-simplification} corresponds to
    \begin{align}\label{eq:additional-constraint-for-hat-Q-2}
        \sum_{j=1}^{K-1} A_{i,j}\cdot x_{\tau_{K-1},j} \geq L_i^r(s_{\tau_{K-1}})-A_{i,K} \cdot  \sum_{t=\tau_{K-1}}^{\tau_K-1}V_t, \qquad \forall i\in[n_A]~\text{and}~r\in[12]. 
    \end{align}
    \item Eq.~\eqref{eq:multi-phase-linear-constraints-explicit-3}:
    substituting the parameter in $\mathcal{Q}$ into Eq.~\eqref{eq:multi-phase-linear-constraints-explicit-3}, we have that for any $\alpha\in[\tau_{K-1}+1,\tau_K]$, $s_\alpha$, $s'_\alpha$, $s_{\tau_K}$, $s'_{\tau_K}$ such that $s_{\tau_K}\in[s_\alpha,s_\alpha+\triangle_\alpha-\triangle_{\tau_K}]$, $s'_{\tau_K}\in[s'_\alpha,s'_\alpha+\triangle_\alpha-\triangle_{\tau_K}]$ and $\exists s_{\alpha-1}\in[s_{\tau_{K-1}},s_{\tau_{K-1}}+\triangle_{\tau_{K-1}}-\triangle_{\alpha-1}]~\text{satisfying}~ s_\alpha,s'_\alpha \in [s_{\alpha-1},s_{\alpha-1}+\triangle_{\alpha-1}-\triangle_\alpha]$, the following holds for each $i\in[n_A]$: 
    \begin{align}\label{eq:multi-phase-linear-constraints-explicit-3-withou-simplification}
        -\frac{1}{B_{1,K}}\left[-R_{1}(s'_{\tau_K})+\sum_{j=1}^{K-1} B_{1,j}\cdot x_{\tau_{K-1},j}\right] \geq -\sum_{t=\alpha}^{\tau_K-1}V_t+ \frac{1}{A_{i,K}}\left[L_i(s_{\tau_K})-\sum_{j=1}^{K-1} A_{i,j}\cdot x_{\tau_{K-1},j}\right].
    \end{align}
    We may directly verify that the possible range for $(s_{\tau_{K}},s'_{\tau_K})$ is 
    $    s_{\tau_K},s'_{\tau_K}\in [s_{\tau_{K-1}},s_{\tau_{K-1}}+\triangle_{\tau_{K-1}}\triangle_{\tau_K}]$, and $|s_{\tau_K}-s'_{\tau_K}|\leq \triangle_{\alpha-1}-\triangle_{\tau_K}$.
    Noticing $R_{1}(\cdot)$, $L_i(\cdot)$ are continuous piecewise linear functions with at most $G$ and $12$ linear pieces, invoking Lemma~\ref{lem:decomposition-of-range-max-linear-function}, we define $L_i^{r,\alpha}(\cdot)$ for each $r\in[12\cdot G]$ and $\alpha\in[\tau_{K-1}+1,\tau_K]$ as
    \[
    \max_{y,z\in [x,x+\triangle_{\tau_{K-1}}-\triangle_{\tau_K}] \atop |y-z|\leq \triangle_{\alpha-1}-\triangle_{\tau_K}} \left[\frac{L_{i}(y)}{A_{i,K}}-\frac{R_{1}(z)}{B_{1,K}}\right]=\max_{r\in[12\cdot G]}L_i^{r,\alpha}(x),\qquad \forall \alpha\in[\tau_{K-1}+1,\tau_K],
    \]
    where $L_i^{r,\alpha}(\cdot)$ for each $r\in[12\cdot G]$ and $\alpha\in[\tau_{K-1}+1,\tau_K]$ is a continuous piecewise linear function with at most $12$ pieces. And the above linear constraint Eq.~\eqref{eq:multi-phase-linear-constraints-explicit-3-withou-simplification} corresponds to 
    \begin{align}\label{eq:additional-constraint-for-hat-Q-3}
        \sum_{j=1}^{K-1} \left[\frac{A_{i,j}}{A_{i,K}}-\frac{B_{1,j}}{B_{i,K}}\right]\cdot x_{\tau_{K-1},j}\geq L_{i}^{r,\alpha}(s_{\tau_{K-1}})-\sum_{t=\alpha}^{\tau_K-1}V_t,~~ \forall i\in[n_A], \alpha\in[\tau_{K-1}+1,\tau_K],r\in[12\cdot G]
    \end{align}
\end{enumerate}

The above process demonstrates how we construct the phase reduction for $\mathcal{Q}$, denoted by $\widehat{\mathcal{Q}}$. Invoking Theorem~\ref{thm:phase-reduction}, one may directly verify that $\mathcal{Q}$ admits a feasible policy if and only if $\widehat{\mathcal{Q}}$ admits a feasible policy. Moreover, we can also derive that if $\mathcal{Q}$ is a monotonous $G$-linear instance, then $\widehat{\mathcal{Q}}$ is also a monotonous $G$-linear instance. We formalize it as the lemma below. 
\begin{lemma}
\label{lem:inherit-of-monotonous-linear-instance}
If $\mathcal{Q}$ is a monotonous $G$-linear instance, then the constructed $\widehat{\mathcal{Q}}$ above is also a monotonous $G$-linear instance.
\end{lemma}
\proof{Proof.}
We may directly verify the first two and the last conditions in Definition~\ref{def:monotonouslinearinstance} through the construction above. The only thing we need to check is the third condition. It suffices to show that 
$\left\{\frac{A_{i,j}}{A_{i,K}}-\frac{B_{1,j}}{B_{i,K}}\right\}_{j\in[K-1]}$
 for each $i\in[n_A]$ is a positive and strictly decreasing sequence. This follows true as $\{A_{i,j}\}_{j\in[K]}$ and $\{-B_{1,j}\}_{j\in[K]}$ are both strictly decreasing positive sequences.
\hfill\Halmos
\endproof

\subsection{Proof of Proposition~\ref{prop:multi-phase-cost-competitive-ratio}}
\label{sec:proof-of-multi-phase-cost-competitive-ratio}
We have constructed a multi-phase minimax-MDP $\mathcal{W}$ in Section~\ref{sec:optimizing-robust-competitive-ratio-in-T-O-K}. There are two non-trivial constraints at time $\tau_K$,
\begin{align}
    &\sum\limits_{j=1}^{K}\gamma_{j}\cdot x_{T+1,j} \leq p\cdot d-\Phi\cdot \mathcal{R}^\sharp (\mathcal{P}, d; \mathcal{V}),\qquad\qquad
    &\sum\limits_{j=1}^{K}(p-\gamma_{j})\cdot x_{T+1,j}\geq \Phi\cdot \mathcal{R}^\sharp (\mathcal{P}, d; \mathcal{V}),
\end{align}
where we may verify that $p\cdot d-\Phi\cdot \mathcal{R}^\sharp (\mathcal{P}, d; \mathcal{V})$ and $\Phi\cdot \mathcal{R}^\sharp (\mathcal{P}, d; \mathcal{V})$ are two continuous, positive valued, strictly increasing and piecewise linear functions with at most $K+1$ linear pieces. 

We next construct an equivalent monotonous $(K+1)$-linear instance $\mathcal{Q}_K$. Invoking Lemma~\ref{lem:decomposition-of-range-max-linear-function}, we may construct $\Phi\cdot \mathcal{R}^{\sharp,r} (\mathcal{P}, d; \mathcal{V})$ for each $r\in[K+1]$ defined as $\mathcal{R}^{\sharp} (\mathcal{P}, d; \mathcal{V})=\max_{r\in[K+1]}\mathcal{R}^{\sharp,r} (\mathcal{P}, d; \mathcal{V})$, with each $\mathcal{R}^{\sharp,r} (\mathcal{P}, d; \mathcal{V})$ being continuous, piecewise linear functions with at most $12$ linear pieces. And
we construct $\mathcal{Q}_K$ the same as $\mathcal{W}$ except for the environment-dependent inventory constraints at time $\tau_K$, which are replaced by
\begin{align}
    \sum\limits_{j=1}^{K}\gamma_{j}\cdot x_{T+1,j} \leq p\cdot d-\Phi\cdot \mathcal{R}^\sharp (\mathcal{P}, d; \mathcal{V}), \qquad\qquad
    \sum\limits_{j=1}^{K}(p-\gamma_{j})\cdot x_{T+1,j}\geq \Phi\cdot \mathcal{R}^{\sharp,r} (\mathcal{P}, d; \mathcal{V}), ~~\forall r\in[K+1].\notag
\end{align}
Note that $\{\gamma_j\}_{j=1}^K$ is a positive and strictly increasing sequence with $\gamma_K< p$, we can deduce that $\mathcal{Q}_K$ is a monotonous $(K+1)$-linear instance. 

Then we can use the constructed phase reduction in Section~\ref{sec:phase-reduction-for-Q} to solve the problem. We denote the phase reduction instance sequence as $\{\mathcal{Q}_K,\mathcal{Q}_{K-1}\dots,\mathcal{Q}_0\}$. The remaining thing is the total time complexity of the whole $K$ reductions. We use $n_A^i$ and $n_B^i$ to denote the ``$\geq$'' and ``$\leq$'' multi-phase inventory constraints in instance $\mathcal{Q}_i$, while we can directly verify that $n_B^i=1$ for each $i\in[K]$. We can also deduce that 
\begin{align*}
    n_A^i=12\cdot n_A^{i+1}+12\cdot (i+1)\cdot n_A^{i+1} \cdot (\tau_{i+1}-\tau_i) \leq 30\cdot K\cdot T\cdot n_A^{i+1} \leq n_A^{K}\cdot (30\cdot K\cdot T)^{K-i} = T^{O(K)}.
\end{align*}
Moreover, we can derive from the construction in Section~\ref{sec:phase-reduction-for-Q} that the time complexity of reduction from $\mathcal{Q}_i$ to $\mathcal{Q}_{i-1}$ is $O(i\cdot n_A^i\cdot n_B^i \cdot (\tau_{i}-\tau_{i-1}))$. Thus the total time complexity is $O(K\cdot T\cdot T^{O(K)})=T^{O(K)}$. The corresponding feasible policy can also be directly deduced during the reduction process.

\subsection{Proof of Lemma~\ref{lem:decomposition-of-range-max-linear-function}}
\label{sec:proofofsplit}
We first prove the continuity of $w(\cdot)$. As $g(\cdot)$ is piecewise linear and continuous with finite pieces, we deduce that $g(\cdot)$ is Lipschitz continuous and assume the Lipschitz constant of $g(\cdot)$ is $L_{g}$. Actually, $L_{g}$ equals the maximum slope in $g(\cdot)$, which can be computed in $O(M)$ time. $L_{f}$ is defined in the same way. For any $x\in\mathbb{R}$, suppose $w(x)=g(y^{*})-f(z^{*})$ with $y^{*},z^{*}\in[x,x+\triangle]$ and $|y^{*}-z^{*}|\leq h$. Then for any $\delta\in\mathbb{R}$, we may verify that there exists $y'\in[x+\delta,x+\delta+\triangle]$ such that $|y'-y^{*}|\leq |\delta|$, and $z'\in[x+\delta,x+\delta+\triangle]$ with $|z'-y'|\leq h$ such that $|z'-z^{*}|\leq 4\cdot|\delta|$. We can derive that
\begin{align*}
w(x+\delta)\geq g(x')-f(z')\geq g(x^{*})-f(z^{*})-(L_{g}+4\cdot L_{f})\cdot |\delta| =w(x)-(L_{g}+4\cdot L_{f})\cdot |\delta|.
\end{align*}
Thus $w(\cdot)$ is Lipschitz continuous with constant $L_{g}+4\cdot L_{f}$.

Second, we present how to calculate $\{w_{i}\}_{i\in[M\cdot N]}$. Consider $g(\cdot)$'s breakpoints $-\infty=a_{0}<a_{1}<\dots<a_{M-1}<a_{M}=\infty$, i.e., $g(\cdot)$ is linear on each interval distinguished by these points. And further we define $\{g_{i}(\cdot)\}_{i\in[M]}$ by
\begin{equation*}
g_{i}(x)=
\begin{cases}
-\infty & x\notin [a_{i-1},a_{i}], \\[5pt]
r_{i}\cdot x+b_{i}    & x\in [a_{i-1},a_{i}],
\end{cases}
\end{equation*}
where for each $i\in[M]$, $g_{i}(x)=g(x)$ in $[a_{i-1},a_{i}]$. We have that $g(x)$ equals the maximum value in $\{g_{i}(x)\}_{i\in[M]}$. We define $\{f_{j}\}_{j\in[N]}$ in the same way:
\begin{equation*}
f_{j}(x)=
\begin{cases}
-\infty & x\notin [\widetilde{a}_{j-1},\widetilde{a}_{j}], \\[5pt]
\widetilde{r}_{j}\cdot x+\widetilde{b}_{j}    & x\in [\widetilde{a}_{j-1},\widetilde{a}_{j}],
\end{cases}
\end{equation*}
where $\{\widetilde{a}_{j}\}_{j\in[N]}$ are breakpoints and for each $j\in[N]$, $f_{j}(x)=f(x)$ in $[\widetilde{a}_{j-1},\widetilde{a}_{j}]$. Moreover, the decreasing-ness of $f$ conducts that $\tilde{r}_j\leq 0$ for each $j\in[N]$. We can derive that
\begin{align*}
w(x)=\mathop{\max}\limits_{y,z\in[x,x+\triangle] \atop |y-z|\leq h}(g(y)+f(z)) =\mathop{\max}\limits_{i\in[M] \atop j\in[N]}\mathop{\max}\limits_{y,z\in[x,x+\triangle] \atop |y-z|\leq h}(g_{i}(y)+f_{j}(z)) .
\end{align*}
We define $\hat{w}_{N\cdot(i-1)+j}(x)=\mathop{\max}\limits_{y,z\in[x,x+\triangle] \atop |y-z|\leq h}[g_{i}(y)+f_{j}(z)]$ for each $i \in [M]$ and $j \in [N]$. We then have
\begin{align}\label{eq:decomposition-without-simplification-w}
    w(x)=\mathop{\max}\limits_{i\in[M] \atop j\in[N]}\left[\hat{w}_{N\cdot(i-1)+j}(x)\right].
\end{align}
\begin{lemma}
\label{lem:linearpropertyofhij}
For each $i\in[M]$ and $j\in[N]$, there exist $L(i,j)$ and $R(i,j)$ such that $\hat{w}_{N\cdot(i-1)+j}(x)$ is a continuous piecewise linear bounded function with at most $10$ pieces in $[L(i,j),R(i,j)]$ and equals $-\infty$ outside $[L(i,j),R(i,j)]$. $L(i,j)$ and $R(i,j)$ can be $\infty$ or $-\infty$ and we don't assume that $[L(i,j),R(i,j)]\neq\emptyset$. Besides, the explicit form of $\hat{w}_{N\cdot(i-1)+j}(x)$ can be calculated in $O(1)$ time.
\end{lemma}
The proof of Lemma~\ref{lem:linearpropertyofhij} will be presented in Section~\ref{sec:proofofhwotodivide}. 

Finally, we change these functions $\{\hat{w}_{s}(\cdot)\}_{s\in[M\cdot N]}$ to continuous settings $\{w_{s}(\cdot)\}_{s\in[M\cdot N]}$ as follows:
\begin{equation*}
w_{N\cdot(i-1)+j}(x)=
\begin{cases}
\hat{w}_{N\cdot(i-1)+j}(L(i,j))-(L_{g}+4\cdot L_{f})\cdot(L(i,j)-x) & x<L(i,j), \\[5pt]
\hat{w}(x) & L(i,j)\leq x\leq R(i,j), \\[5pt]
\hat{w}_{N\cdot(i-1)+j}(R(i,j))-(L_{g}+4\cdot L_{f})\cdot(x-R(i,j)). & x>R(i,j)
\end{cases}
\end{equation*}
For the case where $[L(i,j),R(i,j)]=\emptyset$, we just delete the function.
Notice that each $w_{s}(\cdot)$ is a continuous and piecewise linear function with $12$ linear pieces. The left thing is to verify that $\mathop{\max}\limits_{s\in[N\cdot M]}w_{s}(x)= w(x)$. We have proved that $w(\cdot)$ is a continuous function with Lipschitz constant $L_{g}+4\cdot L_{f}$, thus for any $i\in[M]$, $j\in[N]$ and $s=N\cdot(i-1)+j$, we have
\begin{align*}
&\forall x<L(i,j),~w_{s}(x)=w_{s}(L(i,j))-(L_{g}+4\cdot L(f))\cdot (L(i,j)-x) \\
&\qquad =\hat{w}_{s}(L(i,j))-(L_{g}+4\cdot L_{f})\cdot (L(i,j)-x)\leq w(L(i,j))-(L_{g}+4\cdot L_{f})\cdot (L(i,j)-x) \leq w(x).
\end{align*}
Thus for any $x<L(i,j)$, $\hat{w}_{s}(x)<w_{s}(x)\leq w(x)$. Similarly, for any $x>R(i,j)$, $\hat{w}_{s}(x)<w_{s}(x)\leq w(x)$. We can derive that $
\mathop{\max}\limits_{s\in[N\cdot M]}\hat{w}_{s}(x)\leq \mathop{\max}\limits_{s\in[N\cdot M]}w_{s}(x)\leq w(x)$. Invoking Eq.~\eqref{eq:decomposition-without-simplification-w}, we deduce that 
$
\mathop{\max}\limits_{s\in[N\cdot M]}w_{s}(x)= w(x),
$
with each $w_s(\cdot)$ being a continuous and piecewise linear function with at most $12$ linear pieces. Last but not least, we need to confirm the time complexity of computation. In fact, the computation of Lipschitz constant of $g(\cdot)$ and $f(\cdot)$ is $O(N+M)$ which we have stated at the beginning. As for each $w_{s}(\cdot)$, we have given the explicit expression using Lemma~\ref{lem:linearpropertyofhij}, thus the time complexity of the construction is also $O(N\cdot M)$.

\subsection{Proof of Lemma~\ref{lem:linearpropertyofhij}}
\label{sec:proofofhwotodivide}
For convenience, we introduce the following notations, which are only used in this section:
\begin{align*}
w(x)=\mathop{\max}\limits_{y,z\in[x,x+\triangle] \atop |y-z|\leq h}[g(y)+f(z)], \quad \text{where}\quad
g(y)=
\begin{cases}
-\infty & y\notin [a,b] \\[5pt] 
r\cdot y & y\in[a,b]
\end{cases},
\quad
f(z)=
\begin{cases}
-\infty & y\notin [c,d] \\[5pt] 
\widetilde{r}\cdot z & y\in[c,d].
\end{cases} ,
\end{align*}
with $\widetilde{r}\leq 0$. The goal of this section is to calculate $w(x)$. We have 
\begin{align*}
w(x)=\max\{r\cdot y+\widetilde{r}\cdot z;~y,z\in[x.x+\triangle],y\in[a,b],z\in[c,d],|y-z|\leq h\}.
\end{align*} 
We assume $a\leq b$ and $c\leq d$, otherwise $w(\cdot)\equiv -\infty$. Moreover, if $a>c+h$, then 
\begin{align*}
w(x)&=\max\{r\cdot y+\widetilde{r}\cdot z;~y,z\in[x,x+\triangle],y\in[a,b],z\in[c,d],|y-z|\leq h\} \\
&=\max\{r\cdot y+\widetilde{r}\cdot z;~y,z\in[x,x+\triangle],y\in[a,b],z\in[a-h,d],|y-z|\leq h\}.
\end{align*}
Similarly, we may assume $|a-c|\leq h$, $|b-d|\leq h$ and $h\leq\triangle$. Also, we can notice that $w(x)=-\infty$ when $x\notin U\defeq[\max(a,c)-\triangle,\min(b,d)]$. Thus we assume $U\neq\emptyset$. Having the above assumptions, we calculate the situations when $x\in U$.

We first analyze the value of $z$ when achieving maximum. If $y$ is settled, then $z$ has to satisfy $z\geq \max(y-h,c,x)$ and $z\leq \min(y+h,d,x+\triangle)$. If $y\in[x,x+\triangle]\cap[a,b]$, one may verify that $\max(y-h,c,x)\leq \min(y+h,d,x+\triangle)$. From the condition that $\widetilde{r}\leq 0$, we derive that when $y\in[x,x+\triangle]\cap[a,b]$, to achieve the maximum, $z$ equals $\max(y-h,c,x)$. Therefore, it can be derived that for $x\in U$, we have
\begin{align*}
w(x)=\max\{r\cdot y+\widetilde{r}\cdot\max(y-h,c,x);~y\in[a,b]\cap[x,x+\triangle]\}>-\infty .
\end{align*} 

We then prove that $w(x)$ is piecewise linear with at most $10$ pieces in $U$. Notice that
\begin{equation*}
w(x)=
\begin{cases}
h_1(x)\defeq\max\{r\cdot y+\widetilde{r}\cdot\max(y-h,x);~y\in[a,b]\cap[x,x+\triangle]\} & \text{if}~c\leq x\leq\min(b,d), \\[5pt]
h_2(x)\defeq\max\{r\cdot y+\widetilde{r}\cdot\max(y-h,c);~y\in[a,b]\cap[x,x+\triangle]\} & \text{if}~\max(a,c)-\triangle\leq x\leq c.
\end{cases}
\end{equation*}
Thus, it suffices to show that both $h_{1}(\cdot)$ and $h_{2}(\cdot)$ are piecewise linear with at most $5$ pieces in $U$. We may verify the following calculation results for $h_1(\cdot)$ and $h_2(\cdot)$:
\begin{enumerate}
\item $r\leq 0$. We have
\begin{equation*}
h_{1}(x)=
\begin{cases}
r\cdot a+\widetilde{r}\cdot\max(a-h,x) & \text{when}~a-\triangle\leq x\leq a, \\[5pt]
r\cdot x+\widetilde{r}\cdot x & \text{when}~a\leq x\leq b.
\end{cases}
\end{equation*}
\begin{equation*}
h_{2}(x)=
\begin{cases}
r\cdot a+\widetilde{r}\cdot\max(a-h,c) & \text{when}~a-\triangle\leq x\leq a. \\[5pt]
r\cdot x+\widetilde{r}\cdot\max(x-h,c) & \text{when}~a\leq x\leq b,
\end{cases}
\end{equation*}
\item : $-\widetilde{r}\geq r\geq 0$. We have 
\begin{equation*}
h_{1}(x)=
\begin{cases}
r\cdot a+\widetilde{r}\cdot(a-h) & \text{when}~a-\triangle\leq x\leq a-h, \\[5pt]
r\cdot (x+h)+\widetilde{r}\cdot x  & \text{when}~a-h\leq x\leq b-h, \\[5pt]
r\cdot b+\widetilde{r}\cdot x & \text{when}~b-h\leq x\leq b. 
\end{cases}
\end{equation*}
If $b\geq h+c\geq a$, then 
\begin{equation*}
h_{2}(x)=
\begin{cases}
r\cdot (x+\triangle)+\widetilde{r}\cdot c & \text{when}~a-\triangle\leq x\leq h+c-\triangle, \\[5pt]
r\cdot (h+c)+\widetilde{r}\cdot c & \text{when}~h+c-\triangle\leq x\leq h+c, \\[5pt]
r\cdot x+\widetilde{r}\cdot(x-h) & \text{when}~h+c\leq x\leq b.
\end{cases}
\end{equation*}
If $h+c\geq b$, then 
\begin{equation*}
h_{2}(x)=
\begin{cases}
r\cdot(x+\triangle)+\widetilde{r}\cdot c & \text{when}~a-\triangle\leq x\leq b-\triangle, \\[5pt]
r\cdot b+\widetilde{r}\cdot c & \text{when}~b-\triangle\leq x\leq b.
\end{cases}
\end{equation*}
\item $r\geq -\widetilde{r}\geq 0$. We have 
\begin{equation*}
h_{1}(x)=
\begin{cases}
r\cdot(x+\triangle)+\widetilde{r}\cdot(x+\triangle-h) & \text{when}~a-\triangle\leq x\leq b-\triangle, \\[5pt]
r\cdot b+\widetilde{r}\cdot(b-h,x) & \text{when}~b-\triangle\leq x\leq b.
\end{cases}
\end{equation*}
\begin{equation*}
h_{2}(x)=
\begin{cases}
r\cdot (x+\triangle)+\widetilde{r}\cdot\max(x+\triangle-h,c) & \text{when}~a-\triangle\leq x\leq b-\triangle, \\[5pt]
r\cdot b+\widetilde{r}\cdot\max(b-h,c) & \text{when}~b-\triangle\leq x\leq b.
\end{cases}
\end{equation*}
\end{enumerate}
Apparently, all quantities above can be computed in $O(1)$ time.

\subsection{Proof of Lemma~\ref{lem:competitive-ratio-between-pi-V-pi-hat-V}}
\label{sec:proof-of-competitive-ratio-between-pi-V-pi-hat-V}
Consider any regular prediction sequence $\mathcal{P}$ and demand $d$. One may verify that the inventory level of $\mathcal{V}$ under policy $\pi^{\mathcal{V}}$ equals the inventory level of $\hat{\mathcal{V}}$ under policy $\pi^{\hat{\mathcal{V}}}$, denoted by $\bm{x}$. One may also verify that the final inventory level of $\mathcal{V}$ under hindsight optimal policy equals the final inventory level of $\hat{\mathcal{V}}$ under the hindsight optimal policy, denoted by $\bm{x}^*$. We have
\begin{align}
    \varphi_{\pi^{\mathcal{V}}} (\mathcal{P}, d; \mathcal{V})=\frac{p\cdot\min(d,\sum_{j=1}^K x_i)-\sum_{j=1}^K \gamma_i\cdot x_i}{\sum_{j=1}^K (p-\gamma_i)\cdot x_i^*}, \qquad \varphi_{\pi^{\hat{\mathcal{V}}}} (\mathcal{P}, d; \hat{\mathcal{V}})=\frac{p\cdot\min(d,\sum_{j=1}^K x_i)-\sum_{j=1}^K \hat{\gamma}_i\cdot x_i}{\sum_{j=1}^K (p-\hat{\gamma}_i)\cdot x_i^*}. 
\end{align}
We discuss it case by case as follows.
\begin{enumerate}
    \item $d\geq \sum_{j=1}^K x_i$: we can derive that
    \begin{align}
        \varphi_{\pi^{\mathcal{V}}} (\mathcal{P}, d; \mathcal{V})&=\frac{\sum_{j=1}^K (p-\gamma_i)\cdot x_i}{\sum_{j=1}^K (p-\gamma_i)\cdot x_i^*} \geq (1+\delta)^{-1}\cdot\frac{\sum_{j=1}^K (p-\hat{\gamma}_i)\cdot x_i}{\sum_{j=1}^K (p-\gamma_i)\cdot x_i^*} \notag\\
        &\geq (1+\delta)^{-1}\cdot\frac{\sum_{j=1}^K (p-\hat{\gamma}_i)\cdot x_i}{\sum_{j=1}^K (p-\hat{\gamma}_i)\cdot x_i^*}=(1+\delta)^{-1}\cdot\varphi_{\pi^{\hat{\mathcal{V}}}} (\mathcal{P}, d; \widehat{\mathcal{V}}) , \notag
    \end{align}
    where the two inequalities are due to Eq.~\eqref{eq:grid-relationship-between-cost}. 
    \item $d\leq \sum_{j=1}^K x_i$: consider $0\leq x'_i\leq x_i$ such that $d= \sum_{j=1}^K x'_i$ and denote $x''_i$ as $x_i-x'_i$ for each $i\in[K]$. Then we have 
    \begin{align}
        &\varphi_{\pi^{\hat{\mathcal{V}}}} (\mathcal{P}, d; \widehat{\mathcal{V}})=\frac{\sum_{j=1}^K (p-\hat{\gamma}_i)\cdot x'_i-\sum_{j=1}^K \hat{\gamma}_i\cdot x''_i}{\sum_{j=1}^K (p-\hat{\gamma}_i)\cdot x_i^*}\geq \Phi \geq 0,  \label{eq:phi-pi-hat-V-P-d-hat-V}\\
        &\varphi_{\pi^{\mathcal{V}}} (\mathcal{P}, d; \mathcal{V})=\frac{\sum_{j=1}^K (p-\gamma_i)\cdot x'_i-\sum_{j=1}^K \gamma_i\cdot x''_i}{\sum_{j=1}^K (p-\gamma_i)\cdot x_i^*} \notag \\
        &\quad=\frac{\sum_{j=1}^K (p-\hat{\gamma}_i)\cdot x'_i-\sum_{j=1}^K \hat{\gamma}_i\cdot x''_i}{\sum_{j=1}^K (p-\gamma_i)\cdot x_i^*}+\frac{\sum_{j=1}^K \left[(p-\gamma_i)-(p-\hat{\gamma}_i)\right]\cdot x'_i-\sum_{j=1}^K (\gamma_i-\hat{\gamma}_i)\cdot x''_i}{\sum_{j=1}^K (p-\gamma_i)\cdot x_i^*}.  \label{eq:phi-pi-V-P-d-V}
    \end{align}
    Due to Eq.~\eqref{eq:grid-relationship-between-cost}, the second term in Eq.~\eqref{eq:phi-pi-V-P-d-V} is greater than 
    \begin{align}
        -\frac{\delta}{1+\delta}\cdot\frac{\sum_{j=1}^K (p-\hat{\gamma}_i)\cdot x'_i}{\sum_{j=1}^K (p-\gamma_i)\cdot x_i^*}&-\delta\cdot \frac{\sum_{j=1}^K \hat{\gamma}_i \cdot x''_i}{\sum_{j=1}^K (p-\gamma_i)\cdot x_i^*} \\
        \geq &-\delta\cdot\frac{\sum_{j=1}^K (p-\hat{\gamma}_i)\cdot x'_i}{\sum_{j=1}^K (p-\hat{\gamma}_i)\cdot x_i^*}-\delta\cdot(1+\delta)\cdot \frac{\sum_{j=1}^K \hat{\gamma}_i \cdot x''_i}{\sum_{j=1}^K (p-\hat{\gamma}_i)\cdot x_i^*}
    \end{align}
    Notice that $\bm{x}^*$ is the inventory level under hindsight optimal policy, invoking the condition that $0\leq x'_i\leq x_i$ and $d= \sum_{j=1}^K x'_i$, we have $\sum_{j=1}^K (p-\hat{\gamma}_i)\cdot x'_i\leq \sum_{j=1}^K (p-\hat{\gamma}_i)\cdot x_i^*$. Due to Eq.~\eqref{eq:phi-pi-hat-V-P-d-hat-V}, we have $
    \sum_{j=1}^K \hat{\gamma}_i \cdot x''_i\leq \sum_{j=1}^K (p-\hat{\gamma}_i)\cdot x'_i \leq \sum_{j=1}^K (p-\hat{\gamma}_i)\cdot x_i^*$. Invoking above estimation, we have 
    \begin{align}
        \varphi_{\pi^{\mathcal{V}}} (\mathcal{P}, d; \mathcal{V})&\geq \frac{\sum_{j=1}^K (p-\hat{\gamma}_i)\cdot x'_i-\sum_{j=1}^K \hat{\gamma}_i\cdot x''_i}{\sum_{j=1}^K (p-\gamma_i)\cdot x_i^*}- \delta-\delta\cdot(1+\delta) \notag \\
        &\geq \frac{\sum_{j=1}^K (p-\hat{\gamma}_i)\cdot x'_i-\sum_{j=1}^K \hat{\gamma}_i\cdot x''_i}{\sum_{j=1}^K (p-\hat{\gamma}_i)\cdot x_i^*}- 3\delta=\varphi_{\pi^{\hat{\mathcal{V}}}} (\mathcal{P}, d; \widehat{\mathcal{V}})-3\delta, \notag
    \end{align}
    while the second inequality is due to Eq.~\eqref{eq:grid-relationship-between-cost}.
\end{enumerate}
Combining the above discussions, we have $\varphi_{\pi^{\mathcal{V}}} (\mathcal{P}, d; \mathcal{V})\geq \varphi_{\pi^{\hat{\mathcal{V}}}} (\mathcal{P}, d; \widehat{\mathcal{V}})-3\delta$.
\end{document}